\documentclass[twoside,11pt]{article}

\usepackage{amsmath,amsfonts,bm}

\def\eqref#1{equation~\ref{#1}}

\def\ceil#1{\lceil #1 \rceil}

\def\1{\bm{1}}

\def\rmS{{\mathbf{S}}}

\def\vv{{\bm{v}}}

\def\vx{{\bm{x}}}
\def\vy{{\bm{y}}}

\def\mE{{\bm{E}}}

\def\mI{{\bm{I}}}

\def\mW{{\bm{W}}}
\def\mX{{\bm{X}}}
\def\mY{{\bm{Y}}}

\DeclareMathAlphabet{\mathsfit}{\encodingdefault}{\sfdefault}{m}{sl}
\SetMathAlphabet{\mathsfit}{bold}{\encodingdefault}{\sfdefault}{bx}{n}

\def\gD{{\mathcal{D}}}

\def\gF{{\mathcal{F}}}

\def\gK{{\mathcal{K}}}

\def\sS{{\mathbb{S}}}

\def\sX{{\mathbb{X}}}

\newcommand{\colonequiv}{\vcentcolon\mspace{-1.2mu}\equiv}

\usepackage{subcaption}
\usepackage{mathtools}
\usepackage{tikz}
\usetikzlibrary{fit}
\usepackage{blkarray}
\usepackage{paralist}
\usepackage{algorithm}
\usepackage[noend]{algpseudocode}
\usetikzlibrary{backgrounds}

\usepackage{booktabs}
\usepackage{multirow}

\usepackage{tabularx}
\usepackage[export]{adjustbox}

\usepackage[preprint]{jmlr2e}

\tikzset{%
	highlight/.style={rectangle,draw,thick,blue,inner xsep=1pt, inner ysep=-2pt}
}
\newcommand{\tikzmark}[2]{\tikz[overlay,remember picture,
	baseline=(#1.base)] \node (#1) {#2};}
\newcommand{\Highlight}[3][submatrix]{%
	\tikz[overlay,remember picture]{
		\node[highlight,fit=(#2.north west) (#3.south east)] (#1) {};}
}

\jmlrheading{volume}{2022}{X-XX}{X/XX}{XX/XX}{paper-id}{Edith Heiter, Robin Vandaele, Tijl De Bie, Yvan Saeys, Jefrey Lijffijt}

\ShortHeadings{Topologically Regularized Data Embeddings}{Heiter, Vandaele, De Bie, Saeys and Lijffijt}
\firstpageno{1}

\begin{document}

\title{Topologically Regularized Data Embeddings}

\author{\name Edith Heiter\textsuperscript{$\dagger$} \email edith.heiter@ugent.be
		\AND
		\name Robin Vandaele\textsuperscript{$\dagger, \S, \between$} \email robin.vndaele@gmail.com
   	   	\AND
		\name Tijl De Bie\textsuperscript{$\dagger$} \email tijl.debie@ugent.be
		\AND
		\name Yvan Saeys\textsuperscript{$\S, \between$} \email Yvan.Saeys@ugent.be
		\AND
		\name Jefrey Lijffijt\textsuperscript{$\dagger$} \email jefrey.lijffijt@ugent.be\\
		\textsuperscript{$\dagger$}\addr IDLab, Department of Electronics and Information Systems, 
		Ghent University, Ghent, Belgium\\
		\textsuperscript{$\S$}\addr Department of Applied Mathematics, Computer Science and Statistics, 
		Ghent University, Ghent, Belgium\\
		\textsuperscript{$\between$}\addr Data mining and Modelling for Biomedicine (DaMBi), 
		VIB Inflammation Research Center, Ghent, Belgium
		}
\editor{TBD}

\maketitle

\begin{abstract}%
Unsupervised representation learning methods are widely used for gaining insight into high-dimensional, unstructured, or structured data. In some cases, users may have prior topological knowledge about the data, such as a known cluster structure or the fact that the data is known to lie along a tree- or graph-structured topology. However, generic methods to ensure such structure is salient in the low-dimensional representations are lacking. This negatively impacts the interpretability of low-dimensional embeddings, and plausibly downstream learning tasks.
To address this issue, we introduce topological regularization: a generic approach based on algebraic topology to incorporate topological prior knowledge into low-dimensional embeddings. We introduce a class of topological loss functions, and show that jointly optimizing an embedding loss with such a topological loss function as a regularizer yields embeddings that reflect not only local proximities but also the desired topological structure.
We include a self-contained overview of the required foundational concepts in algebraic topology, and provide intuitive guidance on how to design topological loss functions for a variety of shapes, such as clusters, cycles, and bifurcations. We empirically evaluate the proposed approach on computational efficiency, robustness, and versatility in combination with linear and non-linear dimensionality reduction and graph embedding methods.

\end{abstract}

\begin{keywords}
Topological data analysis, persistent homology, representation learning, dimensionality reduction, graph embedding, topological regularization
\end{keywords}

\section{Introduction\label{SEC::intro}}

Data embedding methods are commonly used in data science to convert raw input data into a form that is more suitable for learning and consecutive inference. 
For example, linear dimensionality reduction methods such as Principal Component Analysis \citep[PCA;][]{jolliffe1986principal} and Linear Discriminant Analysis  \citep[LDA;][]{mclachlan2005discriminant} are commonly used to preprocess data for unsupervised or supervised learning. 
Non-linear dimensionality reduction methods such as Diffusion Maps \citep{coifman2005geometric}, Uniform Manifold Approximation and Projection \citep[UMAP;][]{mcinnes2018umap}, and $t$-Stochastic Neighbor Embedding \citep[$t$-SNE;][]{van2008visualizing}, are often used to preprocess or visualize high-dimensional data. 
For relational input data, graph embedding methods such as DeepWalk \citep{perozzi2014deepwalk} allow one to represent the data in a matrix structure that is convenient to visualize or use as an input for common machine learning algorithms.

One of the main arguments for using data embeddings is that they improve the signal-to-noise ratio in the data,
making meaningful information in the data more salient. %
This facilitates robust and automated inference from the data, as well as human understanding and interaction when the embeddings are visualized in a two-dimensional (2D) or three-dimensional (3D) space. 
Therefore, data embeddings are invaluable for summarizing the important high-level structure in data in a way that is useful for both automated and interactive learning in a wide variety of application areas and scientific domains.

\begin{figure}[hbp]
	\begin{subfigure}[t]{\linewidth}
		\centering
		\includegraphics[width=\linewidth]{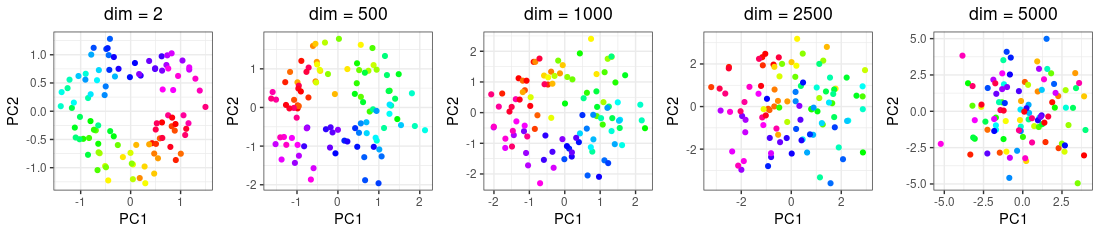}  
		\caption{\label{NoisyBallUniform}}
	\end{subfigure}
	\begin{subfigure}[t]{\linewidth}
		\centering
		\includegraphics[width=\linewidth]{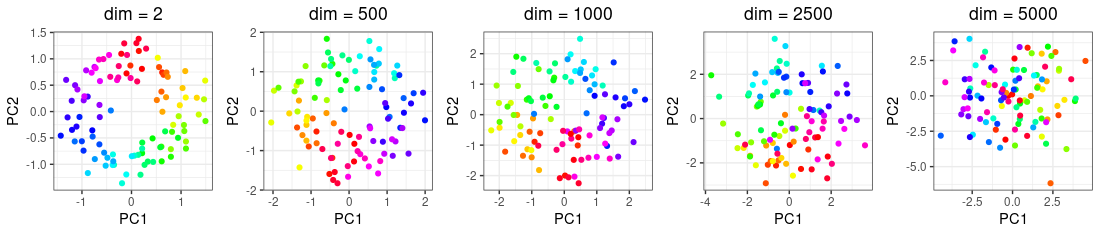}  
		\caption{\label{NoisyBallGaussian}}
	\end{subfigure}
	\caption{Two-dimensional PCA projections of a data set of 100 points on the unit circle with added (a) uniform and (b) Gaussian iid noise in high dimensions, with mean $\mu=0$ and standard deviation $\sigma=0.25$ in each dimension.
		The coloring of points marks their circular angle without noise, and the title of each plot denotes the input dimensionality.
		PCA increasingly struggles to capture the circular topology for increasing dimensionalities.\label{NoisyBalls}}
\end{figure}

Unfortunately, data embeddings themselves are susceptible to the noise they commonly aim to reduce \citep{vandaele2022curse}. 
For example, consider a high-dimensional noisy data set containing a cycle topology as meaningful structure. 
When projecting this data onto its 2D PCA plane, the noise in each dimension will likely shift the PCA plane away from its optimal noise-free orientation, in this way reducing the saliency of the cycle as the dimensionality increases, as illustrated in Figure~\ref{NoisyBalls}.

In the case of graphs, the relevant information may consist of a certain community structure.
Yet, random links between these communities may cause graph embeddings to fail to effectively separate them for data visualization purposes or consecutive learning tasks. 
In other cases, embeddings may truthfully capture the higher-order structure in the data, but the user may be interested in finer or coarser structure that cannot be captured well in a fully automated manner, such as subclusters of a main cluster in the embedding.  
\emph{These are examples where the use of topological prior knowledge is required or desired by the user}. 

These examples motivate the need for a method that can integrate topological information into data embeddings.
In this paper, we thus introduce the concept of \emph{topological regularization}, and a generic approach  for finding \emph{topologically regularized data embeddings} that operationalizes this concept. At a high level, the proposed approach is to find a (matrix) embedding $\mE$ of a data set (a point cloud, a graph,\ldots) $\sX$, with each row from $\mE$ corresponding to the embedding of a corresponding element from the data set (a data point, a vertex from the graph,\ldots), that minimizes a total loss function
\begin{equation}
	\label{totalloss1}
	\mathcal{L}_{\mathrm{tot}}(\mE, \sX)\coloneqq \mathcal{L}_{\mathrm{emb}}(\mE, \sX)+\lambda_{\mathrm{top}} \mathcal{L}_{\mathrm{top}}(\mE),
\end{equation}
where the embedding loss function $\mathcal{L}_{\mathrm{emb}}$ typically aims to preserve a notion of proximity between data elements in the original data, and $\lambda_{\mathrm{top}}>0$ controls the strength of the \emph{topological regularization}.
As we will demonstrate, broad types of topological prior knowledge can be directly encoded through the \emph{topological loss function} $\mathcal{L}_{\mathrm{top}}$.
It is made possible by building on recent developments in the field of \emph{topological data analysis} (TDA), in particular, on \emph{persistent homology}-based optimization \citep{gabrielsson2020topology, solomon2021fast, carriere2021optimizing}.

The theory behind persistent homology---explained in detail in Section \ref{SEC::background}---builds on fundamental concepts from the field of algebraic topology, which is not a standard tool of data scientists.
This may limit accessibility of our work to an expert audience, which may limit adoption in practice.
For this reason, this paper contains an accessible and self-contained overview of relevant concepts of topological optimization, starting from the basic concepts in simplicial homology.
We then introduce topological regularization, and explain in an illustrative manner how to design topological loss functions.
After this, we include extensive computational and experimental analysis of topological regularization, where we study and explore its computational cost, robustness, parameter sensitivity, behavior for different regularization strengths, topological loss functions, and their applications and future challenges, clearly discussing also the existing limitations.

\subsection{Example of Topological Regularization}
To clarify the purpose and possible impact of topological regularization at a non-technical level, we begin by discussing a simple but illustrative example of its application to single-cell omics analysis. (It is discussed in full depth in Sec.~\ref{sec:effec_qualitative}.)

Single-cell omics include various types of data collected on the cell level, such as transcriptomics, proteomics and epigenomics.
Studying the topological model underlying data may lead to a better understanding of the dynamic processes of biological cells and the regulatory interactions involved therein.
Such dynamic processes can be modeled through trajectory inference methods, also called \emph{pseudotime analysis}, which order cells along a trajectory based on the similarities in their expression patterns \citep{Saelens276907}.

For example, the \emph{cell cycle} is a well known biological differentiation model that takes place in a cell as it grows and divides. 
The cell cycle consists of different stages, namely growth (G1), DNA synthesis (S), growth and preparation for mitosis (G2), and mitosis (M).
The latter two stages are often grouped together in a G2M stage.
By studying expression data of cells that participate in the differentiation model, one may identify the genes involved in and between particular stages of the cell cycle \citep{liu2017reconstructing}.
Pseudotime analysis allows such study by assigning to each cell a time during the differentiation process in which it occurs, and thus, the relative positioning of all cells within the cell cycle model.

Thus, \emph{the analysis of single cell cycle data constitutes a problem where prior topological information is available}.
As the signal-to-noise ratio is commonly low in high-dimensional expression data \citep{libralon2009pre, zhang2021noise}, this data is usually preprocessed through a dimensionality reduction method prior to automated pseudotime inference \citep{Cannoodt2016, Saelens276907}.
\emph{Topological regularization provides a tool to enhance the expected topological signal during the embedding procedure, and as such, facilitate automated inference that depends on this signal}.

\begin{figure}[t]
	\centering
	\begin{subfigure}[t]{.9\linewidth}
		\includegraphics[height=.4cm]{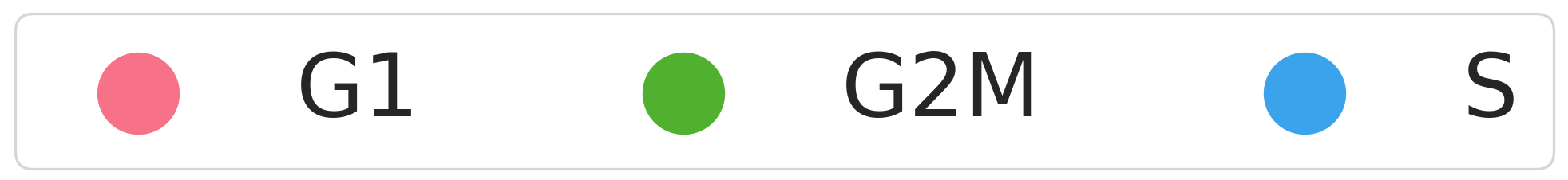}
	\end{subfigure}\\
	\begin{subfigure}[t]{.32\linewidth}
		\centering
		\includegraphics[height=3.7cm]{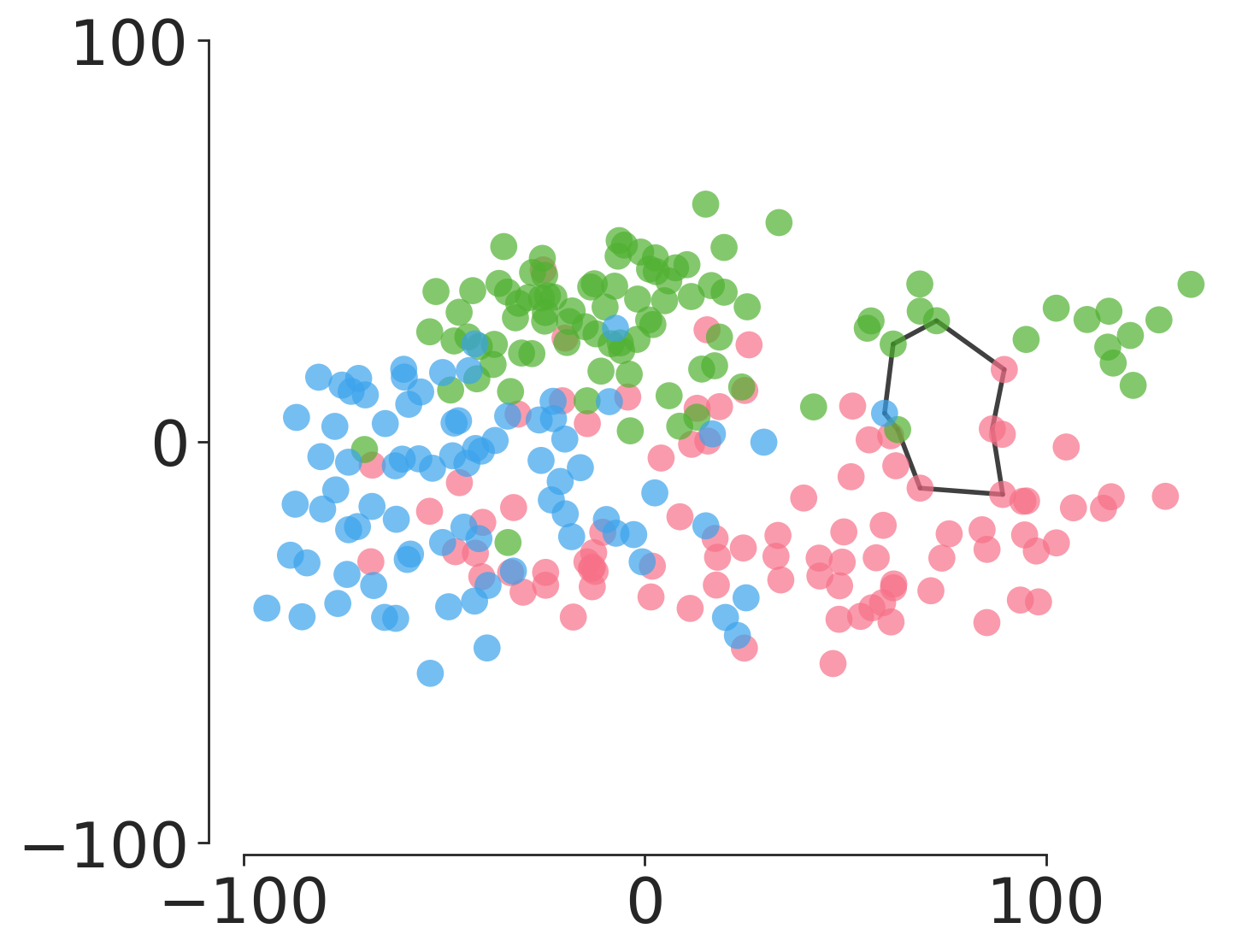}  
		\caption{The representation of the most prominent cycle obtained through persistent homology.}
	\end{subfigure}\hfill%
	\begin{subfigure}[t]{.3\linewidth}
		\centering
		\includegraphics[height=3.8cm]{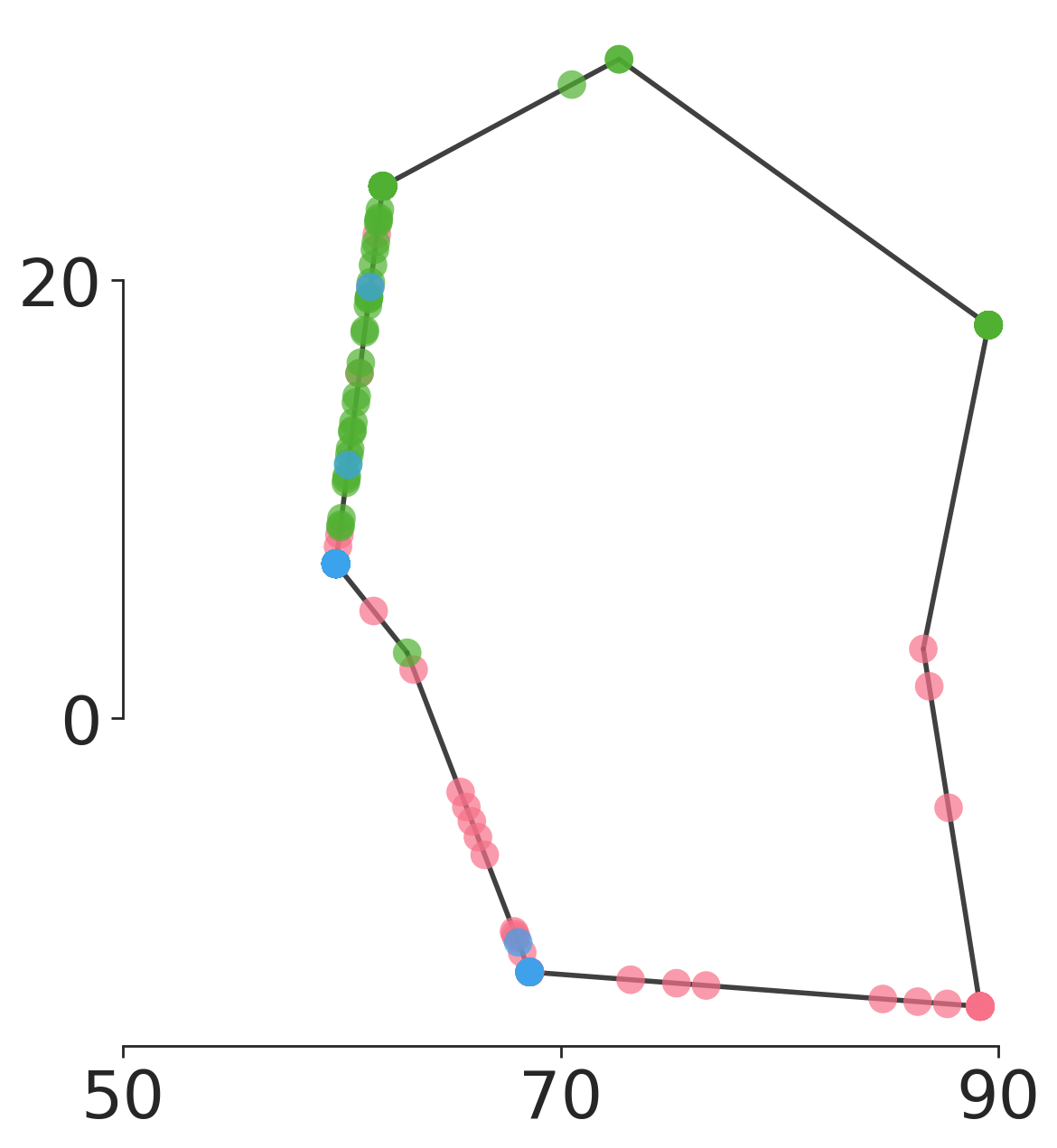}  
		\caption{Orthogonal projection of the embedded data onto the cycle representation.}
	\end{subfigure}\hfill%
	\begin{subfigure}[t]{.34\linewidth}
		\centering
		\includegraphics[height=3.8cm]{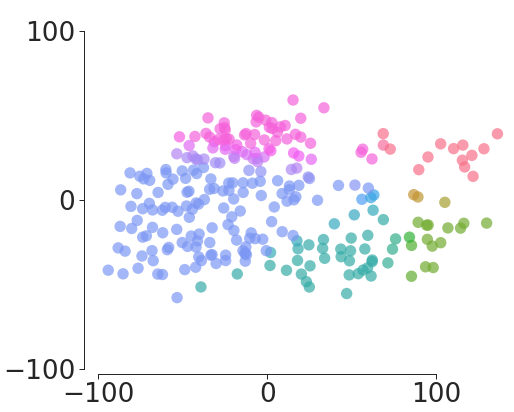}  
		\caption{The pseudotimes inferred from the projection in (b), quantified on a continuous color scale.}
	\end{subfigure}
	\caption{Automated pseudotime inference of real cell cycle data through persistent homology, from the PCA embedding of the data.\label{RepCyclePCA}}
\end{figure}

\begin{figure}[t]
	\centering
	\begin{subfigure}[t]{.9\linewidth}
		\includegraphics[height=.4cm]{Images/CellCycle/CellCircleLegend}
	\end{subfigure}\\
	\begin{subfigure}[t]{.32\linewidth}
		\centering
		\includegraphics[height=3.3cm]{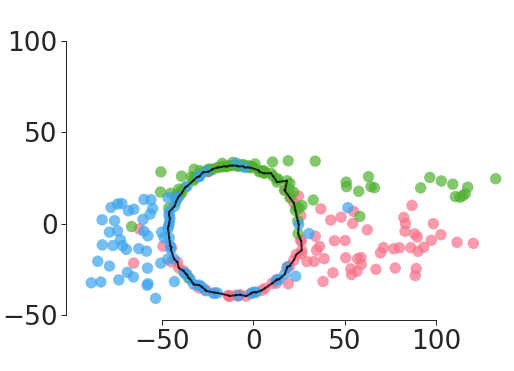}  
		\caption{The representation of the most prominent cycle obtained through persistent homology.}
	\end{subfigure}
	\hfill
	\begin{subfigure}[t]{.3\linewidth}
		\centering
		\includegraphics[height=3.3cm]{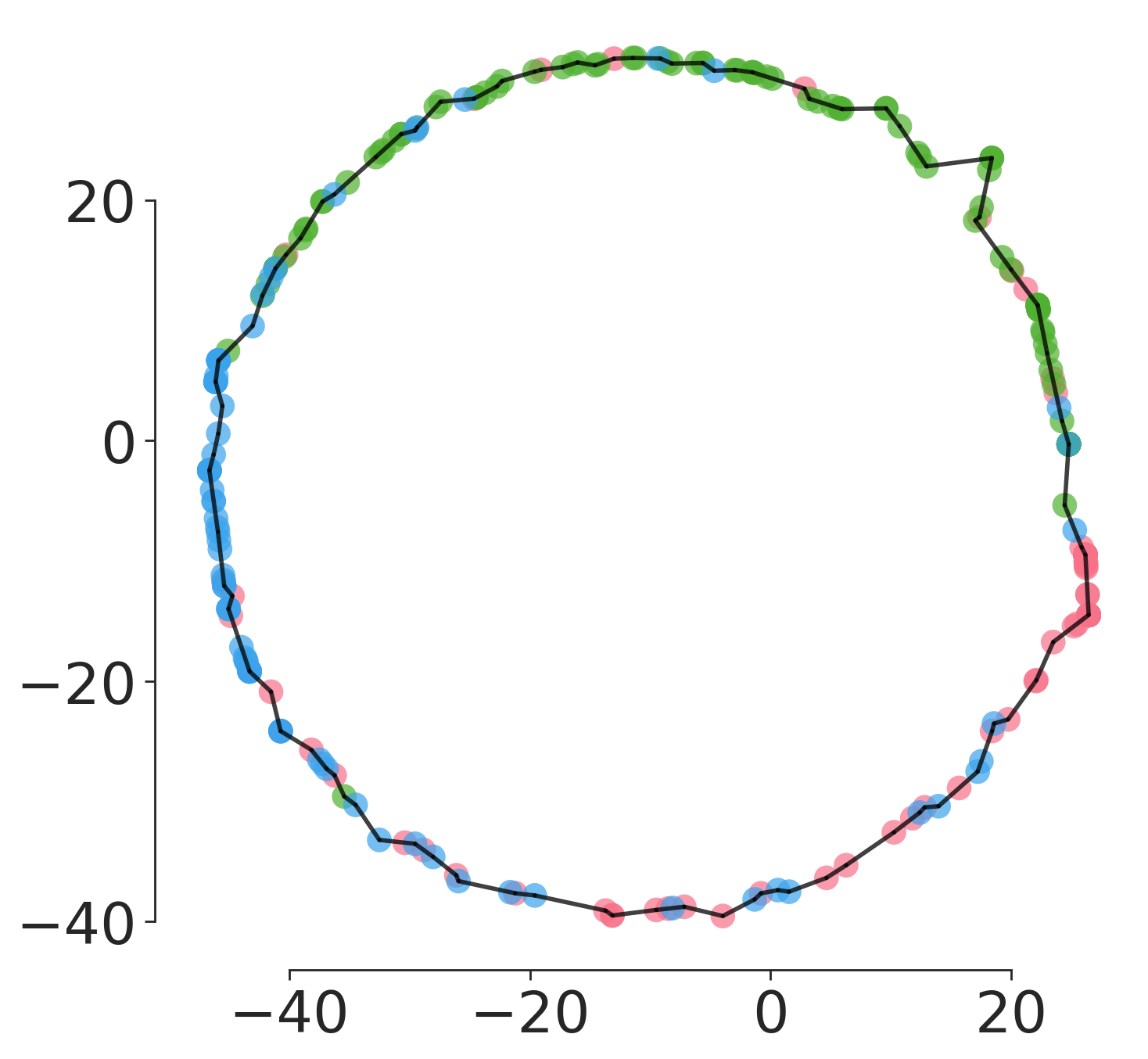}  
		\caption{Orthogonal projection of the embedded data onto the cycle representation.}
	\end{subfigure}
	\hfill
	\begin{subfigure}[t]{.34\linewidth}
		\centering
		\includegraphics[height=3.3cm]{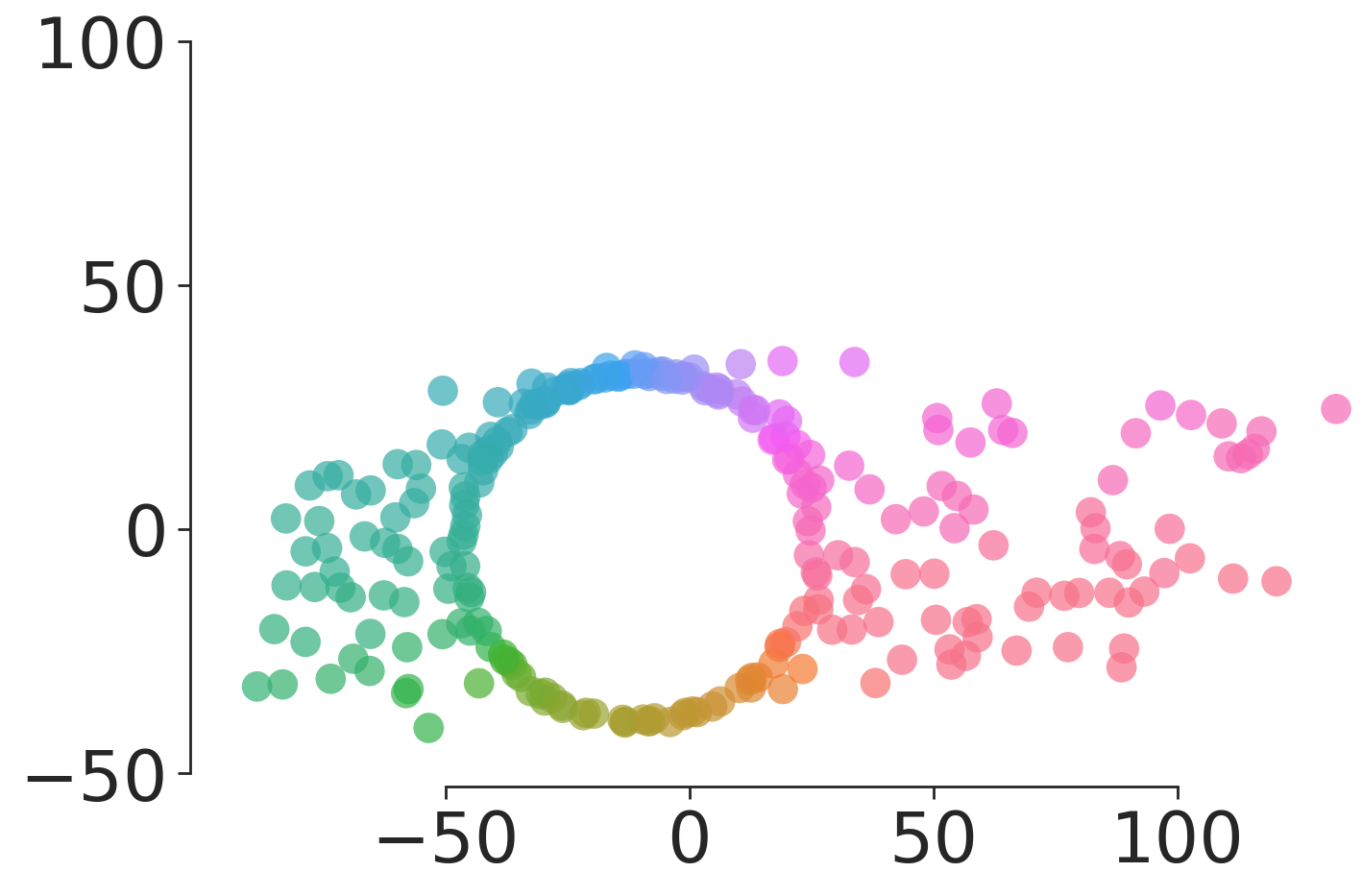}  
		\caption{The pseudotimes inferred from the projection in (b), quantified on a continuous color scale.}
	\end{subfigure}
	\caption{Automated pseudotime inference of real cell cycle data through persistent homology, from the topologically regularized PCA embedding of the data.}
	\label{RepCycleTop}
\end{figure}

To illustrate this, we applied an automated (cell) cycle and pseudotime inference method based on persistent homology (see Sec.~\ref{sec:effec_qualitative} for full details) on the real cell cycle data presented in \citep{buettner2015computational}, without (Figure~\ref{RepCyclePCA}) and with (Figure~\ref{RepCycleTop}) our proposed topological regularization of a PCA embedding of the data.
Hereby, we designed the topological loss function to bias the embedding towards a circular model.
Similar to the embedding in Figure \ref{NoisyBalls}, the ordinary PCA embedding struggles to capture the circular topology (Figure \ref{RepCyclePCA}~a).
Therefore, the cycle that is captured in an automated manner is rather spurious, and mostly linearly separates G1 and G2M cells.
Projecting cells onto the edges of this cycle---which is an intermediate step to derive continuous pseudotimes from a discretized topological representation as earlier described in \citet{Saelens276907}---places the majority of the cells onto a single edge (Figure~\ref{RepCyclePCA}~b).
The resulting pseudotimes are mostly continuous for cells projected onto this edge, whereas they are more discretized for all other cells (Figure~\ref{RepCyclePCA}~c).
By incorporating prior topological knowledge into the PCA embedding, the cycle in the topologically regularized embedding that is captured in an automated manner better characterizes the transmission between the G1, G2M, and S stages in the cell cycle model (Figure~\ref{RepCycleTop}~a).
The automated procedure for pseudotime inference also reflects a more continuous transmission between the cell stages (Figures~\ref{RepCycleTop}~b and \ref{RepCycleTop}~c). 

\subsection{Contributions\label{SUBSEC::contributions}}

We introduce the concept of topological regularization, to effectively bias a data embedding towards prior topological information.

\begin{compactitem}
	\item In Section~\ref{SEC::background} we provide an extensive, self-contained, and illustrative introduction to topological optimization of point clouds (or thus embeddings), starting from the basic concepts in simplicial homology.
	\item In Section~\ref{SEC::topreg} we propose a class of topological loss functions, explaining how they can be designed for a variety of topological priors, ranging from clusters, cycles, bifurcations, etc., to any combination of these. Our proposal also includes a novel sampling approach to ensure the topological loss functions are both robust and meaningful, while at the same time reducing computational cost.
	\item In Section~\ref{SEC::experiments} we present an extensive computational and experimental analysis of topological regularization. We study its computational cost, robustness, parameter sensitivity, behavior for different regularization strengths, topological loss functions, or topological priors, and its applications and future challenges. In doing so, we also discuss and empirically analyse current limitations of the proposed approach.
	\item We conclude on our work in Section~\ref{SEC::discconc} and discuss possible applications as well as open challenges for topological regularization in Section~\ref{sec:future_work}.
\end{compactitem}
This paper is a significant extension of an earlier conference paper \citep{vandaele2021topologically}, including substantially more background, improvements to the method, and additional experiments and analysis.

\subsection{Related Work\label{SUBSEC::relwork}}

Here we discuss the main related research and how it relates to the current paper.

\subsubsection{Foundations of Topological Optimization}

Most directly, topological regularization is building on a series of recent papers \citep{gabrielsson2020topology, solomon2021fast, carriere2021optimizing}, that showed that topological optimization---thus optimizing for $\mathcal{L}_{\mathrm{top}}$ in (\ref{totalloss1})---is possible in various settings, and in which the mathematical foundation is developed.
Topological optimization is inherently difficult due to the combinatorial nature of how topological information is quantified from point clouds through persistent homology, as the mapping from input data to its persistence diagrams is highly non-linear without an explicit analytical representation.
To accommodate this, topological optimization makes use of Clarke subderivatives \citep{clarke1990optimization}, whose applicability to persistent homology builds on arguments from o-minimal geometry \citep{van1998tame, carriere2021optimizing}.
Thanks to this recent work \citep{gabrielsson2020topology,carriere2021optimizing}, powerful tools for topological optimization have been developed for software libraries such as PyTorch and TensorFlow.
This allows their use without deeper knowledge of the mathematical foundation of persistent homology. 

\subsubsection{Incorporating Topological Optimization into Embeddings}

Topological autoencoders \citep{moor2020topological}, DIPOLE  \citep[Distributed Persistence- Optimized Local Embeddings][]{wagner2021improving}, and Interleaving Dimension Reduction \citep{nelson2022topology} have already combined topological optimization with a data embedding procedure.
The main difference to our work is that the topological information used for optimization is obtained from the original high-dimensional data, and not passed as a prior as in this paper.
While this can be useful, obtaining such topological information heavily relies on distances between observations, which are often meaningless and unstable in high dimensions \citep{vandaele2022curse,aggarwal2001surprising}.
Furthermore, certain constructions such as the \emph{weak Alpha filtration} obtained from the \emph{Delaunay triangulation}---which we will use extensively for topological regularization and are discussed in detail in Section \ref{SEC::background}---are expensive to obtain from high-dimensional data \citep{cignoni1998dewall}, and are therefore best computed from a low-dimensional embedding.

\subsubsection{Incorporating Topological Information into Embeddings}

Besides topological regularization, other methods that incorporate topological information into data embeddings have been developed as well.
For example, Deep Embedded Clustering \citep{xie2016unsupervised} simultaneously learns feature representations and cluster assignments using deep neural networks.
Constrained embeddings of Euclidean data on spheres have also been studied by \citep{bai2015constrained}, and self-organizing stars have been developed to embed data according to a star-shaped topology with a given number of branches \citep{come2010self}.
The common thread in these methods is that they require an extensive development for one particular kind of input data and one particular kind of topological model.
Contrary to this, incorporating topological optimization into representation learning provides a unifying yet versatile approach towards combining data embedding methods with topological priors, that generalizes well to any input structure as long as the output is a point cloud.

\subsubsection{Other Forms of Topological Regularization in Machine Learning}

Changing topological properties of an object to improve consecutive learning and inference has known recent applications in image segmentation \citep{Vandaele2020} and supervised machine learning \citep{chen2019topological, gabrielsson2020topology}. 
For example, basic image smoothing (which can also be conducted through topological optimization; \citealp{gabrielsson2020topology}) decreases the prominence of spurious topological features and improves consecutive unsupervised segmentation of skin lesions \citep{Vandaele2020}.
For supervised learning, topological regularization has been used to prevent spurious topological components in classification boundaries \citep{chen2019topological} or obtain fewer local extrema in the weights of machine learning models \citep{gabrielsson2020topology}, to reduce overfitting.
Similar to topological regularization through (\ref{totalloss1}), they decompose the loss function into an ordinary training loss function (cross-entropy loss, quadratic loss, hinge loss, \ldots) and a topological penalty that is evaluated on the obtained model. 

During topological regularization of data embeddings however, we are concerned with enhancing the topological signal for which the topological loss function $\mathcal{L}_{\mathrm{top}}$ has been designed, rather than decreasing the prominence of spurious topological components.
Since $\mathcal{L}_{\mathrm{top}}$ must be designed for the domain-specific application at hand, this approach is not generic.
However, in the current paper we show that topological priors can have a similar effect as regular regularization in machine learning, in the sense that the learned embedding becomes less vulnerable to noise.
\section{From Simplicial Homology to Topological Optimization}
\label{SEC::background}

The purpose of this section is to present how topological optimization works for point clouds---thus embeddings---in an illustrative manner.
In Section \ref{SUBSEC::filtrations}, we introduce \emph{simplicial complexes} and \emph{filtrations}. 
These are the fundamental combinatorial structures from which \emph{persistent homology} quantifies topological information, as will be explained in Section \ref{SUBSEC::persistence}.
In Section \ref{SUBSEC::alpha}, we introduce two popular types of simplicial complexes and filtrations, namely the \emph{Vietoris-Rips complex and filtration}, and the \emph{weak Alpha complex and filtration}, respectively, the latter of which are more convenient for performing low-dimensional topological optimization (such as in the embedding space) due to their constrained size.
Subsequently, in Section \ref{SUBSEC::toploss} we illustrate how \emph{topological loss functions} defined on the \emph{persistence diagram(s)} of a point cloud (embedding) can be used to conduct \emph{topological optimization}.
Finally, in Section \ref{SUBSEC::compcost}, we discuss the computational cost that is associated with persistent homology and topological optimization.

\begin{figure}
	\centering
	\begin{subfigure}[t]{.475\linewidth}
		\centering
		\includegraphics[width=.63\linewidth]{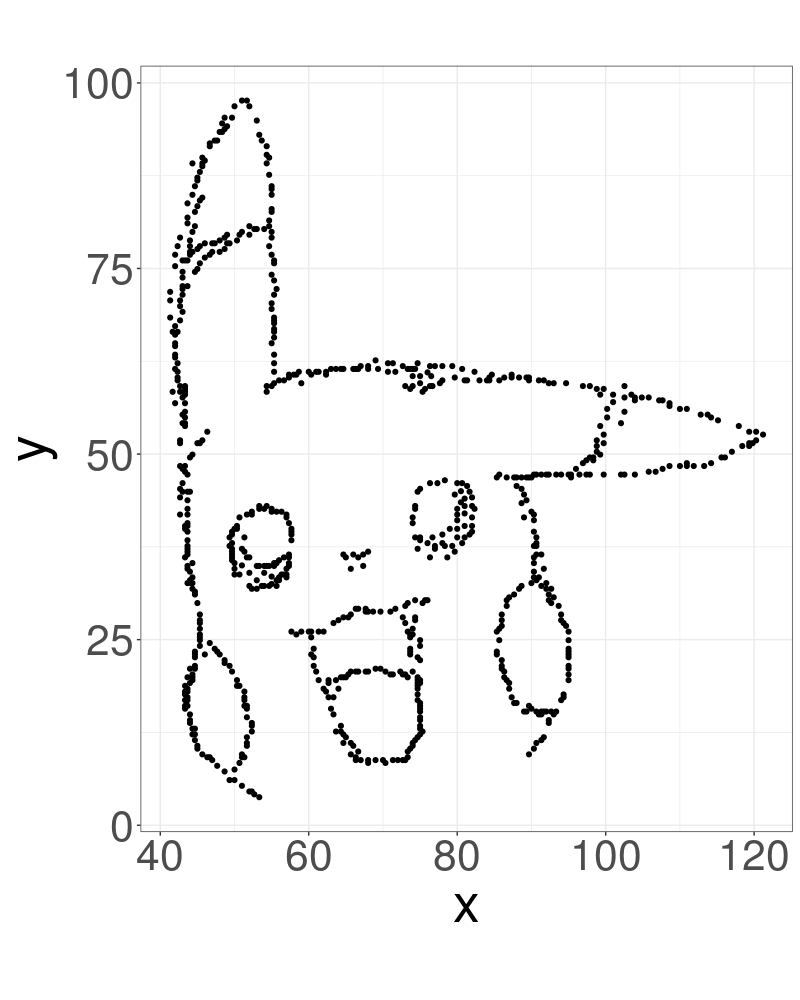}  
		\caption{}
		\label{Pikachu}
	\end{subfigure}
	\hfill
	\begin{subfigure}[t]{.475\linewidth}
		\centering
		\includegraphics[width=.63\linewidth]{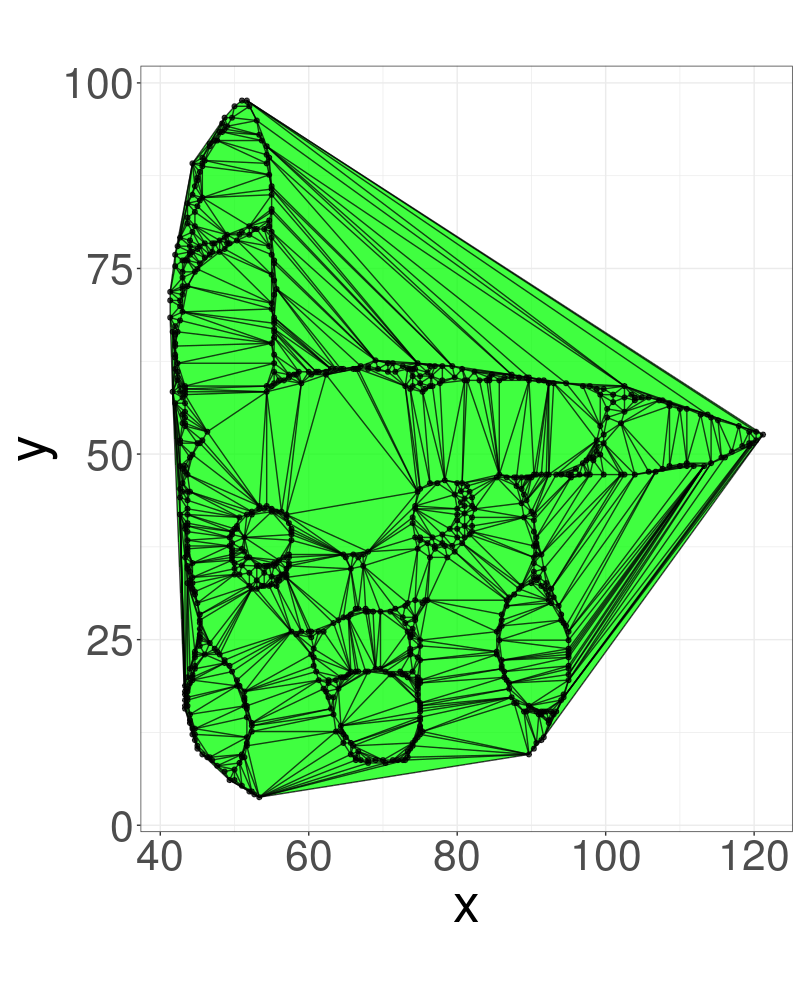}  
		\caption{}
		\label{PikachuAlpha}
	\end{subfigure}
	\vskip\floatsep
	\begin{subfigure}[t]{\linewidth}
		\centering
		\includegraphics[width=.85\linewidth]{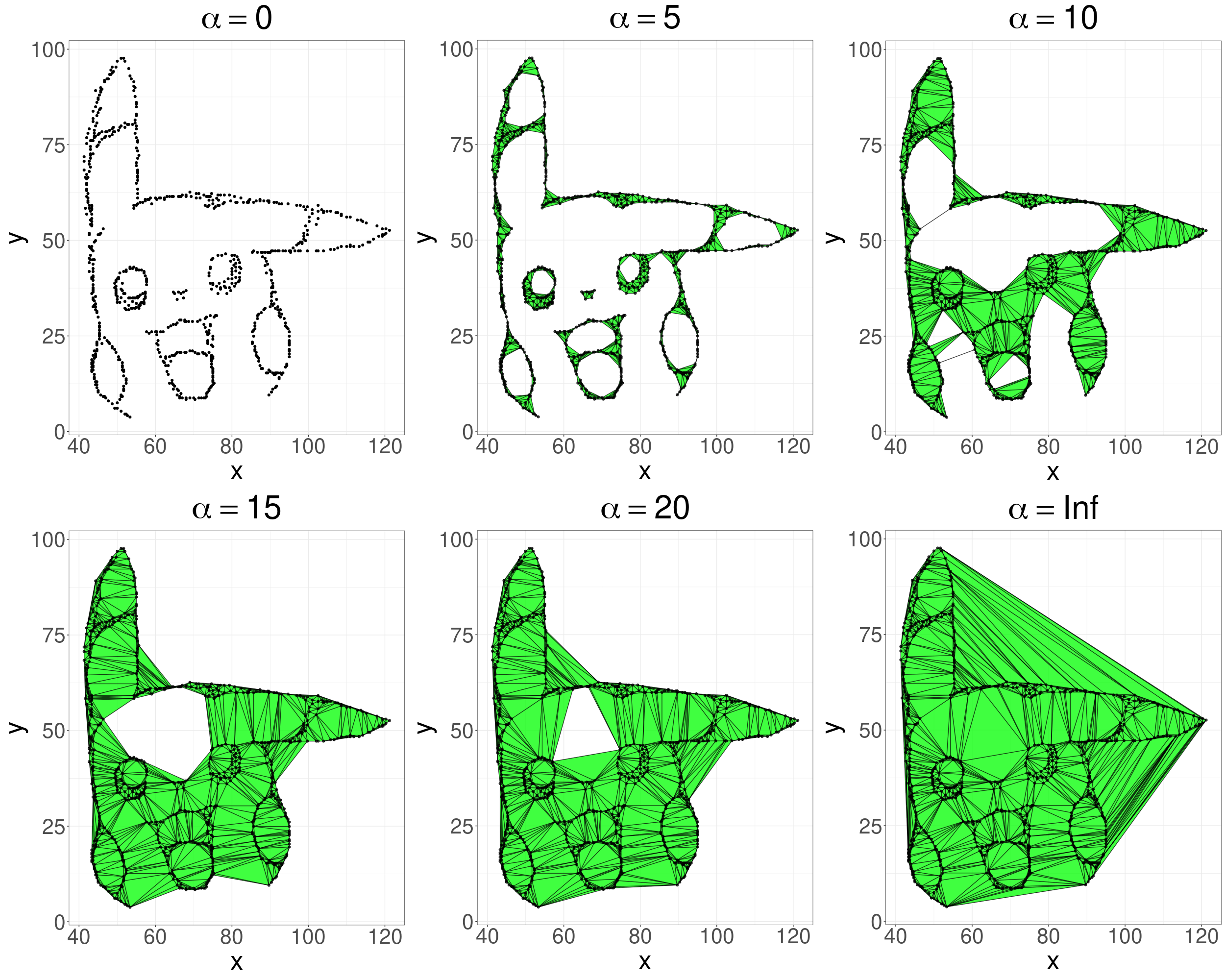}  
		\caption{}
		\label{PikachuFiltration}
	\end{subfigure}
	\caption{(a) A point cloud data set $\mX$ resembling Pikachu.
		(b) The Delaunay triangulation constructed from $\mX$.
		(c) Various simplicial complexes in the weak Alpha filtration---a sequence of subcomplexes from the Delaunay triangulation ordered by inclusion---of $\mX$.
		The time parameter $\alpha=\infty$ is used to informally denote any time parameter above which the weak Alpha complex equals the Delaunay triangulation.}
	\label{PikachuFull}
\end{figure}

Although many of the definitions in this section extend to more general metric spaces, for simplicity, we will restrict to \emph{point clouds}, i.e., finite sets of distinct points in some Euclidean space $\mathbb{R}^k$.
The working example to explain the concepts of persistent homology and topological optimization will be the two-dimensional point cloud data set $\mX$ resembling Pikachu, shown in Figure \ref{Pikachu}.
A simpler example to explain (persistent) simplicial homology computation will also be used (Figure \ref{boundarymap}).
The notation `$\mX$' is used to denote the point cloud data set, while introducing the background material in this section in a broad setting.
Within the context of topological regularization, $\mX$ is the embedding $\mE$.

\subsection{Simplicial Complexes and Filtrations}
\label{SUBSEC::filtrations}

As mentioned above, simplicial complexes and filtrations are the fundamental combinatorial structures from which topological information is quantified through persistent homology (Section \ref{SUBSEC::persistence}).
In general, a simplicial complex can be seen as a generalization of a graph, where apart from nodes or vertices (0-simplices) and edges (1-simplices), it may also include triangles (2-simplices), tetrahedra (3-simplices), and so on.

\begin{definition}
	\label{def::abstractcomplex}
	(Abstract Simplicial Complex).
	An \emph{abstract simplicial complex} $\gK$ is a family of sets that is closed under taking subsets.
	More specifically, the two defining properties are
	\begin{enumerate}
		\item $\gK$ is a set of finite subsets of some given set $\sS$;
		\item if $\Delta'\subseteq\Delta \in \gK$, then $\Delta'\in \gK$.
	\end{enumerate}
	Each element $\Delta$ in $\gK$ is called a \emph{face}, and its \emph{dimension} equals $|\Delta|-1$.
	The \emph{dimension} of $\gK$ is the maximal dimension of all faces in $\gK$.
\end{definition}

Loosely speaking, a \emph{simplicial complex} is an abstract simplicial complex with associated geometry.
A simplicial complex consists of \emph{simplices}, which generalize the notions of triangle and tetrahedron to arbitrary dimensions.

\begin{definition}
	\label{def::simplex}
	(Simplex).
	A \emph{simplex} $\Delta$ is the convex hull of $k+1$ affinely independent points $\vv_0,\ldots,\vv_k\in\mathbb{R}^k$, in which case $\Delta$ is also called a \emph{$k$-simplex}.
	The convex hull of any nonempty subset of $\{\vv_0,\ldots,\vv_k\}$ is called a \emph{face} of $\Delta$.
	Thus, faces are simplices themselves.
\end{definition}

In graph theory, vertices and lines are commonly glued together in a particular way to form a drawing, or thus visualization, of a graph in the plane.
Formally, this is a \emph{geometric realization} of a graph \citep{vandaele2021graph}.
In a similar way, simplices are glued together in a particular way to form a simplicial complex.

\begin{definition}
	\label{def::simpcomplex}
	(Simplicial Complex).
	A simplicial complex $\gK$ is a set of simplices that satisfies
	\begin{enumerate}
		\item every face in $\gK$ is also contained in $\gK$;
		\item the nonempty intersection of two faces $\Delta,\Delta'\in\gK$ is a face of both $\Delta$ and $\Delta'$.
	\end{enumerate}
\end{definition}

Thus, the first defining property in Definition \ref{def::simpcomplex} is similar to the second defining property of an abstract simplicial complex in Definition \ref{def::abstractcomplex}.
The second property in Definition \ref{def::simpcomplex} is similar to how in planar drawing of a graph $G$, two lines representing two edges $e,e'$ of $G$ are not allowed to intersect in a point, unless that point represents a vertex of $G$ that is incident to both $e$ and $e'$.

For the purpose of this paper, it is not an issue for one to mix up the terminology and notations associated to `abstract simplicial complexes' and `simplicial complexes', i.e., to identify simplicial complexes with their discrete counterparts.
The most important thing to be aware of, is that \emph{(persistent) homology} (Section \ref{SUBSEC::persistence}) is concerned with topological properties of the `continuous versions' (geometric realizations) of simplicial complexes, which are computed through their `discrete' (abstract) counterparts.
Compare this to how the connectedness of a graph (as a graph) can be computed through its purely combinatorial and finite representation, but determines the connectedness of any of its geometric realizations (as a topological space containing uncountably many points).

Each separate plot in Figure \ref{PikachuFull} illustrates a simplicial complex.
Furthermore, each simplicial complex in Figure \ref{PikachuFiltration} is a \emph{subcomplex} of the next, that is, all of its simplices (in this case, vertices, edges, and triangles,) are also simplices of the next simplicial complex.
This is the formal defining property of a \emph{filtration}.

\begin{definition}
	\label{def::filtration}
	(Filtration).
	A \emph{filtration} $\gF$ is a sequence of simplicial complexes
	$$
	\gF=(\gK_0\subseteq\gK_1\subseteq\ldots\subseteq\gK_n=\gK).
	$$
\end{definition}

Although filtrations do not have to be of the finite sequence form such as in Definition \ref{def::filtration} \citep{oudot:hal-01247501}, in practice they always are.

\subsection{The Vietoris-Rips and Alpha Filtration}
\label{SUBSEC::alpha}

One of the most popular filtrations on point cloud data for practical applications is the \emph{Vietoris-Rips filtration}, formally defined as follows.

\begin{definition}
	\label{def::VRfilt}
	(Vietors-Rips Complex and Filtration).
	Let $\mX$ be a point cloud and $\epsilon\in\mathbb{R}_{\geq 0}$.
	The \emph{Vietoris-Rips complex} of $\mX$ at time $\epsilon$ is defined as
	$$
	\mathrm{VR}_\epsilon(\mX)\coloneqq\{\Delta\subseteq\mX:\mathrm{diam}(\Delta)\leq\epsilon\},
	$$
	where $\mathrm{diam}(\Delta)\coloneqq\max_{\vx,\vy\in\Delta}\|x-y\|$ is the diameter of $\Delta$.
	The \emph{Vietoris-Rips filtration} is the parameterized sequence of simplicial complexes $(\mathrm{VR}_\epsilon)_{\epsilon\in\mathbb{R}_{\geq 0}}$.
\end{definition}

Thus, the Vietoris-Rips complex at time $\epsilon$ contains a simplex for every subset of points with diameter at most $\epsilon$, and the Vietoris-Rips filtration varies the resulting simplicial complex by increasing the parameter $\epsilon$.
For practical purposes, simplices with dimension larger than $k+1$ are excluded from the filtration whenever $\mX\subseteq\mathbb{R}^k$, since---as will be discussed below---one would be unable to characterize topological information through them.
Even then the Vietoris-Rips filtration may remain challenging to fit into (local) memory.
Therefore, and especially for low-dimensional point clouds, \emph{weak Alpha complexes and filtrations} may be more convenient to use for practical applications.
The simplicial complexes in Figures \ref{PikachuAlpha} and \ref{PikachuFiltration} are examples thereof.
Formally, weak Alpha complexes are \emph{subcomplexes} of a particular simplicial complex termed the \emph{Delaunay triangulation} \citep{delaunay1934sphere}.

\begin{definition}
	\label{def::delaunay}
	(Delaunay Triangulation).
	Let $\mX$ be a point cloud.
	A \emph{triangulation} $\gK$ of $\mX$ is a simplicial complex that covers the convex hull of $\mX$.
	A \emph{Delaunay triangulation} of $\mX$ is a triangulation $\gK$ of $\mX$ such that no point in $\mX$ is inside the circum-hypersphere of any $k$-simplex in $\gK$.
\end{definition}

Intuitively, Delaunay triangulations maximize the minimum angle of all the triangle angles in the triangulation, i.e., they tend to avoid `skinny triangles'.
It is known that a point cloud $\mX$ in $\mathbb{R}^k$ has a unique Delaunay triangulation if $\mX$ is in \emph{general position} \citep{delaunay1934sphere}, that is, the affine hull of $\mX$, i.e., the smallest affine space containing $\mX$, is $k$-dimensional, and no $k+2$ points in $\mX$ lie on the boundary of a ball whose interior does not intersect $\mX$. 
When one deals with finite Euclidean data derived from a continuous distribution, this condition is generally satisfied with probability 1.
Figure \ref{PikachuAlpha} illustrates the Delaunay triangulation of the data $\mX$ in Figure \ref{Pikachu}.

Intuitively, weak Alpha filtrations can now be regarded as Vietoris-Rips filtrations defined on the Delaunay triangulation.

\begin{definition}
	\label{def::WAfilt}
	(Weak Alpha Complex and Filtration).
	Let $\mX$ be a point cloud with Delaunay triangulation $\gK$ and $\alpha\in\mathbb{R}_{\geq 0}$.
	The \emph{weak Alpha complex} of $\mX$ at time $\alpha$ is defined as.
	$$
	\mathrm{WA}_\alpha(\mX)\coloneqq\{\Delta\in\gK:\mathrm{diam}(\Delta)\leq\alpha\},
	$$
	The \emph{weak Alpha filtration} is the parameterized sequence of simplicial complexes $(\mathrm{WA}_\alpha)_{\alpha\in\mathbb{R}_{\geq 0}}$.
\end{definition}

Note that although the filtrations in Definitions~\ref{def::VRfilt} and~\ref{def::WAfilt} are parameterized over an uncountable set, in practice, they are always equivalent to a filtration of finite sequence form such as in Definition~\ref{def::filtration}.
That is because for a point cloud $\mX$, the defined filtrations can only change at finitely many time values.

\subsection{Simplicial Homology}
\label{SUBSEC::homology}

In algebraic and computational topology, \emph{homology} is concerned with studying and quantifying algebraic objects associated to (abstract) simplicial complexes.
It is is defined through the \emph{boundary map} of an abstract simplicial complex.
For simplicity of the definitions in this section, we will assume that we have a fixed ordering of all vertices in the abstract simplicial complex, e.g., as induced through the ordering of data points that define the complex, and that computations are performed over a field (such as modulo 2).

\begin{definition}
	\label{def::boundarymap}
	(Boundary Map).
	Let $\gK$ be an abstract simplicial complex of which the vertices $(v_0, \ldots, v_m)$ are ordered, and $\mathbb{F}$ a field.
	The ordering of vertices in $\gK$ naturally identifies an ordered tuple $(v_{i_0},\ldots,v_{i_k})$, $i_0<\ldots<i_k$, with every simplex $\Delta=\{v_{i_0},\ldots,v_{i_k}\}\in \gK$, and we say that $\Delta$ is \emph{oriented}.
	A \emph{simplicial $k$-chain} is a finite formal sum
	$$
	\lambda_1\Delta_1+\ldots+\lambda_N\Delta_N,
	$$
	where $N\in\mathbb{N}$, $\lambda_i$ are coefficients in $\mathbb{F}$, and $\Delta_i$ are oriented $k$-simplices in $\gK$ (to be interpreted as basis `vectors'), for $1\leq i\leq N$.
	The set of all simplicial $k$-chains forms an $\mathbb{F}$-vector space $\gK_k$ along with the conventional operations for vector spaces.
	The \emph{boundary map} $\partial_k:\Delta_k\rightarrow\Delta_{k-1}$ between the $\mathbb{F}$-vector spaces $\Delta_{k}$ and $\Delta_{k-1}$ is defined by letting for each simplex $\Delta=(v_{i_0},\ldots,v_{i_k})$,
	\begin{equation}
		\label{eq::boundarymap}
		\partial_k(\Delta)\coloneqq\sum_{j=0}^{k}(-1)^j(v_{i_0},\ldots,v_{i_{j-1}},v_{i_{j+1}}\ldots,v_{i_k}).
	\end{equation}
	By convention, $\partial_0\equiv 0$, and $\Delta_{-1}=\{0\}$.
	The boundary map extends linearly to all $k$-chains in $\gK_k$.
	If $C$ is a $k$-chain in $\Delta_k$, we also refer to $\partial_k(C)$ as the \emph{boundary} of $C$.
\end{definition}

It can be shown \citep{Hatcher2002} that $\partial_{k}\circ\partial_{k+1}\equiv 0$.
Thus, the image $\mathrm{Im}(\partial_{k+1})$ is a vector subspace of the kernel $\mathrm{Ker}(\partial_{k})$.
As a result, we can define the \emph{homology module} of an abstract simplicial complex $\gK$.
The term `module' refers to the fact that these structures can be defined over more general algebraic structures, i.e., \emph{rings} \citep{Hatcher2002}.
For fields however, these are just vector spaces.

\begin{definition}
	\label{def::homology}
	(Homology Module).
	Let $\gK$ be an abstract simplicial complex and $\mathbb{F}$ a field, with associated boundary maps $\partial_k$.
	The kernel $\mathrm{ker}(\partial_k)$ is called the \emph{$k$-th cycle module}, and the image $\mathrm{Im}(\partial_k)$ the \emph{$k$-th boundary module}.
	The \emph{$k$-th homology module} of $\gK$ is the $\mathbb{F}$-vector space
	$$
	H_k\coloneqq \mathrm{Ker}(\partial_{k}) \,/\, \mathrm{Im}(\partial_{k+1}).
	$$
\end{definition}

The relation $\mathrm{Im}\,\partial_{k+1}\subseteq\mathrm{ker}\,\partial_k$ has a geometric interpretation.
It implies that every \emph{$(k+1)$-boundary}, which is a $k$-chain that is the image of some $(k+1)$-chain under $\partial_{k+1}$, has zero boundary under $\partial_k$, and hence, is a \emph{$k$-cycle}.
However, the reverse inclusion does not hold in general.
Thus, there may exist $k$-cycles that are not the boundary of any $(k+1)$-chain. 
These correspond to `holes' in the simplicial complex, as illustrated by Figure \ref{boundarymap}.

\begin{figure}
	\centering
	\begin{tikzpicture}
		\node[circle, fill=black, inner sep=0pt, minimum size=5pt, label=above:$x_0$] (x0) at (2, 3.5) {};
		\node[circle, fill=black, inner sep=0pt, minimum size=5pt, label=left:$x_1$] (x1) at (1, 1.75) {};
		\node[circle, fill=black, inner sep=0pt, minimum size=5pt, label=right:$x_2$] (x2) at (3, 1.75) {};
		\node[circle, fill=black, inner sep=0pt, minimum size=5pt, label=below:$x_3$] (x3) at (0, 0) {};
		\node[circle, fill=black, inner sep=0pt, minimum size=5pt, label=below:$x_4$] (x4) at (2, 0) {};
		\node[circle, fill=black, inner sep=0pt, minimum size=5pt, label=below:$x_5$] (x5) at (4, 0) {};
		\draw (x2) -- (x0) -- (x1) -- (x2);
		\draw (x4) -- (x1) -- (x3) -- (x4);
		\draw (x5) -- (x2) -- (x4) -- (x5);
		\begin{pgfonlayer}{background}
			\fill[green] (x0.center) -- (x1.center) -- (x2.center) -- cycle;
			\fill[green] (x1.center) -- (x3.center) -- (x4.center) -- cycle;
			\fill[green] (x2.center) -- (x4.center) -- (x5.center) -- cycle;
		\end{pgfonlayer}
		\draw[->, thick](4, 1.75) -- (6, 1.75) node[pos=0.5, above]{$\partial_2$};
		\node[circle, fill=black, inner sep=0pt, minimum size=5pt, label=above:$x_0$] (y0) at (8, 3.5) {};
		\node[circle, fill=black, inner sep=0pt, minimum size=5pt, label=left:$x_1$] (y1) at (7, 1.75) {};
		\node[circle, fill=black, inner sep=0pt, minimum size=5pt, label=right:$x_2$] (y2) at (9, 1.75) {};
		\node[circle, fill=black, inner sep=0pt, minimum size=5pt, label=below:$x_3$] (y3) at (6, 0) {};
		\node[circle, fill=black, inner sep=0pt, minimum size=5pt, label=below:$x_4$] (y4) at (8, 0) {};
		\node[circle, fill=black, inner sep=0pt, minimum size=5pt, label=below:$x_5$] (y5) at (10, 0) {};
		\draw (y2) -- (y0) -- (y1) -- (y2);
		\draw (y4) -- (y1) -- (y3) -- (y4);
		\draw (y5) -- (y2) -- (y4) -- (y5);
		\draw[red, opacity=0.35, line width=5pt] (y1) -- (y2) -- (y4) -- (y1);
	\end{tikzpicture}
	\caption{Geometric interpretation of the boundary operator. 
		$\partial_{k+1}$ maps a ($k+1$)-chain to a $k$-cycle.
		Some $k$-cycles, such as the one marked in red, are not the boundary of any $(k+1)$-chain.
		These correspond to `holes' in the simplicial complex.}
	\label{boundarymap}
	\vskip\floatsep
	{
		\makeatletter\setlength\BA@colsep{2.25pt}\makeatother
		\begin{blockarray}{ccccccccccccccccccc}
			& $x_0$ & $x_1$ & $x_2$ & $x_3$ & $x_4$ & $x_5$ & $x_{01}$ & $x_{02}$ & $x_{12}$ & $x_{13}$ & $x_{14}$ & $x_{24}$ & $x_{25}$ & $x_{34}$ & $x_{45}$ & $x_{012}$ & $x_{134}$ & $x_{245}$ \\
			\begin{block}{c@{\hspace{10pt}}(cccccccccccccccccc)}
				$x_0$ & 0 & 0 & 0 & 0 & 0 & 0 & \tikzmark{left1}{1} & 1 & 0 & 0 & 0 & 0 & 0 & 0 & 0 & 0 & 0 & 0 \\
				$x_1$ & 0 & 0 & 0 & 0 & 0 & 0 & 1 & 0 & 1 & 1 & 1 & 0 & 0 & 0 & 0 & 0 & 0 & 0 \\
				$x_2$ & 0 & 0 & 0 & 0 & 0 & 0 & 0 & 1 & 1 & 0 & 0 & 1 & 1 & 0 & 0 & 0 & 0 & 0 \\
				$x_3$ & 0 & 0 & 0 & 0 & 0 & 0 & 0 & 0 & 0 & 1 & 0 & 0 & 0 & 1 & 0 & 0 & 0 & 0 \\
				$x_4$ & 0 & 0 & 0 & 0 & 0 & 0 & 0 & 0 & 0 & 0 & 1 & 1 & 0 & 1 & 1 & 0 & 0 & 0 \\
				$x_5$ & 0 & 0 & 0 & 0 & 0 & 0 & 0 & 0 & 0 & 0 & 0 & 0 & 1 & 0 & \tikzmark{right1}{1} & 0 & 0 & 0\\
				$x_{01}$ & 0 & 0 & 0 & 0 & 0 & 0 & 0 & 0 & 0 & 0 & 0 & 0 & 0 & 0 & 0 & \tikzmark{left2}{1} & 0 & 0 \\
				$x_{02}$ & 0 & 0 & 0 & 0 & 0 & 0 & 0 & 0 & 0 & 0 & 0 & 0 & 0 & 0 & 0 & 1 & 0 & 0 \\
				$x_{12}$ & 0 & 0 & 0 & 0 & 0 & 0 & 0 & 0 & 0 & 0 & 0 & 0 & 0 & 0 & 0 & 1 & 0 & 0 \\
				$x_{13}$ & 0 & 0 & 0 & 0 & 0 & 0 & 0 & 0 & 0 & 0 & 0 & 0 & 0 & 0 & 0 & 0 & 1 & 0 \\
				$x_{14}$ & 0 & 0 & 0 & 0 & 0 & 0 & 0 & 0 & 0 & 0 & 0 & 0 & 0 & 0 & 0 & 0 & 1 & 0 \\
				$x_{24}$ & 0 & 0 & 0 & 0 & 0 & 0 & 0 & 0 & 0 & 0 & 0 & 0 & 0 & 0 & 0 & 0 & 0 & 1 \\
				$x_{25}$ & 0 & 0 & 0 & 0 & 0 & 0 & 0 & 0 & 0 & 0 & 0 & 0 & 0 & 0 & 0 & 0 & 0 & 1 \\
				$x_{34}$ & 0 & 0 & 0 & 0 & 0 & 0 & 0 & 0 & 0 & 0 & 0 & 0 & 0 & 0 & 0 & 0 & 1 & 0 \\
				$x_{45}$ & 0 & 0 & 0 & 0 & 0 & 0 & 0 & 0 & 0 & 0 & 0 & 0 & 0 & 0 & 0 & 0 & 0 & \tikzmark{right2}{1}\\
				$x_{012}$ & 0 & 0 & 0 & 0 & 0 & 0 & 0 & 0 & 0 & 0 & 0 & 0 & 0 & 0 & 0 & 0 & 0 & 0 \\
				$x_{134}$ & 0 & 0 & 0 & 0 & 0 & 0 & 0 & 0 & 0 & 0 & 0 & 0 & 0 & 0 & 0 & 0 & 0 & 0 \\
				$x_{245}$ & 0 & 0 & 0 & 0 & 0 & 0 & 0 & 0 & 0 & 0 & 0 & 0 & 0 & 0 & 0 & 0 & 0 & 0 \\
			\end{block}
		\end{blockarray}
		\Highlight[first]{left1}{right1}
		\Highlight[second]{left2}{right2}
		\tikz[overlay,remember picture]{
			\node at (.7, 2.95) (B1) {\textcolor{blue}{$\partial_1\sim B_1$}};
			\node at (.7, -0.65) (B2) {\textcolor{blue}{$\partial_2 \sim B_2$}};
			\draw[->,thick,blue] (first) -- (B1);
			\draw[->,thick,blue] (second) -- (B2);
		}
	}
	\caption{The full boundary matrix $B$ of the left simplicial complex in Figure \ref{boundarymap} over $\mathbb{F}=\mathbb{F}_2$, containing the matrix representations of all boundary operators $\partial_k$.
		The labels $x_i$, $x_{ij}$, and $x_{ijk}$, are short for the simplices $\{x_i\}$, $\{x_i,x_j\}$, and $\{x_i,x_j,x_k\}$, respectively.}
	\label{boundarymatrix}
\end{figure}

The \emph{$k$-th Betti-number} $\beta_k$ quantifies the extend to which the reverse inclusion `$\mathrm{Ker}(\partial_k)\subseteq\mathrm{Im}(\partial_{k+1})$' fails.

\begin{definition}
	\label{def::betti}
	(Betti Number).
	Let $\gK$ be an abstract simplicial complex and $\mathbb{F}$ a field, with associated homology modules $H_k$.
	The \emph{$k$-th Betti number} of $\gK$, denoted $\beta_k$, is the rank of $H_k$, i.e., its number of generators.
\end{definition}

The $k$-th Betti number quantifies the number of $k$-dimensional holes in a (geometric realization of) an abstract simplicial complex.
For example, the simplicial complex in Figure \ref{boundarymap} (Left) has one 0-dimensional hole (one connected component), one 1-dimensional hole (one loop), and no higher-dimensional holes.
Betti numbers can vary over the choice of field however.
For example, for the triangulation $\gK$ of the \emph{Klein bottle}, one may find that $\beta_1(\gK)=2$ over $\mathbb{F}=\mathbb{F}_2$, and  $\beta_1(\gK)=1$ over $\mathbb{F}=\mathbb{F}_p$ for any prime number $p>2$ \citep{maria2014algorithms}.
This is related to the fact that the Klein bottle is not torsion-free.
For more intuitive shapes in low dimensions however, which may be of particular interest during topological regularization, this is rarely an issue.
Without further specification, all (persistent) homology computations in this paper are performed over $\mathbb{F}=\mathbb{F}_2$.

\paragraph{Computing Simplicial Homology}

From a given simplicial complex $\gK$, we can compute the Betti numbers by performing linear operations on the \emph{boundary matrices} of $\gK$.
These are the matrix representations of the boundary operators $\partial_k$.
For $\mathbb{F}=\mathbb{F}_2$, these are just sparse incidence matrices $B_{k}$, where $B_{k_{ij}}=1$ if the $i$-th $(k-1)$-dimensional simplex $\Delta_i$ is a face of the $j$-th $k$-dimensional simplex $\Delta_j$, and $B_{k_{ij}}=0$ otherwise. 
Figure \ref{boundarymatrix} shows the boundary matrices $B_k$ of the left simplicial complex in Figure \ref{boundarymap} collected into the `full' boundary matrix $B$, which contains a row and column for every simplex in the simplicial complex.
Ranks of the homology modules, or thus Betti numbers, can then be derived from the ranks of the matrices $B_k$ through the rank–nullity theorem.

For example for our current example (Figure \ref{boundarymatrix}), over $\mathbb{F}=\mathbb{F}_2$, we have
$$
\begin{cases}
	\mathrm{rank}(B_0)=\mathrm{rank}(\mathrm{Im}(\partial_0))=0 \mbox{ (Definition \ref{def::boundarymap})},\\
	\mathrm{rank}(B_1)=\mathrm{rank}(\mathrm{Im}(\partial_1))=5,\\
	\mathrm{rank}(B_2)=\mathrm{rank}(\mathrm{Im}(\partial_2))=3.
\end{cases}
$$
Note that these ranks were computed in SageMath\citep{sagemath}.
From the rank-nullity theorem, we obtain
$$
\begin{cases}
	\mathrm{rank}(\mathrm{Im}(\partial_0)) + \mathrm{rank}(\mathrm{Ker}(\partial_0)) = 6\iff \mathrm{rank}(\mathrm{Ker}(\partial_0)) = 6,\\
	\mathrm{rank}(\mathrm{Im}(\partial_1)) + \mathrm{rank}(\mathrm{Ker}(\partial_1)) = 9\iff \mathrm{rank}(\mathrm{Ker}(\partial_1)) = 4.
\end{cases}
$$
We thus find that
$$
\begin{cases}
	\beta_0=\mathrm{rank}(H_0) = \mathrm{rank}(\mathrm{Ker}(\partial_0)) - \mathrm{rank}(\mathrm{Im}(\partial_1)) = 6 - 5 = 1,\\
	\beta_1=\mathrm{rank}(H_1) = \mathrm{rank}(\mathrm{Ker}(\partial_1)) - \mathrm{rank}(\mathrm{Im}(\partial_2)) = 4 - 3 = 1.
\end{cases}
$$
Hence, we find that the left simplicial complex in Figure \ref{boundarymap} has $\beta_0=1$ connected component, and $\beta_1=1$ loop, which is consistent with what we observe from the figure.

\subsection{Persistent Homology and Persistence Diagrams}
\label{SUBSEC::persistence}

When given a filtration parameterized by a time parameter $t$, such as $t=\alpha$ in the case of the weak Alpha filtration, one may observe changes in the topological holes in the (abstract) simplicial complex when $t$ increases.
For example, as illustrated in Figure \ref{PikachuFiltration}, different connected components may become connected through edges, or cycles may either appear or disappear.
When a topological hole comes to existence in the filtration, we say that it is \emph{born}.
Vice versa, we say that a topological hole \emph{dies} when it disappears. 
The filtration times at which these events occur are called the \emph{birth time} $b$ and \emph{death time} $d$, respectively.
Simply put, \emph{persistent homology tracks the birth and death times of topological holes across a filtration}. 

\begin{figure}[t]
	\centering
	\includegraphics[width=.725\linewidth]{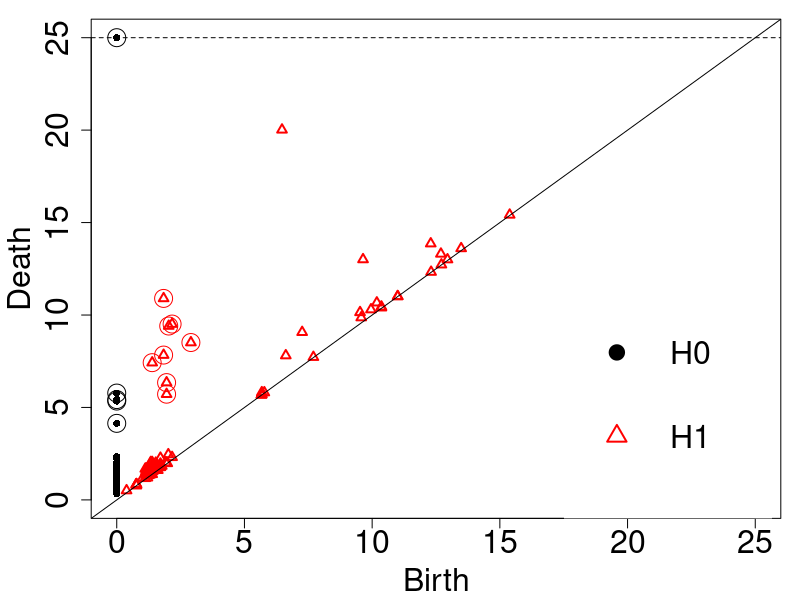}  
	\caption{The persistence diagrams $\gD_0$ and $\gD_1$ of the weak Alpha filtration in Figure \ref{PikachuFiltration} visualized on top of each other. 
		$H_k$ refers to homology dimension $k$. 
		The 13 encircled points represent characteristic shapes of Pikachu's eyes, mouth, nose, cheeks, and ears.}
	\label{PikachuDiagram}
\end{figure}

Persistent homology is commonly quantified through a \emph{persistence diagram} (Figure \ref{PikachuDiagram}).
Formally, a $k$-dimensional persistence diagram is a set $\gD_k\subseteq\mathbb{R}^2$, decomposed as
\begin{align*}
	\underbrace{\left\{(a_i,\infty):1\leq i\leq N\right\}}_{\eqqcolon \gD^{\mathrm{ess}}_k}\cup\underbrace{\left\{(b_j,d_j):1\leq j\leq M\wedge b_j<d_j\right\}}_{\eqqcolon \gD^{\mathrm{reg}}_k}\cup\left\{(x,x):x\in\mathbb{R}\right\},
\end{align*}
where $a_i, b_j, d_j$, $1\leq i\leq N$, $1\leq j\leq M$, correspond to the birth and death times of $k$-dimensional holes across the filtration.
Points $(a_i,\infty)$, are usually displayed on top of the diagram.
They correspond to holes that never die across the filtration, and form the \emph{essential part} of the persistence diagram.
In the case of filtrations defined in Section \ref{SUBSEC::alpha}, one always has $\gD^{\mathrm{ess}}_0=\{(0,\infty)\}$, and $\gD^{\mathrm{ess}}_k=\emptyset$ for $k\geq 1$.
This is because eventually, the simplicial complex in the filtration will consist of one connected component that never dies, and any higher dimensional hole will be `filled in', as illustrated by Figure \ref{PikachuFiltration}.
The points $(t_{b_j},t_{d_j})$ in $\mathcal{D}_k$ with finite coordinates $t_{b_j}<t_{d_j}$ form the \emph{regular part} $D^{\mathrm{reg}}_k$ of the persistence diagram.
Finally, the diagonal is included in a persistence diagram, as to allow for a well-defined distance metric between persistence diagram, termed the \emph{bottleneck distance}.

\begin{definition}
	(bottleneck distance).
	Let $\gD$ and $\gD'$ be two persistence diagrams.
	The \emph{bottleneck distance} between them is defined as
	$$
	d_{\mathrm{b}}\left(\gD, \gD'\right)\coloneqq\inf_{\varphi}\sup_x\|x-\varphi(x)\|_{\infty}\in\mathbb{R}\cup\{\infty\},
	$$
	where $\varphi$ ranges over all bijections from $\gD$ to $\gD'$, and $x$ ranges over all points in $\gD$.
	By convention, we let $\infty-\infty=0$ when calculating the distance between two diagram points.
	Since persistence diagrams include the diagonal by definition, $|\gD|=|\gD'|=|\mathbb{R}|$. 
	Thus, $d_{\mathrm{b}}\left(\gD, \gD'\right)$ is well-defined, as unmatched points in the diagram can always be matched to the diagonal.
\end{definition}

The bottleneck distance commonly justifies the use of persistent homology as a stable method for quantifying topological information in data, i.e., robust to small data permutations \citep{oudot:hal-01247501}.

Figure \ref{PikachuDiagram} shows the 0- and 1-dimensional persistence diagram of the weak Alpha filtration in Figure \ref{PikachuFiltration}. 
Prominent holes during the filtration are those that persist for a long time interval $d-b$, even indefinitely if $d=\infty$.
In the persistence diagrams, these are points that are `highly' elevated above the diagonal.
Points $(b,\infty)$ are commonly plotted on top of the diagram, such as in Figure \ref{PikachuDiagram}.
From the persistence diagrams $\gD_0$ and $\gD_1$ that are plotted on top of each other in Figure \ref{PikachuDiagram}, one may observe 13 prominently elevated points with an early birth-time $b$, that we encircled in Figure \ref{PikachuDiagram}.
The 5 encircled points in $\gD_0$ capture the characteristic connected components of Pikachu's face: its two eyes, mouth, nose, and contour.
The 8 encircled points in $\gD_0$ capture the characteristic loops of Pikachu's face: its two eyes, two cycles in its mouth, its two cheeks, and the two tops of its ears.
While birth-times of connected components naturally equal zero in a weak Alpha filtration, the 8 characteristic loops have an early but nonzero birth-time in the filtration as they occur on a small scale, i.e., they can only be defined by connecting distinct data points that are close to each other (Figure \ref{PikachuFiltration}).
Nevertheless, one may also observe prominently elevated points in $\gD_1$ with larger birth-times.
These correspond to loops that occur on a larger scale and can only be defined by connecting more distant points in the point cloud data.
For example, the loop around Pikachu's forehead, present at times $\alpha=10$ and $\alpha=20$ in Figure \ref{PikachuFiltration}, is the most persisting cycle in the point cloud. 

\paragraph{Computing Persistent Homology}

\begin{algorithm}[b!]
	\caption{The standard algorithm for computing persistence.}
	\label{alg:ph}
	\hspace*{\algorithmicindent} \textbf{Input} boundary matrix $B$ of filtration $\gF$ on simplicial complex $\gK$.    
	\begin{algorithmic}[1]
		\For{$j=1$ to $|\gK|$}
		\While{there exists $i<j$ with $\mathrm{low}(i)=\mathrm{low}(i)$}
		\State add column $i$ to column $j$.
		\EndWhile
		\EndFor
	\end{algorithmic}
\end{algorithm}

Computing persistent homology from a filtration $\gF=(\gK_0\subseteq\gK_1\subseteq\ldots\subseteq\gK_n=\gK)$ is similar to computing regular simplicial homology in that it is performed on the (now full) boundary matrix $B$ of the final simplicial complex $\gK$ (Figure \ref{boundarymatrix}).
However, the rows and columns of $B$ must first be ordered by the time of appearance of their corresponding simplices in the filtration $\gF$.
For example, the boundary matrix $B$ in Figure \ref{boundarymatrix} would be the matrix representation of the filtration
\begin{equation}
	\label{eq:filtrationexample}
	\gF=(\{\{x_0\}\}\subseteq\{\{x_0\},\{x_1\}\}\subseteq\ldots\subseteq\{\{x_0\}, \{x_1\}, \ldots, \{x_2, x_4, x_5\}\}=\gK),
\end{equation}
where $\gK$ is the left simplicial complex in Figure \ref{boundarymap}.
Thus, the simplices that appear in $\gF$ are ordered by size first, then by lexicographical ordering of the indices of their included vertices, as shown in Figure \ref{examplefiltration}.
Note that we did not specify the exact times-of-appearances associated to these simplicial complexes, or resulting from this, the simplices in $\gK$.

\begin{figure}[p]
	\centering
	\begin{tikzpicture}[scale=.3]
		\node[circle, fill=black, inner sep=0pt, minimum size=3pt, label=above:\scriptsize $x_0$] (x0) at (2, 3.5) {};
		\node[circle, fill=white, inner sep=0pt, minimum size=3pt, label=left:{\color{white} \scriptsize $x_1$}] (x1) at (1, 1.75) {};
		\node[circle, fill=white, inner sep=0pt, minimum size=3pt, label=right:{\color{white} \scriptsize $x_2$}] (x2) at (3, 1.75) {};
		\node[circle, fill=white, inner sep=0pt, minimum size=3pt, label=below:{\color{white} \scriptsize $x_3$}] (x3) at (0, 0) {};
		\node[circle, fill=white, inner sep=0pt, minimum size=3pt, label=below:{\color{white} \scriptsize $x_4$}] (x4) at (2, 0) {};
		\node[circle, fill=white, inner sep=0pt, minimum size=3pt, label=below:{\color{white} \scriptsize $x_5$}] (x5) at (4, 0) {};
	\end{tikzpicture}
	\hfill
	\begin{tikzpicture}[scale=.3]
		\node[circle, fill=black, inner sep=0pt, minimum size=3pt, label=above:\scriptsize $x_0$] (x0) at (2, 3.5) {};
		\node[circle, fill=black, inner sep=0pt, minimum size=3pt, label=left:\scriptsize $x_1$] (x1) at (1, 1.75) {};
		\node[circle, fill=white, inner sep=0pt, minimum size=3pt, label=right:{\color{white} \scriptsize $x_2$}] (x2) at (3, 1.75) {};
		\node[circle, fill=white, inner sep=0pt, minimum size=3pt, label=below:{\color{white} \scriptsize $x_3$}] (x3) at (0, 0) {};
		\node[circle, fill=white, inner sep=0pt, minimum size=3pt, label=below:{\color{white} \scriptsize $x_4$}] (x4) at (2, 0) {};
		\node[circle, fill=white, inner sep=0pt, minimum size=3pt, label=below:{\color{white} \scriptsize $x_5$}] (x5) at (4, 0) {};
	\end{tikzpicture}
	\hfill
	\begin{tikzpicture}[scale=.3]
		\node[circle, fill=black, inner sep=0pt, minimum size=3pt, label=above:\scriptsize $x_0$] (x0) at (2, 3.5) {};
		\node[circle, fill=black, inner sep=0pt, minimum size=3pt, label=left:\scriptsize $x_1$] (x1) at (1, 1.75) {};
		\node[circle, fill=black, inner sep=0pt, minimum size=3pt, label=right:\scriptsize $x_2$] (x2) at (3, 1.75) {};
		\node[circle, fill=white, inner sep=0pt, minimum size=3pt, label=below:{\color{white} \scriptsize $x_3$}] (x3) at (0, 0) {};
		\node[circle, fill=white, inner sep=0pt, minimum size=3pt, label=below:{\color{white} \scriptsize $x_4$}] (x4) at (2, 0) {};
		\node[circle, fill=white, inner sep=0pt, minimum size=3pt, label=below:{\color{white} \scriptsize $x_5$}] (x5) at (4, 0) {};
	\end{tikzpicture}
	\hfill
	\begin{tikzpicture}[scale=.3]
		\node[circle, fill=black, inner sep=0pt, minimum size=3pt, label=above:\scriptsize $x_0$] (x0) at (2, 3.5) {};
		\node[circle, fill=black, inner sep=0pt, minimum size=3pt, label=left:\scriptsize $x_1$] (x1) at (1, 1.75) {};
		\node[circle, fill=black, inner sep=0pt, minimum size=3pt, label=right:\scriptsize $x_2$] (x2) at (3, 1.75) {};
		\node[circle, fill=black, inner sep=0pt, minimum size=3pt, label=below:\scriptsize $x_3$] (x3) at (0, 0) {};
		\node[circle, fill=white, inner sep=0pt, minimum size=3pt, label=below:{\color{white} \scriptsize $x_4$}] (x4) at (2, 0) {};
		\node[circle, fill=white, inner sep=0pt, minimum size=3pt, label=below:{\color{white} \scriptsize $x_5$}] (x5) at (4, 0) {};	
	\end{tikzpicture}
	\hfill
	\begin{tikzpicture}[scale=.3]
		\node[circle, fill=black, inner sep=0pt, minimum size=3pt, label=above:\scriptsize $x_0$] (x0) at (2, 3.5) {};
		\node[circle, fill=black, inner sep=0pt, minimum size=3pt, label=left:\scriptsize $x_1$] (x1) at (1, 1.75) {};
		\node[circle, fill=black, inner sep=0pt, minimum size=3pt, label=right:\scriptsize $x_2$] (x2) at (3, 1.75) {};
		\node[circle, fill=black, inner sep=0pt, minimum size=3pt, label=below:\scriptsize $x_3$] (x3) at (0, 0) {};
		\node[circle, fill=black, inner sep=0pt, minimum size=3pt, label=below:\scriptsize $x_4$] (x4) at (2, 0) {};
		\node[circle, fill=white, inner sep=0pt, minimum size=3pt, label=below:{\color{white} \scriptsize $x_5$}] (x5) at (4, 0) {};
	\end{tikzpicture}
	\hfill
	\begin{tikzpicture}[scale=.3]
		\node[circle, fill=black, inner sep=0pt, minimum size=3pt, label=above:\scriptsize $x_0$] (x0) at (2, 3.5) {};
		\node[circle, fill=black, inner sep=0pt, minimum size=3pt, label=left:\scriptsize $x_1$] (x1) at (1, 1.75) {};
		\node[circle, fill=black, inner sep=0pt, minimum size=3pt, label=right:\scriptsize $x_2$] (x2) at (3, 1.75) {};
		\node[circle, fill=black, inner sep=0pt, minimum size=3pt, label=below:\scriptsize $x_3$] (x3) at (0, 0) {};
		\node[circle, fill=black, inner sep=0pt, minimum size=3pt, label=below:\scriptsize $x_4$] (x4) at (2, 0) {};
		\node[circle, fill=black, inner sep=0pt, minimum size=3pt, label=below:\scriptsize $x_5$] (x5) at (4, 0) {};
	\end{tikzpicture}
	\newline
	\begin{tikzpicture}[scale=.3]
		\node[circle, fill=black, inner sep=0pt, minimum size=3pt, label=above:\scriptsize $x_0$] (x0) at (2, 3.5) {};
		\node[circle, fill=black, inner sep=0pt, minimum size=3pt, label=left:\scriptsize $x_1$] (x1) at (1, 1.75) {};
		\node[circle, fill=black, inner sep=0pt, minimum size=3pt, label=right:\scriptsize $x_2$] (x2) at (3, 1.75) {};
		\node[circle, fill=black, inner sep=0pt, minimum size=3pt, label=below:\scriptsize $x_3$] (x3) at (0, 0) {};
		\node[circle, fill=black, inner sep=0pt, minimum size=3pt, label=below:\scriptsize $x_4$] (x4) at (2, 0) {};
		\node[circle, fill=black, inner sep=0pt, minimum size=3pt, label=below:\scriptsize $x_5$] (x5) at (4, 0) {};
		\draw (x0) -- (x1);
	\end{tikzpicture}
	\begin{tikzpicture}[scale=.3]
		\node[circle, fill=black, inner sep=0pt, minimum size=3pt, label=above:\scriptsize $x_0$] (x0) at (2, 3.5) {};
		\node[circle, fill=black, inner sep=0pt, minimum size=3pt, label=left:\scriptsize $x_1$] (x1) at (1, 1.75) {};
		\node[circle, fill=black, inner sep=0pt, minimum size=3pt, label=right:\scriptsize $x_2$] (x2) at (3, 1.75) {};
		\node[circle, fill=black, inner sep=0pt, minimum size=3pt, label=below:\scriptsize $x_3$] (x3) at (0, 0) {};
		\node[circle, fill=black, inner sep=0pt, minimum size=3pt, label=below:\scriptsize $x_4$] (x4) at (2, 0) {};
		\node[circle, fill=black, inner sep=0pt, minimum size=3pt, label=below:\scriptsize $x_5$] (x5) at (4, 0) {};
		\draw (x2) -- (x0) -- (x1);
	\end{tikzpicture}
	\hfill
	\begin{tikzpicture}[scale=.3]
		\node[circle, fill=black, inner sep=0pt, minimum size=3pt, label=above:\scriptsize $x_0$] (x0) at (2, 3.5) {};
		\node[circle, fill=black, inner sep=0pt, minimum size=3pt, label=left:\scriptsize $x_1$] (x1) at (1, 1.75) {};
		\node[circle, fill=black, inner sep=0pt, minimum size=3pt, label=right:\scriptsize $x_2$] (x2) at (3, 1.75) {};
		\node[circle, fill=black, inner sep=0pt, minimum size=3pt, label=below:\scriptsize $x_3$] (x3) at (0, 0) {};
		\node[circle, fill=black, inner sep=0pt, minimum size=3pt, label=below:\scriptsize $x_4$] (x4) at (2, 0) {};
		\node[circle, fill=black, inner sep=0pt, minimum size=3pt, label=below:\scriptsize $x_5$] (x5) at (4, 0) {};
		\draw (x2) -- (x0) -- (x1) -- (x2);
	\end{tikzpicture}
	\hfill
	\begin{tikzpicture}[scale=.3]
		\node[circle, fill=black, inner sep=0pt, minimum size=3pt, label=above:\scriptsize $x_0$] (x0) at (2, 3.5) {};
		\node[circle, fill=black, inner sep=0pt, minimum size=3pt, label=left:\scriptsize $x_1$] (x1) at (1, 1.75) {};
		\node[circle, fill=black, inner sep=0pt, minimum size=3pt, label=right:\scriptsize $x_2$] (x2) at (3, 1.75) {};
		\node[circle, fill=black, inner sep=0pt, minimum size=3pt, label=below:\scriptsize $x_3$] (x3) at (0, 0) {};
		\node[circle, fill=black, inner sep=0pt, minimum size=3pt, label=below:\scriptsize $x_4$] (x4) at (2, 0) {};
		\node[circle, fill=black, inner sep=0pt, minimum size=3pt, label=below:\scriptsize $x_5$] (x5) at (4, 0) {};
		\draw (x2) -- (x0) -- (x1) -- (x2);
		\draw (x1) -- (x3);
	\end{tikzpicture}
	\hfill
	\begin{tikzpicture}[scale=.3]
		\node[circle, fill=black, inner sep=0pt, minimum size=3pt, label=above:\scriptsize $x_0$] (x0) at (2, 3.5) {};
		\node[circle, fill=black, inner sep=0pt, minimum size=3pt, label=left:\scriptsize $x_1$] (x1) at (1, 1.75) {};
		\node[circle, fill=black, inner sep=0pt, minimum size=3pt, label=right:\scriptsize $x_2$] (x2) at (3, 1.75) {};
		\node[circle, fill=black, inner sep=0pt, minimum size=3pt, label=below:\scriptsize $x_3$] (x3) at (0, 0) {};
		\node[circle, fill=black, inner sep=0pt, minimum size=3pt, label=below:\scriptsize $x_4$] (x4) at (2, 0) {};
		\node[circle, fill=black, inner sep=0pt, minimum size=3pt, label=below:\scriptsize $x_5$] (x5) at (4, 0) {};
		\draw (x2) -- (x0) -- (x1) -- (x2);
		\draw (x4) -- (x1) -- (x3);
	\end{tikzpicture}
	\hfill
	\begin{tikzpicture}[scale=.3]
		\node[circle, fill=black, inner sep=0pt, minimum size=3pt, label=above:\scriptsize $x_0$] (x0) at (2, 3.5) {};
		\node[circle, fill=black, inner sep=0pt, minimum size=3pt, label=left:\scriptsize $x_1$] (x1) at (1, 1.75) {};
		\node[circle, fill=black, inner sep=0pt, minimum size=3pt, label=right:\scriptsize $x_2$] (x2) at (3, 1.75) {};
		\node[circle, fill=black, inner sep=0pt, minimum size=3pt, label=below:\scriptsize $x_3$] (x3) at (0, 0) {};
		\node[circle, fill=black, inner sep=0pt, minimum size=3pt, label=below:\scriptsize $x_4$] (x4) at (2, 0) {};
		\node[circle, fill=black, inner sep=0pt, minimum size=3pt, label=below:\scriptsize $x_5$] (x5) at (4, 0) {};
		\draw (x2) -- (x0) -- (x1) -- (x2);
		\draw (x4) -- (x1) -- (x3);
		\draw (x2) -- (x4);
	\end{tikzpicture}
	\newline
	\begin{tikzpicture}[scale=.3]
		\node[circle, fill=black, inner sep=0pt, minimum size=3pt, label=above:\scriptsize $x_0$] (x0) at (2, 3.5) {};
		\node[circle, fill=black, inner sep=0pt, minimum size=3pt, label=left:\scriptsize $x_1$] (x1) at (1, 1.75) {};
		\node[circle, fill=black, inner sep=0pt, minimum size=3pt, label=right:\scriptsize $x_2$] (x2) at (3, 1.75) {};
		\node[circle, fill=black, inner sep=0pt, minimum size=3pt, label=below:\scriptsize $x_3$] (x3) at (0, 0) {};
		\node[circle, fill=black, inner sep=0pt, minimum size=3pt, label=below:\scriptsize $x_4$] (x4) at (2, 0) {};
		\node[circle, fill=black, inner sep=0pt, minimum size=3pt, label=below:\scriptsize $x_5$] (x5) at (4, 0) {};
		\draw (x2) -- (x0) -- (x1) -- (x2);
		\draw (x4) -- (x1) -- (x3);
		\draw (x5) -- (x2) -- (x4);
	\end{tikzpicture}
	\begin{tikzpicture}[scale=.3]
		\node[circle, fill=black, inner sep=0pt, minimum size=3pt, label=above:\scriptsize $x_0$] (x0) at (2, 3.5) {};
		\node[circle, fill=black, inner sep=0pt, minimum size=3pt, label=left:\scriptsize $x_1$] (x1) at (1, 1.75) {};
		\node[circle, fill=black, inner sep=0pt, minimum size=3pt, label=right:\scriptsize $x_2$] (x2) at (3, 1.75) {};
		\node[circle, fill=black, inner sep=0pt, minimum size=3pt, label=below:\scriptsize $x_3$] (x3) at (0, 0) {};
		\node[circle, fill=black, inner sep=0pt, minimum size=3pt, label=below:\scriptsize $x_4$] (x4) at (2, 0) {};
		\node[circle, fill=black, inner sep=0pt, minimum size=3pt, label=below:\scriptsize $x_5$] (x5) at (4, 0) {};
		\draw (x2) -- (x0) -- (x1) -- (x2);
		\draw (x4) -- (x1) -- (x3) -- (x4);
		\draw (x5) -- (x2) -- (x4);
	\end{tikzpicture}
	\hfill
	\begin{tikzpicture}[scale=.3]
		\node[circle, fill=black, inner sep=0pt, minimum size=3pt, label=above:\scriptsize $x_0$] (x0) at (2, 3.5) {};
		\node[circle, fill=black, inner sep=0pt, minimum size=3pt, label=left:\scriptsize $x_1$] (x1) at (1, 1.75) {};
		\node[circle, fill=black, inner sep=0pt, minimum size=3pt, label=right:\scriptsize $x_2$] (x2) at (3, 1.75) {};
		\node[circle, fill=black, inner sep=0pt, minimum size=3pt, label=below:\scriptsize $x_3$] (x3) at (0, 0) {};
		\node[circle, fill=black, inner sep=0pt, minimum size=3pt, label=below:\scriptsize $x_4$] (x4) at (2, 0) {};
		\node[circle, fill=black, inner sep=0pt, minimum size=3pt, label=below:\scriptsize $x_5$] (x5) at (4, 0) {};
		\draw (x2) -- (x0) -- (x1) -- (x2);
		\draw (x4) -- (x1) -- (x3) -- (x4);
		\draw (x5) -- (x2) -- (x4) -- (x5);
	\end{tikzpicture}
	\hfill
	\begin{tikzpicture}[scale=.3]
		\node[circle, fill=black, inner sep=0pt, minimum size=3pt, label=above:\scriptsize $x_0$] (x0) at (2, 3.5) {};
		\node[circle, fill=black, inner sep=0pt, minimum size=3pt, label=left:\scriptsize $x_1$] (x1) at (1, 1.75) {};
		\node[circle, fill=black, inner sep=0pt, minimum size=3pt, label=right:\scriptsize $x_2$] (x2) at (3, 1.75) {};
		\node[circle, fill=black, inner sep=0pt, minimum size=3pt, label=below:\scriptsize $x_3$] (x3) at (0, 0) {};
		\node[circle, fill=black, inner sep=0pt, minimum size=3pt, label=below:\scriptsize $x_4$] (x4) at (2, 0) {};
		\node[circle, fill=black, inner sep=0pt, minimum size=3pt, label=below:\scriptsize $x_5$] (x5) at (4, 0) {};
		\draw (x2) -- (x0) -- (x1) -- (x2);
		\draw (x4) -- (x1) -- (x3) -- (x4);
		\draw (x5) -- (x2) -- (x4) -- (x5);
		\begin{pgfonlayer}{background}
			\fill[green] (x0.center) -- (x1.center) -- (x2.center) -- cycle;
		\end{pgfonlayer}
	\end{tikzpicture}
	\hfill
	\begin{tikzpicture}[scale=.3]
		\node[circle, fill=black, inner sep=0pt, minimum size=3pt, label=above:\scriptsize $x_0$] (x0) at (2, 3.5) {};
		\node[circle, fill=black, inner sep=0pt, minimum size=3pt, label=left:\scriptsize $x_1$] (x1) at (1, 1.75) {};
		\node[circle, fill=black, inner sep=0pt, minimum size=3pt, label=right:\scriptsize $x_2$] (x2) at (3, 1.75) {};
		\node[circle, fill=black, inner sep=0pt, minimum size=3pt, label=below:\scriptsize $x_3$] (x3) at (0, 0) {};
		\node[circle, fill=black, inner sep=0pt, minimum size=3pt, label=below:\scriptsize $x_4$] (x4) at (2, 0) {};
		\node[circle, fill=black, inner sep=0pt, minimum size=3pt, label=below:\scriptsize $x_5$] (x5) at (4, 0) {};
		\draw (x2) -- (x0) -- (x1) -- (x2);
		\draw (x4) -- (x1) -- (x3) -- (x4);
		\draw (x5) -- (x2) -- (x4) -- (x5);
		\begin{pgfonlayer}{background}
			\fill[green] (x0.center) -- (x1.center) -- (x2.center) -- cycle;
			\fill[green] (x1.center) -- (x3.center) -- (x4.center) -- cycle;
		\end{pgfonlayer}
	\end{tikzpicture}
	\hfill
	\begin{tikzpicture}[scale=.3]
		\node[circle, fill=black, inner sep=0pt, minimum size=3pt, label=above:\scriptsize $x_0$] (x0) at (2, 3.5) {};
		\node[circle, fill=black, inner sep=0pt, minimum size=3pt, label=left:\scriptsize $x_1$] (x1) at (1, 1.75) {};
		\node[circle, fill=black, inner sep=0pt, minimum size=3pt, label=right:\scriptsize $x_2$] (x2) at (3, 1.75) {};
		\node[circle, fill=black, inner sep=0pt, minimum size=3pt, label=below:\scriptsize $x_3$] (x3) at (0, 0) {};
		\node[circle, fill=black, inner sep=0pt, minimum size=3pt, label=below:\scriptsize $x_4$] (x4) at (2, 0) {};
		\node[circle, fill=black, inner sep=0pt, minimum size=3pt, label=below:\scriptsize $x_5$] (x5) at (4, 0) {};
		\draw (x2) -- (x0) -- (x1) -- (x2);
		\draw (x4) -- (x1) -- (x3) -- (x4);
		\draw (x5) -- (x2) -- (x4) -- (x5);
		\begin{pgfonlayer}{background}
			\fill[green] (x0.center) -- (x1.center) -- (x2.center) -- cycle;
			\fill[green] (x1.center) -- (x3.center) -- (x4.center) -- cycle;
			\fill[green] (x2.center) -- (x4.center) -- (x5.center) -- cycle;
		\end{pgfonlayer}
	\end{tikzpicture}
	\caption{The filtration on the left simplicial complex in Figure \ref{boundarymap} for which the matrix $B$ in Figure \ref{boundarymatrix} is the boundary matrix.}
	\label{examplefiltration}
	\vskip\floatsep
	{
		\makeatletter\setlength\BA@colsep{2.25pt}\makeatother
		\begin{blockarray}{ccccccccccccccccccc}
			& $x_0$ & $x_1$ & $x_2$ & $x_3$ & $x_4$ & $x_5$ & $x_{01}$ & $x_{02}$ & $x_{12}$ & $x_{13}$ & $x_{14}$ & $x_{24}$ & $x_{25}$ & $x_{34}$ & $x_{45}$ & $x_{012}$ & $x_{134}$ & $x_{245}$ \\
			\begin{block}{c@{\hspace{10pt}}(cccccccccccccccccc)}
				$x_0$ & 0 & 0 & 0 & 0 & 0 & 0 & 1 & 1 & 0 & 0 & 0 & 0 & 0 & 0 & 0 & 0 & 0 & 0 \\
				$x_1$ & 0 & 0 & 0 & 0 & 0 & 0 & 1 & 0 & 0 & 1 & 1 & 0 & 0 & 0 & 0 & 0 & 0 & 0 \\
				$x_2$ & 0 & 0 & 0 & 0 & 0 & 0 & 0 & 1 & 0 & 0 & 0 & 0 & 1 & 0 & 0 & 0 & 0 & 0 \\
				$x_3$ & 0 & 0 & 0 & 0 & 0 & 0 & 0 & 0 & 0 & 1 & 0 & 0 & 0 & 0 & 0 & 0 & 0 & 0 \\
				$x_4$ & 0 & 0 & 0 & 0 & 0 & 0 & 0 & 0 & 0 & 0 & 1 & 0 & 0 & 0 & 0 & 0 & 0 & 0 \\
				$x_5$ & 0 & 0 & 0 & 0 & 0 & 0 & 0 & 0 & 0 & 0 & 0 & 0 & 1 & 0 & 0 & 0 & 0 & 0\\
				$x_{01}$ & 0 & 0 & 0 & 0 & 0 & 0 & 0 & 0 & 0 & 0 & 0 & 0 & 0 & 0 & 0 & 1 & 0 & 0 \\
				$x_{02}$ & 0 & 0 & 0 & 0 & 0 & 0 & 0 & 0 & 0 & 0 & 0 & 0 & 0 & 0 & 0 & 1 & 0 & 0 \\
				$x_{12}$ & 0 & 0 & 0 & 0 & 0 & 0 & 0 & 0 & 0 & 0 & 0 & 0 & 0 & 0 & 0 & 1 & 0 & 0 \\
				$x_{13}$ & 0 & 0 & 0 & 0 & 0 & 0 & 0 & 0 & 0 & 0 & 0 & 0 & 0 & 0 & 0 & 0 & 1 & 0 \\
				$x_{14}$ & 0 & 0 & 0 & 0 & 0 & 0 & 0 & 0 & 0 & 0 & 0 & 0 & 0 & 0 & 0 & 0 & 1 & 0 \\
				$x_{24}$ & 0 & 0 & 0 & 0 & 0 & 0 & 0 & 0 & 0 & 0 & 0 & 0 & 0 & 0 & 0 & 0 & 0 & 1 \\
				$x_{25}$ & 0 & 0 & 0 & 0 & 0 & 0 & 0 & 0 & 0 & 0 & 0 & 0 & 0 & 0 & 0 & 0 & 0 & 1 \\
				$x_{34}$ & 0 & 0 & 0 & 0 & 0 & 0 & 0 & 0 & 0 & 0 & 0 & 0 & 0 & 0 & 0 & 0 & 1 & 0 \\
				$x_{45}$ & 0 & 0 & 0 & 0 & 0 & 0 & 0 & 0 & 0 & 0 & 0 & 0 & 0 & 0 & 0 & 0 & 0 & 1\\
				$x_{012}$ & 0 & 0 & 0 & 0 & 0 & 0 & 0 & 0 & 0 & 0 & 0 & 0 & 0 & 0 & 0 & 0 & 0 & 0 \\
				$x_{134}$ & 0 & 0 & 0 & 0 & 0 & 0 & 0 & 0 & 0 & 0 & 0 & 0 & 0 & 0 & 0 & 0 & 0 & 0 \\
				$x_{245}$ & 0 & 0 & 0 & 0 & 0 & 0 & 0 & 0 & 0 & 0 & 0 & 0 & 0 & 0 & 0 & 0 & 0 & 0 \\
			\end{block}
		\end{blockarray}
	}
	\caption{The reduced boundary matrix $B$ in Figure \ref{boundarymatrix} through Algorithm \ref{alg:ph}.}
	\label{boundaryreduced}
\end{figure}

For the $j$-th column of $B$, we define $\mathrm{low}(j)$ as the largest index $i$ for which $B_{ij}\neq 0$.
The \emph{standard algorithm} for computing persistence, also sometimes called the \emph{column algorithm} \citep{Otter2017}, reduces the boundary matrix $B$ in the chosen field (e.g., modulo 2) through Algorithm \ref{alg:ph}.
One can then read off the the birth-death pairs from the reduced matrix as follows \citep{Otter2017}.

\begin{itemize}
	\item If $\mathrm{low}(j) = i$, then the simplex $\Delta_j$ is paired with $\Delta_i$, and the entrance of $\Delta_i$ in the filtration causes the birth of a topological hole that dies with the entrance of $\Delta_j$.
	\item If $\mathrm{low}(j)$ is undefined, i.e., the $j$-th column of the reduced boundary matrix only contains zeroes, then the the entrance of the simplex $\Delta_j$ in the filtration causes the
	birth of a topological hole.
	If there exists some $k$ such that $\mathrm{low}(k) = j$, then the simplex $\Delta_j$ is paired with $\Delta_k$, whose entrance in the filtration causes the death of the topological hole that was born with $\Delta_j$.
	If no such $k$ exists, $\Delta_j$ is unpaired, and the topological hole born with it persists indefinitely.
\end{itemize}

Figure \ref{boundaryreduced} shows the reduced form of the matrix in $B$ corresponding to the filtration in (\ref{eq:filtrationexample}).
As an example, we read off some of the birth-death pairs.

\begin{itemize}
	\item $\mathrm{low}(x_i)$ is undefined for every $i=1,\ldots,5$: every vertex results in the birth of a connected component.
	We find that only for $x_0$, there is no column $x_{ij}$ of the reduced matrix for which $\mathrm{low}(x_{ij})=x_0$. 
	Thus, only the connect component born through  $x_0$ persists indefinitely.
	\item $\mathrm{low}(x_{01})=x_1$: with the entrance of the edge $\{x_0,x_1\}$, the connected component that was born with $\{x_1\}$ dies.
	\item $\mathrm{low}(x_{12})$ is undefined, however $\mathrm{low}(x_{012})=x_{12}$: the entrance of the edge $\{x_1,x_2\}$ resulted in the birth of a cycle that died with the entrance of the simplex $\{x_0, x_1, x_2\}$.
	\item $\mathrm{low}(x_{24})$ is undefined, and there is no column $x_{ijk}$ of the reduced matrix for which $\mathrm{low}(x_{ijk})=x_{24}$: the entrance of the edge $\{x_2,x_4\}$ resulted in the birth of a cycle that persists indefinitely.
\end{itemize}

Note that all of these observations can be made in Figure \ref{examplefiltration} as well.
Furthermore, observe that the topological holes that persists indefinitely correspond to the regular simplicial homology of the final simplicial complex $\gK$ in the filtration, as computed at the end of Section \ref{SUBSEC::homology}.

By mapping the paired simplices to their times of appearance---which are available in practical applications---one obtains birth-death tuples $(b, d)$, or thus persistence diagrams.
Note that the assignment of a birth-death tuple to a particular dimension of persistence diagram can be done based on the dimensions of the paired simplices.
For example, one knows that the entrance of an edge can only result in the birth of a cycle, or the death of a connected component.
\emph{What is most important to realize from this example for topological optimization, which we discuss below, is that the computation of persistent homology and persistence diagrams allows one to trace a birth-death tuple $(b, d)$ corresponding to a particular topological hole, back to the two simplices in the filtration that resulted in the birth and death of that hole}.

\subsection{Topological Loss Functions and Topological Optimization}
\label{SUBSEC::toploss}

Persistent homology allows one to quantify all from the finest to coarsest topological information in data.
Therefore, persistence diagrams are often regarded as \emph{topological signatures} of data, which can be transformed into (topological) features to be incorporated into various machine learning models.
While methods that learn from persistent homology are now both well developed and diverse \citep{pun2018persistent}, optimizing the data representation for the persistent homology thereof only gained recent attention \citep{gabrielsson2020topology, solomon2021fast, carriere2021optimizing}.
As presented in Sections \ref{SUBSEC::homology} \& \ref{SUBSEC::persistence}, persistent homology has a rather abstract mathematical foundation within algebraic topology, and its computation is inherently combinatorial.
This complicates working with usual derivatives for optimization.
To accommodate this, topological optimization makes use of Clarke subderivatives \citep{clarke1990optimization}, whose applicability to persistence builds on arguments from o-minimal geometry \citep{van1998tame, carriere2021optimizing}.
Fortunately, thanks to the recent work of \citep{gabrielsson2020topology} and \citep{carriere2021optimizing}, powerful tools for topological optimization have been developed for software libraries such as PyTorch and TensorFlow, allowing their usage without deeper knowledge about these subjects. 

In this section, \emph{we will present topological optimization for point clouds} in particular.
Note that topological optimization can be applied to other input data structures, for example for optimizing the pixel values in images for denoising \citep{gabrielsson2020topology}.
However, point cloud embeddings are the structures we seek to topologically optimize in this paper.
In Section \ref{SUBSUBSEC::pointcloudopt}, we introduce the basic form of the topological loss functions that we use in this paper following the approach by \citep{gabrielsson2020topology}, and provide some first examples.
In Section \ref{SUBSUBSEC::techback}, we provide an overview of the technical background behind topological optimization of point clouds, where we discuss its dependence on subgradient methods.

\subsubsection{Topological Optimization of Point Clouds: an Example}
\label{SUBSUBSEC::pointcloudopt}

Topological optimization of a point cloud $\mX$ optimizes the placement of points in $\mX$ with respect to the topological information summarized by one or multiple persistence diagrams $\gD$ of $\mX$.
In this paper, these will be obtained from the weak Alpha filtration of $\mX$ (Figures \ref{PikachuAlpha} \& \ref{PikachuDiagram}).
We will use the approach by \citep{gabrielsson2020topology}, where (birth, death)-tuples $(b_1, d_1),(b_2, d_2),\ldots, (b_{|\gD|}, d_{|\gD|})$ in $\gD$ are first ordered by decreasing persistence $d_k-b_k$.
In case of point clouds, one and only one topological hole, i.e., a connected component born at time $\alpha=0$, will always persist indefinitely. 
Other gaps and holes will eventually be filled (Figures \ref{PikachuAlpha} \& \ref{PikachuDiagram}).
Thus, we only optimize for the regular part of the persistence diagram in this paper, i.e., for the diagram points with finite coordinates.
This is done through a \emph{topological loss function}, which for a choice of $i\leq j$, a dimension $k$ of topological hole, and a function $g:\mathbb{R}^2\rightarrow\mathbb{R}$, is defined as
\begin{equation}
	\label{eq:topo_loss}
	\mathcal{L}_{\mathrm{top}}(\gD_k)\coloneqq\sum_{l=i, d_l < \infty}^jg(b_l, d_l), \hspace{2em}\mbox{ where } d_1-b_1\geq d_2-b_2\geq\ldots,
\end{equation}
with $(b_l, d_l)\in \gD_k$, $l=1,\ldots |\gD_k|$. 
By first ordering the points in $\gD_k$ by decreasing persistence, (\ref{eq:topo_loss}) is a \emph{function of persistence} \citep{carriere2021optimizing}, meaning that it is invariant
to permutations of the points of the persistence diagram $\gD_k$.

\begin{figure}[p]
	\centering
	\includegraphics[width=\linewidth]{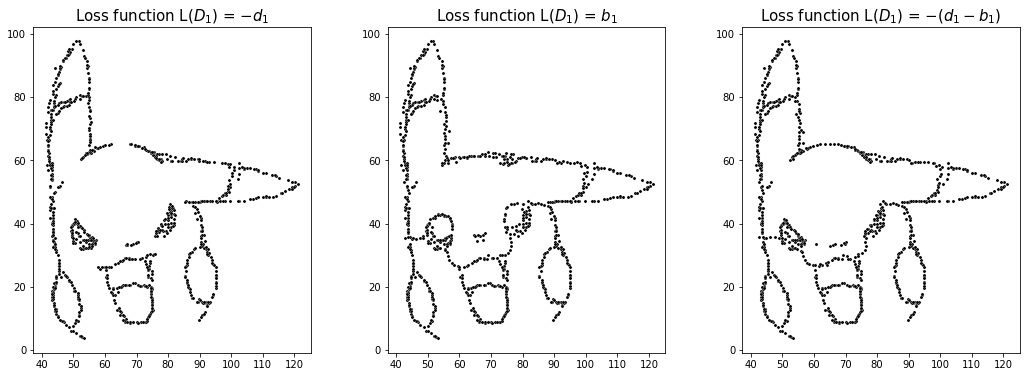}  
	\caption{The point cloud $\mX$ optimized for various topological loss functions, which are all designed to increase the persistence $d_1-b_1$ of the most prominent cycle that corresponds to the contour of Pikachu's forehead.}
	\label{PikaOpt}
	\vskip\floatsep
	\includegraphics[width=0.45\linewidth]{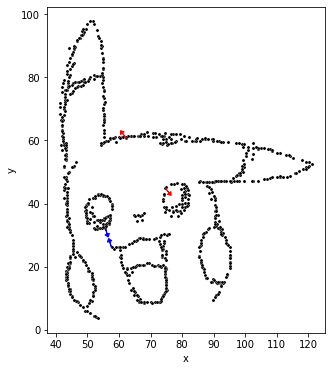}  
	\caption{The negative gradients of the topological loss function $\mathcal{L}_{\mathrm{top}}(\gD_1)=b_1-d_1$ show the direction of point movements during the first topological optimization iteration of $\mX$. The gradients are nonzero only for four points. The connection of two of these during the weak Alpha filtration causes the birth of the cycle (blue arrows). These points move towards each other to decrease the birth-time of the cycle.
	The connection of the other two points during the weak Alpha filtration causes the death of the cycle (red arrows). These points move away from each other to increase the death-time of the cycle.}
	\label{PikaGrad}
\end{figure}

For example, consider the 1st-dimensional persistence diagram $\gD_1$ of the point cloud representation $\mX$ of Pikachu obtained from the weak Alpha filtration (Figures \ref{PikachuAlpha} \& \ref{PikachuDiagram}).
As discussed in Section \ref{SUBSEC::persistence}, the most persisting cycle---represented by the point $(b_1, d_1)\in \gD_1$ according to the ordering in (\ref{eq:topo_loss})---captures the contour of Pikachu's forehead.
The following three functions are then examples of topological loss functions that might be used to increase the persistence $d_1-b_1$ of this cycle, with $i=j=1$ in (\ref{eq:topo_loss}).
\begin{compactitem}
	\item $\mathcal{L}_{{\mathrm{top}}_1}(\gD_k)=-d_1$,
	\item $\mathcal{L}_{{\mathrm{top}}_2}(\gD_k)=b_1$,
	\item $\mathcal{L}_{{\mathrm{top}}_3}(\gD_k)=-(d_1-b_1)$.
\end{compactitem}
Figure \ref{PikaOpt} shows the result of optimizing for each of these loss functions.

When optimizing for $\mathcal{L}_{{\mathrm{top}}_1}$, we see that the cycle enlarges.
However, gaps in the cycle are neglected or even introduced (Figure \ref{PikaOpt}, Left).
Conversely, optimizing for $\mathcal{L}_{{\mathrm{top}}_2}$ closes such gaps, and ensures that the cycle occurs on a smaller scale in the data, or more formally, for earlier filtration times (Figure \ref{PikaOpt}, Middle), but does not enlarge the cycle.
Naturally, both effects are achieved when optimizing for $\mathcal{L}_{{\mathrm{top}}_3}$ (Figure \ref{PikaOpt}, Right).

Figure \ref{PikaGrad} shows the direction of the negative gradients, or thus the direction of point movements, for the topological loss function $\mathcal{L}_{{\mathrm{top}}_3}(\gD_k)$, for the first topological optimization iteration of $\mX$.
We observe that the gradients are nonzero for only four points in $\mX$.
These correspond to either the two points that cause the birth of the most persisting cycle (blue arrows in Figure \ref{PikaGrad}) or the two points that cause the death of this cycle (red arrows in Figure \ref{PikaGrad}).
This is because the point $(b_1, d_1)\in \gD_1$ is traced back to the paired simplices whose entrance cause the birth and death of this cycle (Section \ref{SUBSEC::persistence}).
The birth of the cycle at time $b_1$ is caused when the points marked with blue gradients in Figure \ref{PikaGrad} are connected by an edge.
The death of the cycle at time $d_1$ is caused when the points marked with red gradients in Figure \ref{PikaGrad} are connected by an edge, whose entrance forms a triangle with a third unmarked point that closes the cycle.
Since the filtration times of these edges, i.e., the times of their first occurrence during the weak Alpha filtration, are exactly the distances between their endpoints, the gradients are in opposite direction as to either decrease the birth time $b_1$ or increase the death time $d_1$, and thus increase the overall persistence $d_1-b_1$.

\subsubsection{Technical Background of Topological Optimization of Point Clouds}
\label{SUBSUBSEC::techback}

Topological optimization rests on (stochastic) subgradient methods for optimizing the data representation with respect to its topological properties summarized by its persistence diagrams.
To explain this, it is easiest to start from topological optimization based on the Vietoris-Rips filtration (Definition \ref{def::VRfilt}).

For a given point cloud $\mX\subseteq\mathbb{R}^d$, the Vietoris-Rips filtration is a filtration on the simplicial complex $\gK$ that contains a simplex for every subset of points in $\mX$, possibly constrained by some dimension.
If now $\mX$ consists of $n$ points, we can regard the Vietoris-Rips filtration construction as a mapping $\mathrm{VR}:A=\left(\mathbb{R}^d\right)^n\rightarrow\mathbb{R}^{|\gK|}$,
where an element in $A$ equals a point cloud of $n$ points in $\mathbb{R}^d$, which gets mapped by $\mathrm{VR}$ onto a (Vietoris-Rips) filtration which can be written as an assignment of a real value (filtration time) to every simplex in $\gK$.

Persistent homology now pairs simplices that cause the birth and death of the same topological hole during a filtration of $\gK$.
For simplicity of notation, we consider one full persistence diagram $\gD$ that is the union of the persistence diagrams in all dimensions. 
Let $p$ be the number of paired simplices and $q$ the number of unpaired simplices (Section \ref{SUBSEC::persistence}) during the filtration.
Then $|\gK|=2p+q$.
Furthermore, in particular for the Vietoris-Rips filtration, the numbers $p$ and $q$ will depend only on the data size $n$ and the dimension of $\gK$.
Choosing the lexicographical order on $\mathbb{R}\times(\mathbb{R}\cup\{\infty\})$, we can regard the persistence diagram $\gD$ as a vector in $\mathbb{R}^{2p+q}=\mathbb{R}^{|\gK|}$.
Thus, we view the persistent homology operation as a mapping $\mathrm{Pers}:\mathbb{R}^{|\gK|}\rightarrow\mathbb{R}^{|\gK|}$.
Since birth and death times are necessarily filtration values of simplices, the persistent homology operation $\mathrm{Pers}$ is thus simply a permutation of the coordinates of the input filtration, seen as a vector in $\mathbb{R}^{|\gK|}$.

Having defined a topological loss function $\mathcal{L}_{\mathrm{top}}$ as in (\ref{eq:topo_loss}), the goal of topological optimization is thus to minimize the function
\begin{equation}
	\label{VRfunction}
	\mathcal{L}_{\mathrm{top}}\circ\mathrm{Pers}\circ\mathrm{VR}:A=\left(\mathbb{R}^d\right)^n\rightarrow\mathbb{R},
\end{equation}
with respect to the point cloud $\mX\in A$.

\emph{O-minimal geometry} now provides a well-suited setting for describing parametrized families of filtrations (here by the point cloud $\mX\in A$) and to exhibit interesting differentiability properties of their composition with the persistence mapping and a topological loss function \citep{carriere2021optimizing}.

\begin{definition}
	\label{def::ominimal}
	(O-minimal Structure \citep{carriere2021optimizing}).
	An \emph{o-minimal structure} on the field of real numbers $\mathbb{R}$ is a collection $(\sS_n)_{n\in\mathbb{N}}$, where each $\sS_n$ is a set of subsets of $\mathbb{R}^n$ such that:
	\begin{enumerate}
		\item $\sS_1$ is exactly the collection of finite unions of points and intervals;
		\item all algebraic subsets (0-level sets of polynomials) of $\mathbb{R}^n$ are in $\sS_n$;
		\item $\sS_n$ is a Boolean subalgebra of $\mathbb{R}^n$ for any $n \in \mathbb{N}$;
		\item if $A \in \sS_n$ and $B \in \sS_m$, then $A\times B \in \sS_{n+m}$;
		\item if $\pi: \mathbb{R}^{n+1} \rightarrow \mathbb{R}^n$ is the linear projection onto the first $n$ coordinates and $A \in \sS_{n+1}$, then $\pi(A) \in \sS_n$.
	\end{enumerate}
	An element $A \in \sS_n$ for some $n \in \mathbb{N}$ is called a definable set in the o-minimal structure. 
	For a definable set $A \subseteq \mathbb{R}^n$, a map $f:A \rightarrow \mathbb{R}^m$ is said to be definable if its graph is a definable set in $\mathbb{R}^{n+m}$.
\end{definition}

Definable sets are stable under various geometric operations \citep{coste2000introduction, carriere2021optimizing}.
While the exact details behind this are beyond the scope of this paper, \citep{carriere2021optimizing} showed that if a function of persistence $\mathcal{L}_{\mathrm{top}}$ is locally Lipschitz and definable in an o-minimal structure, then $\mathcal{L}_{\mathrm{top}}\circ\mathrm{Pers}$ is also locally Lipschitz and the composition (\ref{VRfunction}) is definable.
As a consequence, the composition (\ref{VRfunction}) has a well-defined Clarke subdifferential \citep{clarke1990optimization}, since it is differentiable almost
everywhere \citep{carriere2021optimizing}.
Standard stochastic subgradient algorithms can then be used for optimizing (\ref{VRfunction}) with respect to the point cloud that parameterizes the (Vietoris-Rips) filtration \citep{carriere2021optimizing}.

The theory behind topologically optimizing the mapping
\begin{equation}
	\label{WAfunction}
	\mathcal{L}_{\mathrm{top}}\circ\mathrm{Pers}\circ\mathrm{WA}:A\rightarrow\mathbb{R},
\end{equation}
where $\mathrm{WA}$ maps a point cloud to its weak Alpha filtration, is analogous to the theory for Vietoris-Rips filtrations.
However, the Delanauy triangulation (as an abstract simplicial complex) may vary over the input point cloud $\mX\in(\mathbb{R}^d)^n$.
If the points in $\mX$ are in general position (Section \ref{SUBSEC::alpha}), we can restrict $A$ to the connected open subset of $(\mathbb{R}^d)^n$ which has the property that the Delaunay triangulation $\gK_{\mY}$ of every $\mY\in A$ is the same as the Delaunay triangulation $\gK$ of $\mX$, apart from the associated geometry \citep{carriere2021optimizing}.
This means that while the exact positioning of their simplices in $\mathbb{R}^d$ may differ, the same sets of points (e.g., marked by indices $1,\ldots,n$,) define the simplices of $\gK_{\mY}$ as of $\gK$, and each mapping in the composition (\ref{WAfunction}) can again be well defined.
Note that this does not imply that the Delaunay triangulation will necessarily remain the same over the entire process of topological optimization.
\subsection{Computational Analysis}
\label{SUBSEC::compcost}

Topological optimization currently requires recomputing the persistence diagram(s) for every iteration of the optimization.
Thus, the computational analysis of topological optimization boils down to the computational analysis of persistent homology, which unfortunately remains challenging to this day. 
Although persistent homology is an unparalleled tool for quantifying all from the finest to coarsest topological holes in data, it relies on algebraic computations (Algorithm  \ref{alg:ph}) which can be costly for larger sized simplicial complexes.
In the worst case, computing persistent homology is cubic in the number of simplices \citep{Otter2017}.

For a data set $\mX\subseteq\mathbb{R}^d$ of $n$ points, the weak Alpha complex has size and computation complexity $\mathcal{O}\left(n^{\ceil{\frac{d}{2}}}\right)$ \citep{Otter2017, toth2017handbook, somasundaram2021benchmarking}, resulting in a computation complexity $\mathcal{O}\left(n^{3\ceil{\frac{d}{2}}}\right)$ for persistent homology.
However, this can be significantly lower in practice, for example, if the points are distributed nearly uniformly on a polyhedron \citep{amenta2007complexity}.
The Vietoris-Rips filtration has size and computation complexity $\mathcal{O}(\min(2^n,n^{k+1}))$, where $d\geq k$ is the homology dimension of interest \citep{Otter2017, somasundaram2021benchmarking}, resulting in a computation complexity $\mathcal{O}(\min(2^{3n},n^{3(k+1)}))$ for persistent homology.
In practice, the choice of $k$ will also significantly reduce the size of the weak Alpha complex and thus the complexity of the persistent homology computation.
However, the full complex, i.e., the Delaunay triangulation, still needs to be constructed first \citep{boissonnat2018geometric}.
This is often the main bottleneck when working with weak Alpha complexes of high-dimensional data.

In practice, weak Alpha complexes are favorable for computing persistent homology from low dimensional point clouds, whereas fewer points in higher dimensions will favor Vietoris-Rips complexes \citep{somasundaram2021benchmarking}.
Within the context of topological regularization and data embeddings, we aim to achieve a low dimensionality $d$ of the point cloud embedding.
This justifies the choice for the weak Alpha filtrations.

Note that many computational improvements as well as approximation algorithms for persistent homology are already available \citep{Otter2017}.
However, their integration into topological optimization is open to further research.

\section{Topologically Regularized Data Embeddings}
\label{SEC::topreg}

In this section we propose a range of useful topological loss functions to optimize for topological priors in point cloud embeddings. We illustrate the effect of these losses on two-dimensional point cloud data  based on persistent homology from the weak $\alpha$-filtration. In Section~\ref{SUBSEC::k-dimensional-holes} we introduce the building blocks that optimize $k$-dimensional holes. A sampling strategy presented in Section~\ref{SUBSEC::sampling} improves the runtime and the representation of particular structures. In Section~\ref{SUBSEC::flares} we describe a more involved topological loss function to optimize flare-like structures. Finally we point out how to incorporate topological regularization into dimensionality reduction methods.

\subsection{Describing Topological Structures as $k$-Dimensional Holes}
\label{SUBSEC::k-dimensional-holes}
The topological loss is computed from the persistence diagram of a filtration on the two-dimensional point cloud embedding. As described in Section~\ref{SUBSEC::persistence}, the persistence diagram keeps track of the \textit{birth} and \textit{death} times of $k$-dimensional topological holes. These features or the differences between them, can be used to specify loss functions that bias the resulting embedding towards various topological structures when optimized. We instantiate the general loss from Equation~(\ref{eq:topo_loss}) by measuring the persistence of a $k$-dimensional hole with
\begin{equation}
	\label{eq:kdim-hole-loss}
	\mathcal{L}_{\mathrm{top}}(\gD_k)=\mu \sum_{l=i, d_l < \infty}^j d_l - b_l\,.
\end{equation}
We assume that the birth-death pairs are ordered such that $(b_1, d_1)$ is the pair with the largest persistence. To maximize or minimize we use $\mu \in \{-1, 1\}$ in all our experiments. 
Since we compute low-dimensional embeddings in two dimensions, we can only define loss functions on $\gD_0$ and $\gD_1$. For three dimensional embeddings, $\mathcal{L}_{\mathrm{top}}(\gD_2)$ could be defined as well following a similar intuition as for $\gD_1$ and $\gD_2$. In Section~\ref{PARAGRAPH::vary_g} we will introduce a more flexible version of Equation~(\ref{eq:topo_loss}) but stick with this basic measure of persistence for now.

\paragraph{0-dimensional holes (i.e., connected components)}
{\def\arraystretch{2}\tabcolsep=10pt
	\begin{table*}[t]
		\centering
		\begin{tabularx}{\textwidth}{p{0.5cm}
				>{\raggedright\arraybackslash}X
				>{\centering\arraybackslash}p{3.5cm}
				>{\centering\arraybackslash}p{3.5cm}}
			\toprule[1pt]
			i,j & Description & $\mu=1$ (min) & $\mu=-1$ (max)\\
			\midrule
			2,2 & Distance between points with the longest edge in the minimum spanning tree (MST) 
			& \includegraphics[height=2.5cm, width=3.5cm, valign=t, keepaspectratio]{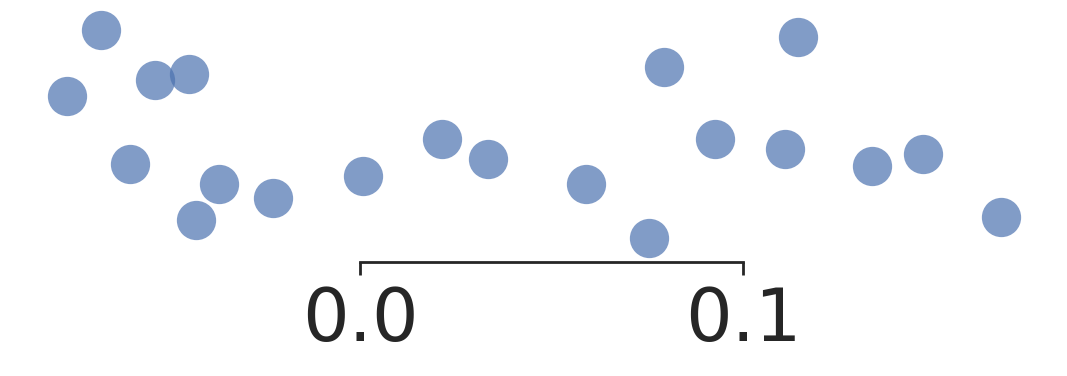} 
			& \includegraphics[height=2.5cm, width=3.5cm, valign=t, keepaspectratio]{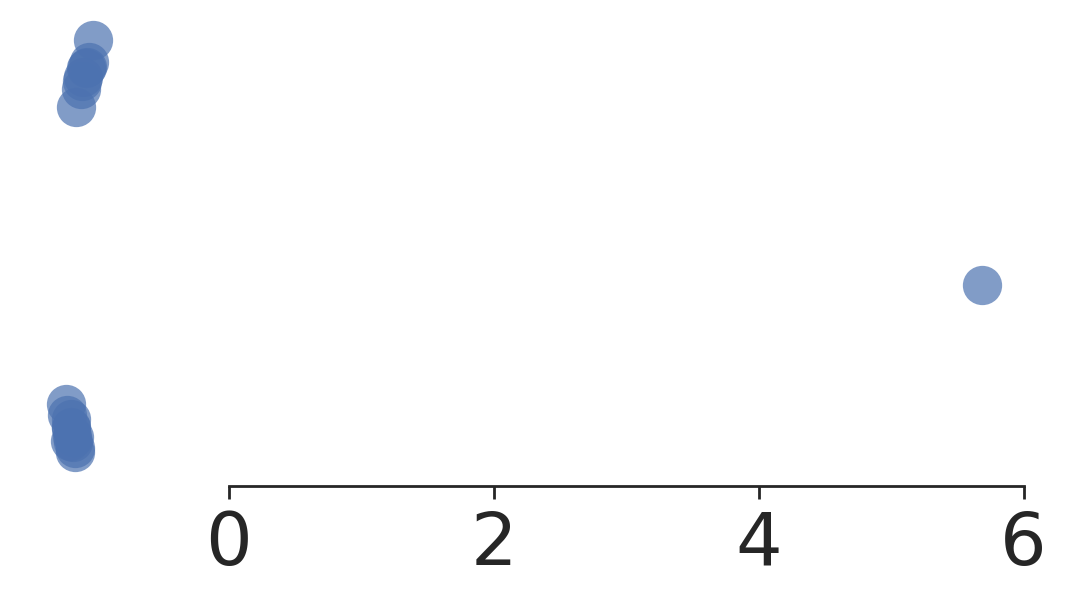}\\
			3,3 & Distance between points with the second longest edge in the MST
			& \includegraphics[height=2.5cm, width=3.5cm, valign=t, keepaspectratio]{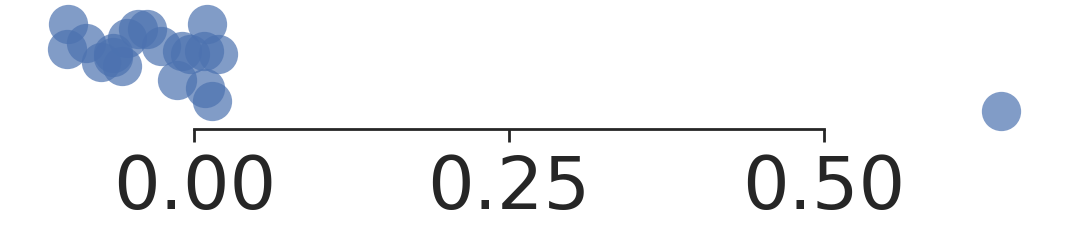} 
			& \includegraphics[height=2.5cm, width=3.5cm, valign=t, keepaspectratio]{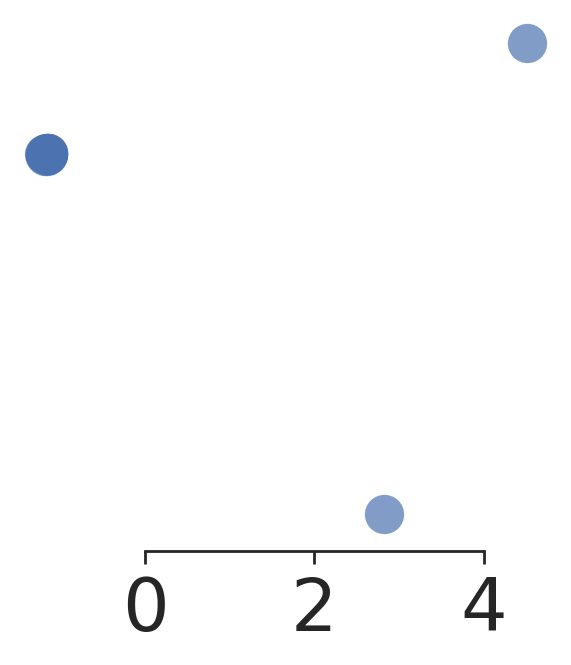}\\
			n,n & Distance between points with the shortest edge in the MST
			& \includegraphics[height=2.5cm, width=3.5cm, valign=t, keepaspectratio]{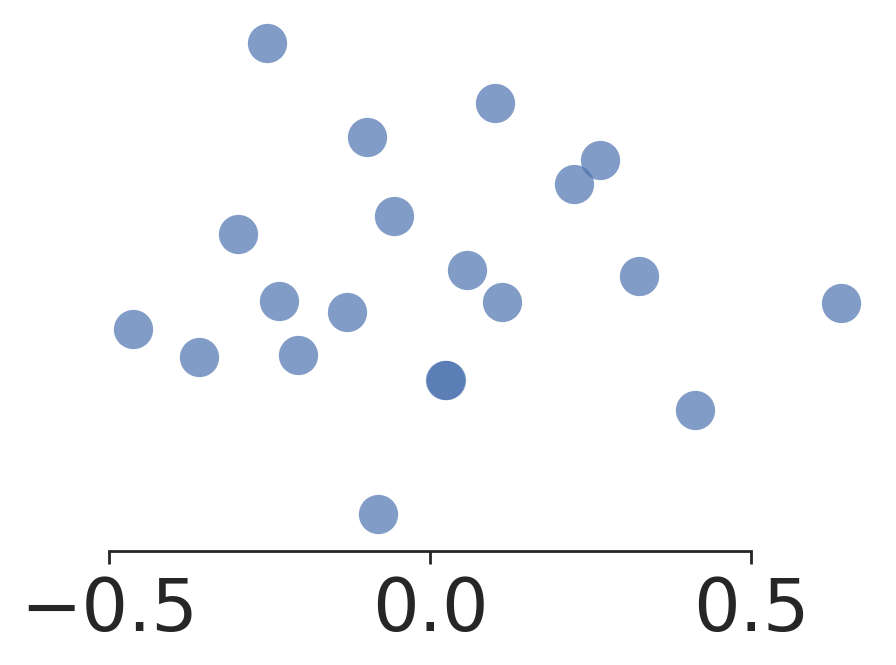} 
			& \includegraphics[height=2.5cm, width=3.5cm, valign=t, keepaspectratio]{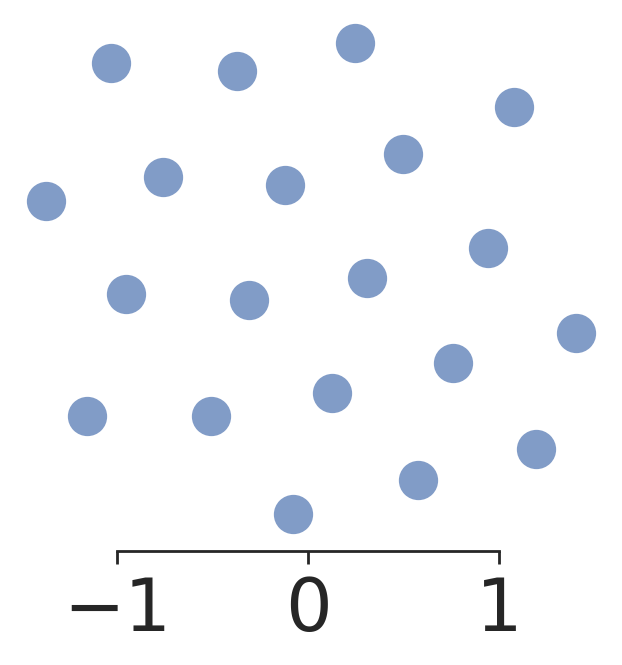}\\
			2,$\infty$ & All distances of the MST
			& \includegraphics[height=2.5cm, width=3.5cm, valign=t, keepaspectratio]{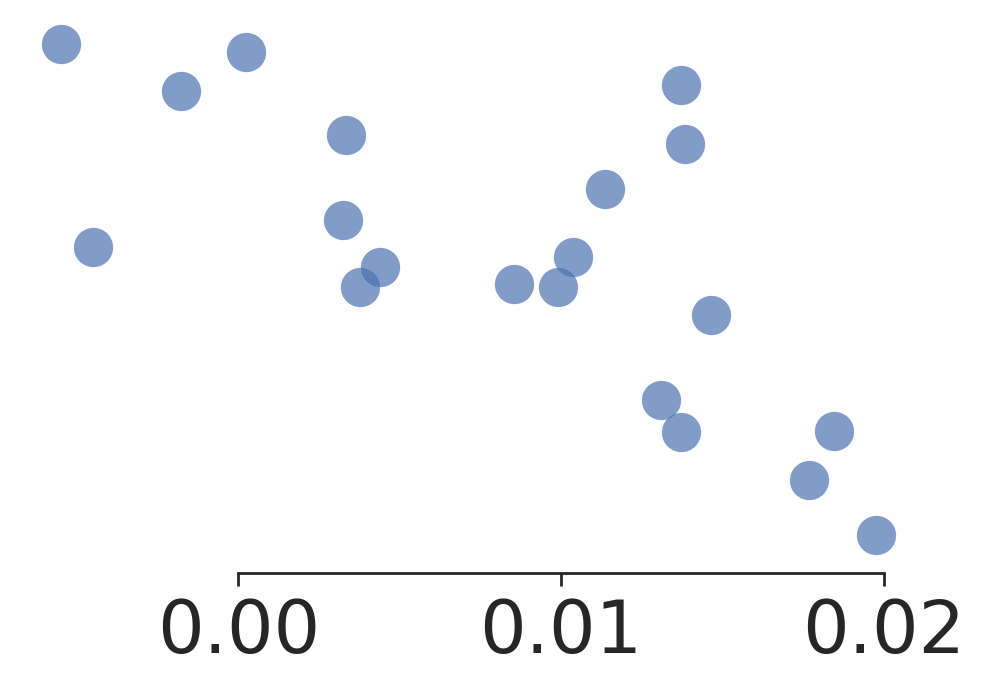} 
			& \includegraphics[height=2.5cm, width=3.5cm, valign=t, keepaspectratio]{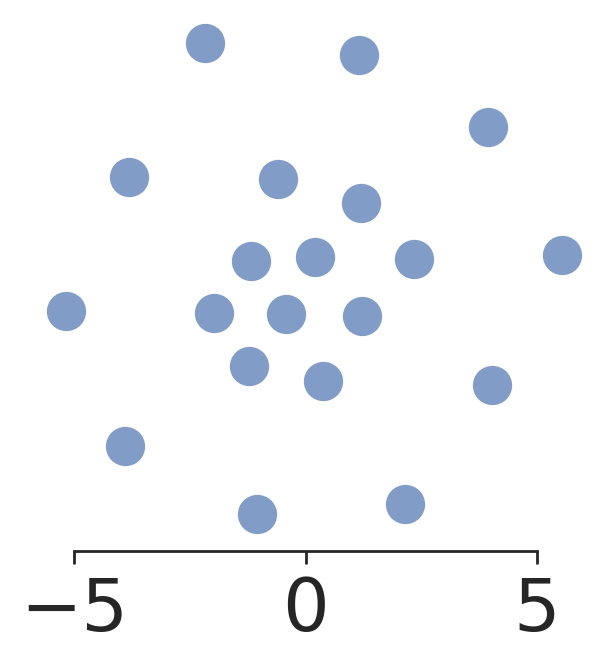}\\
			\bottomrule
		\end{tabularx}
		\caption{Optimizing topological loss functions using $\gD_0$ on synthetic data for 500 epochs.}
		\label{tab:topological_functions_D0}
	\end{table*}
}

For 0-dimensional holes based on the $\alpha$-filtration, the loss function reduces to $\mathcal{L}_{\mathrm{top}}(\gD_0)=\mu\sum_{l=i, d_l < \infty}^j d_l\,$ as the birth times are zero for all pairs. At the start of the filtration (time $\epsilon = 0$), each point constitutes one connected component and with increasing $\epsilon$, components are merged as they become connected through an edge. 
We illustrate the effect of different values for $\mu$, $i$, and $j$ in Table \ref{tab:topological_functions_D0}. The point cloud ($n=20$, $d=2$) is sampled from an isotropic Gaussian and we optimize the topological loss for 500 epochs.  

The 0-dimensional homology always contains $n$ pairs for a point cloud of size $n$ including one pair with infinite lifetime ($d_1 = \infty$). We thus start with $i=j=2$ which acts upon the persistence pair with the largest finite persistence. Depending on $\mu$, this loss will either decrease or increase the distance between the points that are connected the last. 
With $i=j=3$, the distance between points that connect second to last is decreased or increased. Note that when $\mu = 1$, this results in a large gap between the two clusters that connect last (one of which consists of only a single point in this case), as the distance between these two clusters does not influence the loss. When $\mu = -1$ however, the point cloud splits into three clusters where the two largest distances are similar (in this case the clusters happen to all three be equidistant) because delaying the second to last merge will eventually become the last merge of the filtration and vice versa. 

Minimizing the persistence with $i=j=n$ will make the two points that are closest to each other coincide, while maximizing it will distribute all points to have equal distances to their nearest neighbor and then increase the spacing between all points. 
With $i=2$ and $j=\infty$ (or $j=n$) we can either minimize or maximize all finite death times (note the scale of both embeddings). For $\mu = -1$ we observe a circular pattern with some points in the center which only increases in scale if we optimize for more epochs.

\paragraph{1-dimensional holes (i.e. cyclic topologies)}
{\def\arraystretch{2}\tabcolsep=10pt
	\begin{table*}[t]
		\centering
		\begin{tabularx}{\textwidth}{p{0.5cm}
		>{\raggedright\arraybackslash}X
		>{\centering\arraybackslash}p{3.5cm}
		>{\centering\arraybackslash}p{3.5cm}}
			\toprule[1pt]
			i,j & Description & $\mu=1$ (min) & $\mu=-1$ (max)\\
			\midrule
			1,1 & Persistence of the most persistent cycle
			& \includegraphics[height=2.5cm, width=3.5cm, valign=t, keepaspectratio]{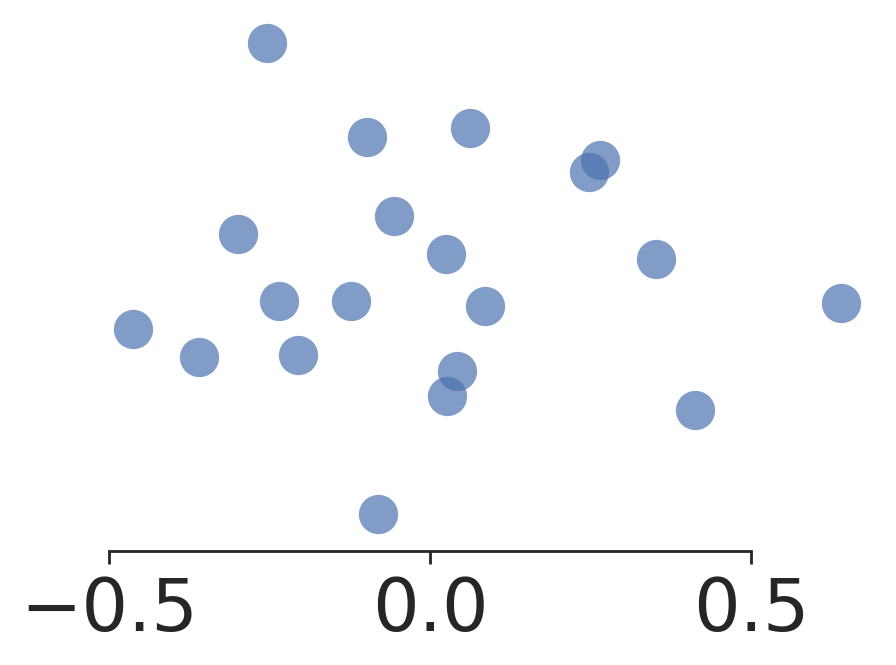}
			& \includegraphics[height=2.5cm, width=3.5cm, valign=t, keepaspectratio]{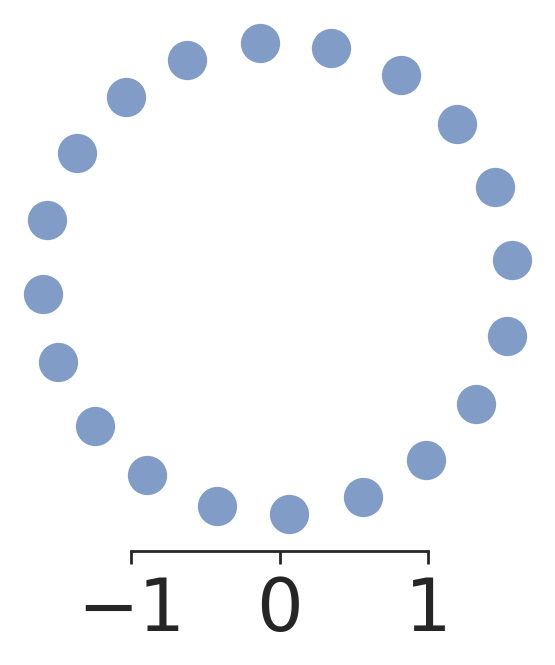}\\
			2,2 & Persistence of the second most persistent cycle
			& \includegraphics[height=2.5cm, width=3.5cm, valign=t, keepaspectratio]{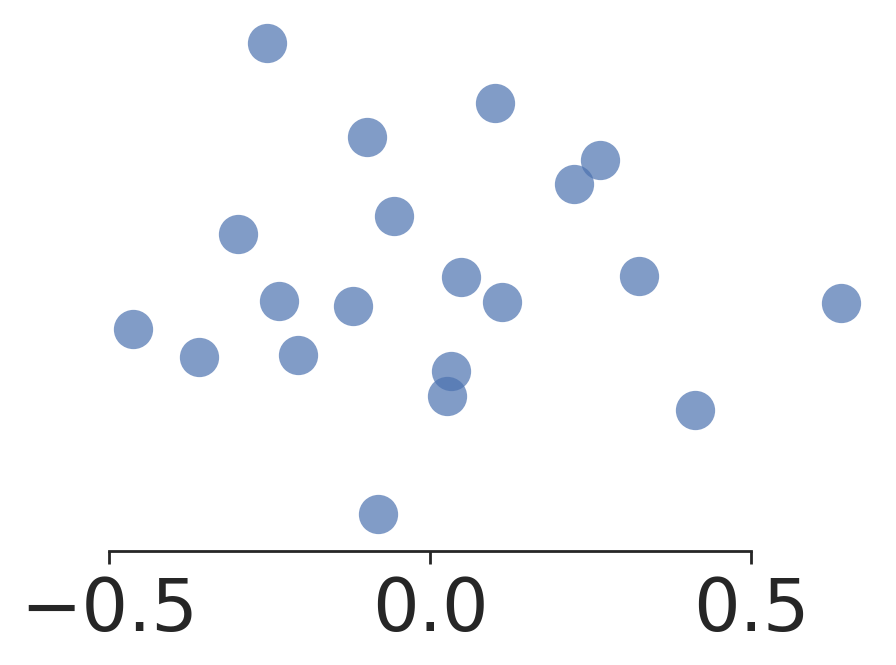}
			& \includegraphics[height=2.5cm, width=3.5cm, valign=t, keepaspectratio]{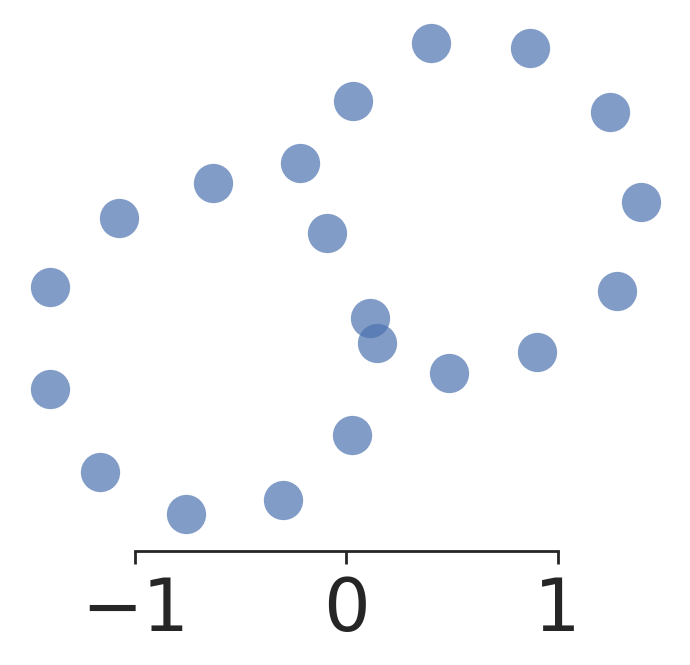}\\
			1,n & Persistence of all cycles
			& \includegraphics[height=2.5cm, width=3.5cm, valign=t, keepaspectratio]{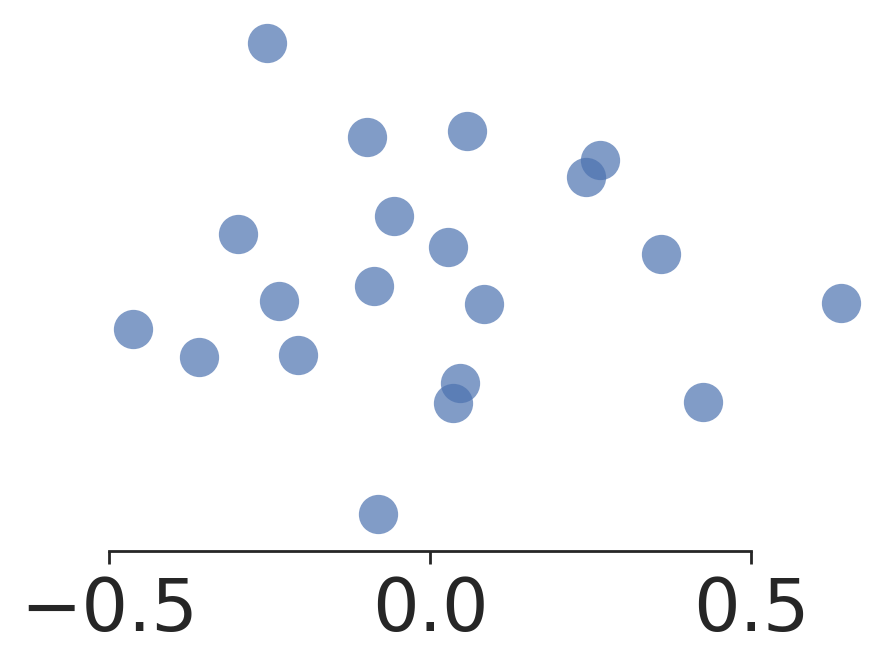}
			& \includegraphics[height=2.5cm, width=3.5cm, valign=t, keepaspectratio]{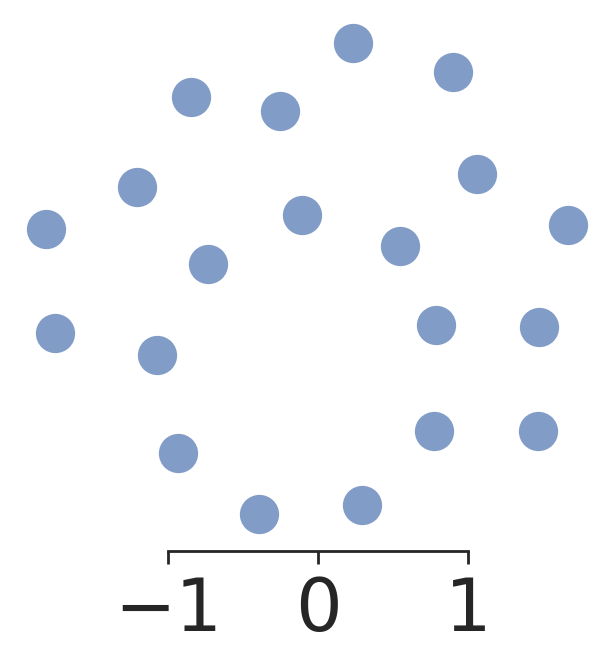}\\
			\bottomrule
		\end{tabularx}
		\caption{Optimizing topological loss functions using $\gD_1$ on synthetic data for 500 epochs.}
		\label{tab:topological_functions_D1}
	\end{table*}
}

With 1-dimensional holes we impose cyclic topologies on the two-dimensional embedding. In contrast to the 0-dimensional topological loss, the birth times $b_l$ in Equation~(\ref{eq:kdim-hole-loss}) are nonzero and contribute to the loss. In Table~\ref{tab:topological_functions_D1} we show the result of optimizing this loss on $\gD_1$ for different $i,j$ and $\mu$. First we point out that minimizing 1-dimensional topology ($\mu = 1$) leads to highly similar embeddings for different values of $i$ and $j$. These embeddings all lack 1-dimensional topologies in their persistence diagram $\gD_1$. No cycle is \textit{born} because the radius of any potential cycle is smaller than the distance of neighboring points that would be part of that cycle.

Optimizing the persistence of the most prominent hole with $i=j=1$ and $\mu = -1$ results in a circle with points evenly distributed on the boundary. Note that points outside the circle would not change the loss measuring only the largest persistence. Maximizing with $i=j=2$ leads to two equally sized circles as the optimization will alternate between the two. Optimizing the persistence of all cycles ($i=1, j=\infty$) results in one larger and several smaller circles. 

With the intuition from these basic examples we could build linear combinations of loss functions on $\gD_0$ and $\gD_1$, e.g. if we want the represented model to both be connected and include a circle, or know there are two circles that are not connected.

\subsection{Subsampling for Runtime and Representation Improvement}
\label{SUBSEC::sampling}

\begin{figure*}[t]
	\centering
	\begin{subfigure}{.3\linewidth}
		\centering
		\includegraphics[width=\linewidth, height=\linewidth, keepaspectratio]{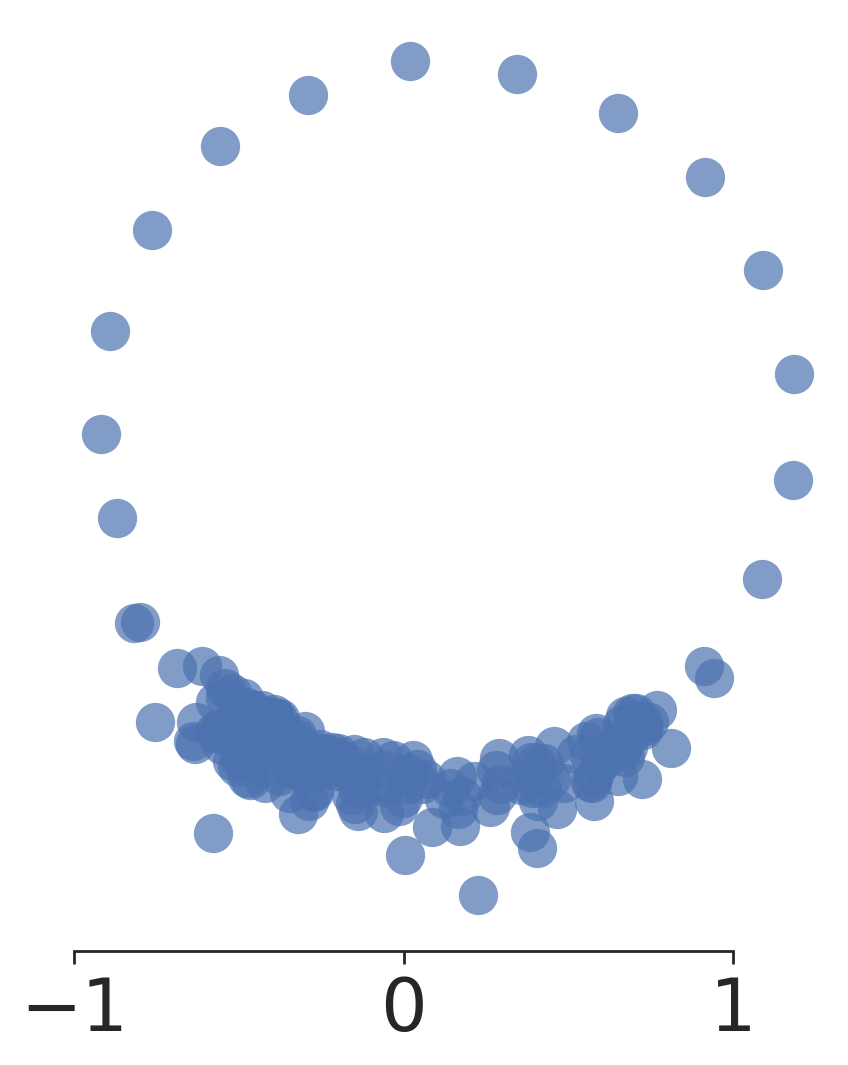}
		\caption{Optimization for 500 epochs (17s)}
		\label{fig:loss_function_circle_nosampling}
	\end{subfigure}\hfill
	\begin{subfigure}{.3\linewidth}
		\centering
		\includegraphics[width=\linewidth, height=\linewidth, keepaspectratio]{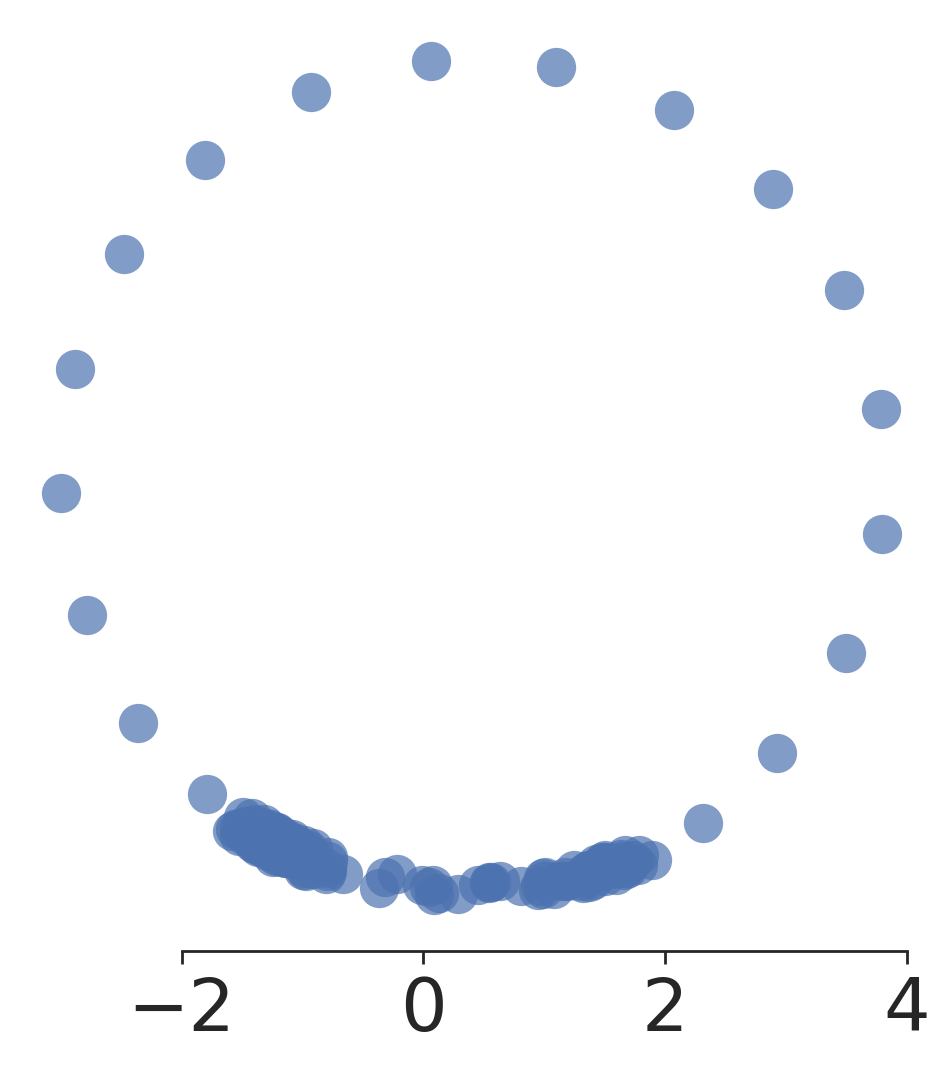}
		\caption{Optimization for 2k epochs (1min 7s)}
		\label{fig:loss_function_circle_2000epochs}
	\end{subfigure}\hfill
	\begin{subfigure}{.3\linewidth}
		\centering
		\includegraphics[width=\linewidth, height=\linewidth, keepaspectratio]{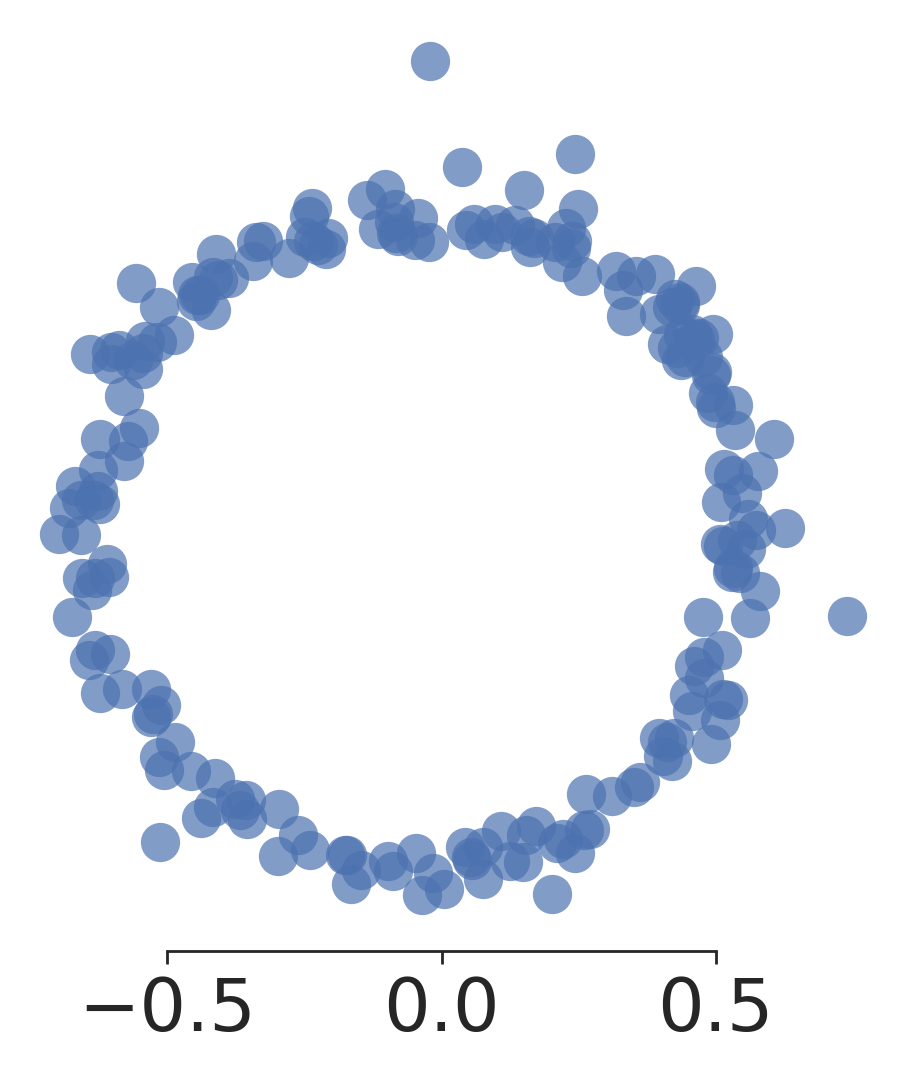}
		\caption{Optimization with sampling loss for 500 epochs (5s) with $f_{\mathcal{S}} = 0.2$}
		\label{fig:loss_function_circle_sampling}
	\end{subfigure}
	\caption{Topological optimization of a two-dimensional dataset with $n=200$ points on $\gD_1$ with with $i=j=1$ and $\mu=-1$.}
	\label{fig:loss_function_circle_sampling_comparison}
\end{figure*}

The examples in Table \ref{tab:topological_functions_D0} and \ref{tab:topological_functions_D1} showed that persistent homology effectively measures the prominence of topological holes. However, it is often ineffective for representing such holes in a natural manner. To illustrate, we optimize the most persistent cycle in a larger two-dimensional dataset with $n=200$ points sampled from an isotropic Gaussian and show the result in Figure \ref{fig:loss_function_circle_nosampling}. While some points are a crucial part of the cycle and will affect the loss when (re-)moved, others lie outside the boundary and do not influence the topological loss. Even more, their position will not change until they are part of the circle as each gradient update only affects the \textit{four} points contributing to birth and death of the cycle (see Figure \ref{PikaGrad}). The representation might improve slightly when optimizing for more epochs (Figure \ref{fig:loss_function_circle_2000epochs}), but the runtime quickly becomes prohibitive. 

To represent topological holes through more points, we propose to optimize the loss
\begin{equation}
	\label{sampleloss}
	\widetilde{\mathcal{L}}_{\mathrm{top}}(\mE)\coloneqq\mathbb{E}\left[\mathcal{L}_{\mathrm{top}}\left(\left\{\vx\in \rmS:\rmS \mbox{ is a random sample of } \mE \mbox{ with sampling fraction } f_{\mathcal{S}}\right\}\right)\right],
\end{equation}
where $\mathcal{L}_{\mathrm{top}}$ is defined as in (\ref{eq:kdim-hole-loss}).
In practice, during each optimization iteration, $\widetilde{\mathcal{L}}_{\mathrm{top}}$ is approximated by the mean of $\mathcal{L}_{\mathrm{top}}$ evaluated over $n_{\mathcal{S}}$ random samples of the point cloud $\mE$.

In Figure \ref{fig:loss_function_circle_sampling} we show the result of optimizing a topological sampling loss of the most persistent cycle with $f_{\mathcal{S}} = 0.2$ and $n_{\mathcal{S}} = 1$. 
Evaluating the topological loss on a subset of the data leads to a more balanced distribution of points around the circular shape. 
An added benefit of the sampling-based loss in Equation~(\ref{sampleloss}) is that optimization can be conducted significantly faster, as persistent homology is evaluated on smaller samples.

\paragraph{Computational Cost of Topological Loss Functions with Sampling}
When one makes use of a sampling fraction $0<f_{\mathcal{S}}\leq 1$ along with $n_{\mathcal{S}}$ repeats, the computational complexity of the topological loss function reduces to $\mathcal{O}\left(n_{\mathcal{S}}\left((f_{\mathcal{S}}n)^{3\ceil{\frac{d}{2}}}\right)\right)$ when using weak $\alpha$-filtrations, where $d$ is the embedding dimensionality.
Nevertheless, as discussed in Section \ref{SUBSEC::compcost}, the computational complexity may often be significantly lower in practice, due to constraining the dimension of the homology for which we optimize, and because common distributions across natural shapes admit reduced size and time complexities of the Delaunay triangulations \citep{dwyer1991higher, amenta2007complexity, devillers2019poisson}.

\subsection{Optimization of Flare Structures}
\label{SUBSEC::flares}

Persistent homology is invariant to certain topological changes.
For example, both a linear `I'-structured model and a bifurcating `Y'-structured model consist of one connected component, and no higher-dimensional holes.
These models are indistinguishable based on the (persistent) homology thereof, even though they are topologically different in terms of their singular points.

Capturing such additional topological phenomena is possible through a refinement of persistent homology under the name of \emph{functional persistence}, also well discussed and illustrated by \citet{carlsson_2014}.
Instead of evaluating persistent homology on a data matrix $\mE$, we evaluate it on a subset $\{\vx\in\mE:f(\vx)\leq \tau\}$ for a well chosen function $f:\mE\rightarrow\mathbb{R}$ and hyperparameter $\tau$. Inspired by this approach, for a diagram $\mathcal{D}$ of a point cloud $\mE$, we propose the topological loss
\begin{equation}
	\label{functionalloss}
	\widetilde{\mathcal{L}}_{\mathrm{top}}(\mE)\coloneqq\mathcal{L}_{\mathrm{top}}\left(\{\vx\in\mE:f_{\mE}(\vx)\leq \tau\}\right),\mbox{ informally denoted } \left[\mathcal{L}_{\mathrm{top}}(\mathcal{D})\right]_{f_{\mE}^{-1}]-\infty,\tau]},
\end{equation}
where $f$ is a real-valued function on $\mE$, possibly dependent on $\mE$---which changes during optimization itself---, $\tau$ a hyperparameter, and $\mathcal{L}_{\mathrm{top}}$ is an ordinary topological loss as defined by Equation~(\ref{eq:kdim-hole-loss}).

In the experiments we will optimize \emph{flares}, i.e. star-shaped structures using the idea of functional persistence. In particular, we will focus on the case where $f$ equals the scaled centrality measure on $\mE$:
\begin{equation}
	\label{centrality}
	f_{\mE}\equiv\mathcal{E}_{\mE}\colonequiv 1-\frac{g_{\mE}}{\max g_{\mE}},\mbox{ where }g_{\mE}(\vx)\coloneqq \left\|\vx-\frac{1}{|\mE|}\sum_{\vy\in\mE}\vy\right\|\,.
\end{equation}
With $\tau\geq 1$ we have $\widetilde{\mathcal{L}}_{\mathrm{top}}(\mE)=\mathcal{L}_{\mathrm{top}}(\mE)$. 
For sufficiently small $\tau > 0$, $\widetilde{\mathcal{L}}_{\mathrm{top}}$ evaluates $\mathcal{L}_{\mathrm{top}}$ on the points `far away' from the mean in the center of $\mE$. In Section \ref{SEC::experiments} we will combine this idea with 0-dimensional persistence to optimize the flares.

\subsection{Incorporating Topological Regularization in Dimensionality Reduction Methods}
\label{SUBSEC::topo_DR}
To regularize point cloud embeddings $\mE$, we combine an embedding loss with the topological loss that represents the prior knowledge on the data.
The goal is to find an embedding that minimizes a total loss function
\begin{equation}
	\label{totalloss}
	\mathcal{L}_{\mathrm{tot}}(\mE, \mathbb{X})\coloneqq \mathcal{L}_{\mathrm{emb}}(\mE, \mathbb{X})+\lambda_{\mathrm{top}} \mathcal{L}_{\mathrm{top}}(\mE),
\end{equation}
where $\mathcal{L}_{\mathrm{emb}}$ aims to preserve structural attributes of the original data, and $\lambda_{\mathrm{top}}>0$ controls the strength of \emph{topological regularization}.
Note that $\mathbb{X}$ itself is not required to be a point cloud, or reside in the same space as $\mE$.
This is especially useful for representation learning of graphs.
We can thus use topological regularization for embedding a graph $G$, to learn a representation of the nodes of $G$ in $\mathbb{R}^d$ that well respects properties of $G$. To embed the graph, we used a DeepWalk variant adapted from \citet{graph_nets}. 
To regularize graph embeddings by DeepWalk or embeddings of tabular data by UMAP, we simply combine their objectives with the topological loss $\mathcal{L}_{\mathrm{top}}$ as in Equation~(\ref{totalloss}).

In addition to DeepWalk and UMAP, we also present experiments on \emph{regularized PCA projections}. 
Topologically regularizing a PCA projection is more involved, as using $$\mathcal{L}_{\mathrm{emb}}(\mW, \mX)\coloneqq \mathrm{MSE}\left(\mX\mW\mW^T, \mX\right)$$
does not ensure orthonormality of the matrix $\mW$. To address this we use Pymanopt \citep{pymanopt} to optimize over the Stiefel manifold of orthonormal matrices. The line-search algorithms to compute the optimal step size for every iteration of the optimization, however, do not work with our topological sampling loss. We therefore removed the backtracking part from the line-search.
Note that in the experiments of the conference version of this paper \citep{vandaele2021topologically} we used an orthogonality loss $\|\mW^T\mW - \mI\|_F$ on the weights $\mW \in \mathbb{R}^{n \times 2}$, which had unwanted effects on the optimization.\footnote{In the experiments on the synthetic cycle, the orthogonality loss shifted the weights of $\mW$ to the first two data dimensions - not the regularization with the topological loss. We suspect this was a numerical artifact of the orthogonality loss combined with a too high learning rate.}
\section{Experiments and Applications}
\label{SEC::experiments}
In this section, we illustrate the effects of topological regularization on synthetic and real-world data. In Section \ref{sec:effectiveness}, we show how to optimize embeddings with the ground truth prior and quantify the effect on subsequent prediction tasks. 
In Section \ref{sec:robustness}, we explore the robustness of topological regularization with different parametrizations of the loss.
We design the experiments around six research questions that are stated in the beginning of these Sections. Our code is available at \url{https://github.com/aida-ugent/TopoEmbedding}.

\subsection{Datasets}
\begin{figure*}[t]
	\centering
	\begin{subfigure}[t]{.325\textwidth}
		\centering
		\includegraphics[width=\linewidth]{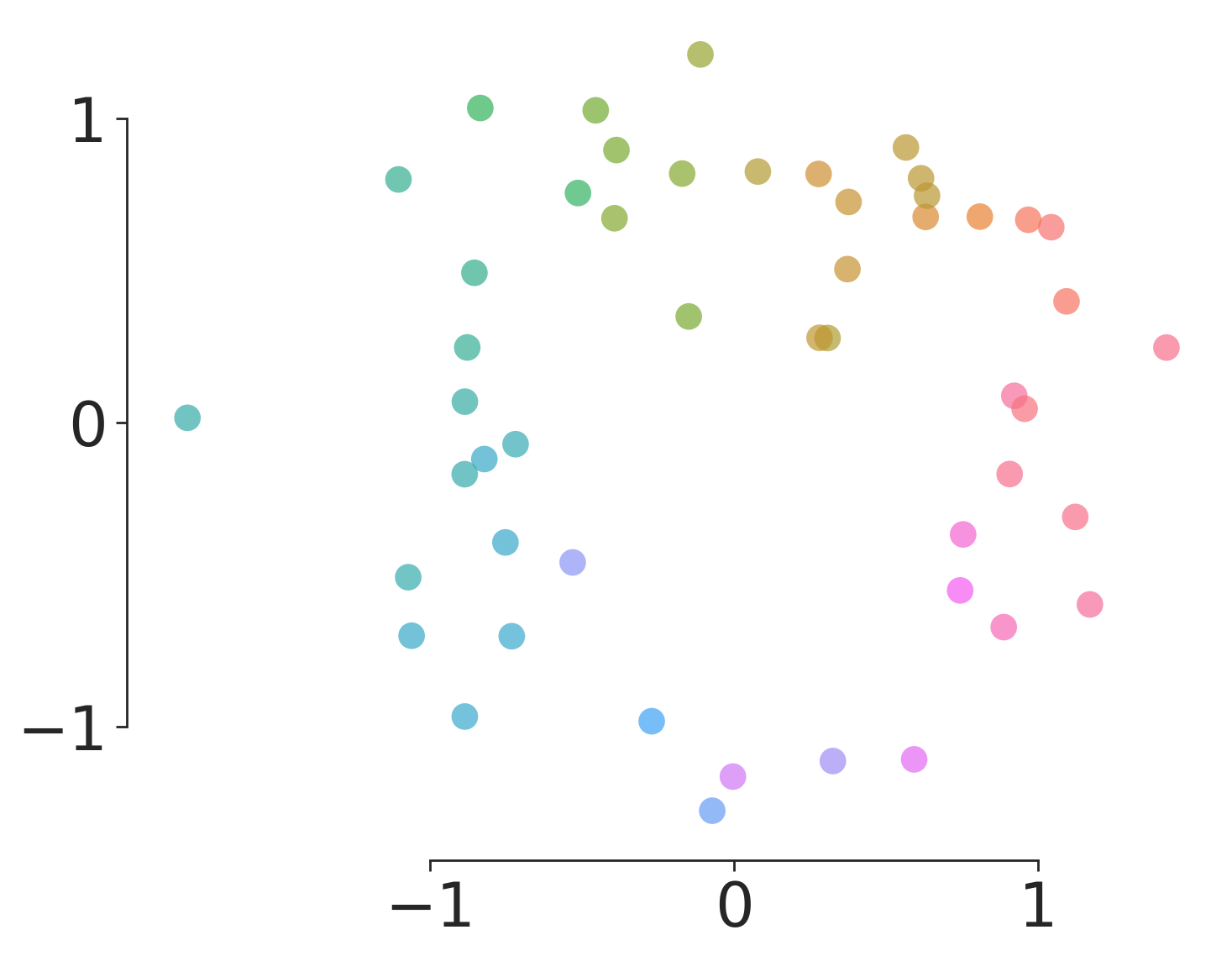}  
		\captionsetup{width=.9\linewidth}
		\caption{First two data coordinates of the 500 dimensional synthetic cycle data.}
		\label{fig:synthcircle12}
	\end{subfigure}\hfill
	\begin{subfigure}[t]{.325\textwidth}
		\centering
		\includegraphics[width=\linewidth]{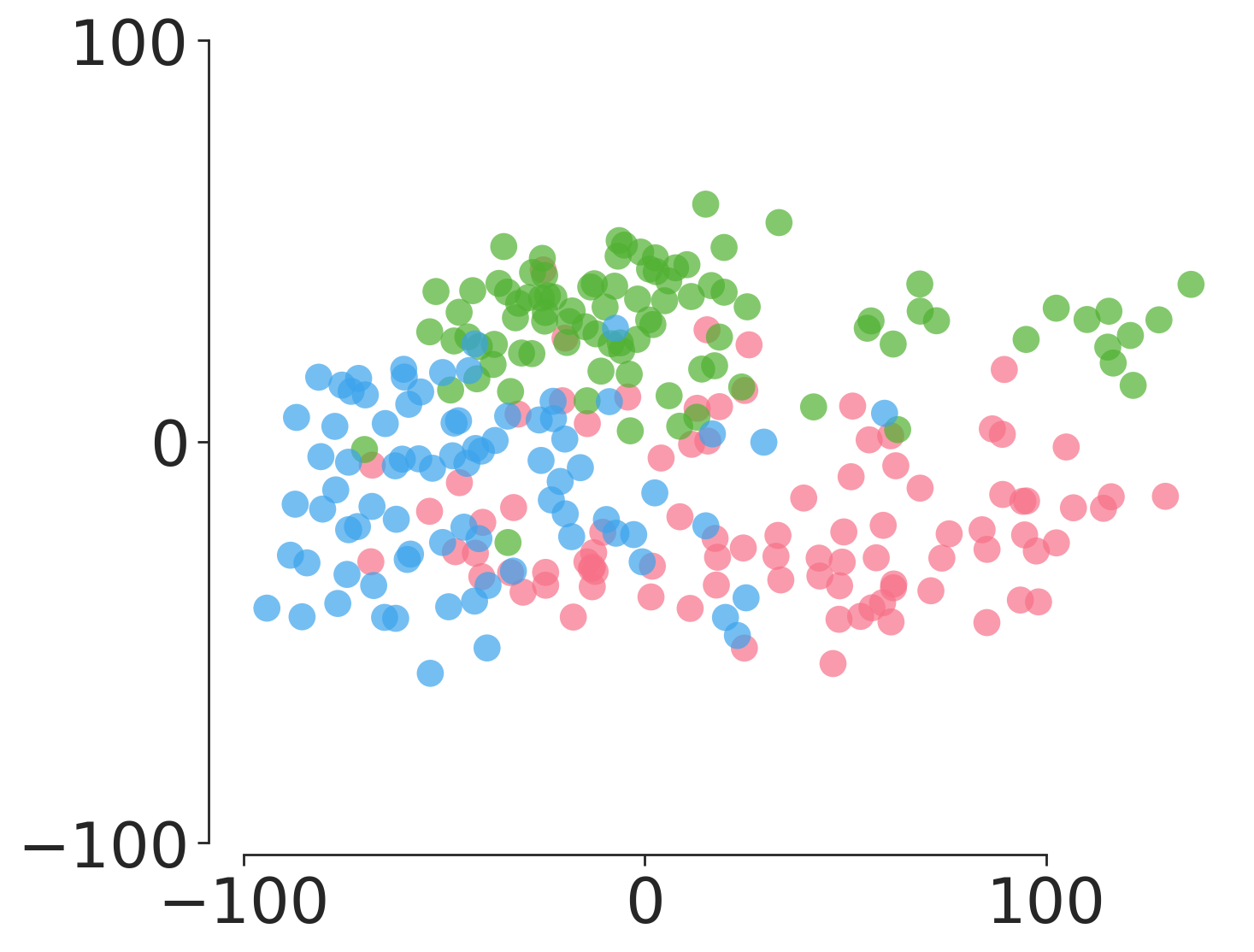}  
		\captionsetup{width=.9\linewidth}
		\caption{Cell cycle data with colors indicating cell state (PCA).}
		\label{fig:cellcycle}
	\end{subfigure}\hfill
	\begin{subfigure}[t]{.325\textwidth}
		\centering
		\includegraphics[width=\linewidth]{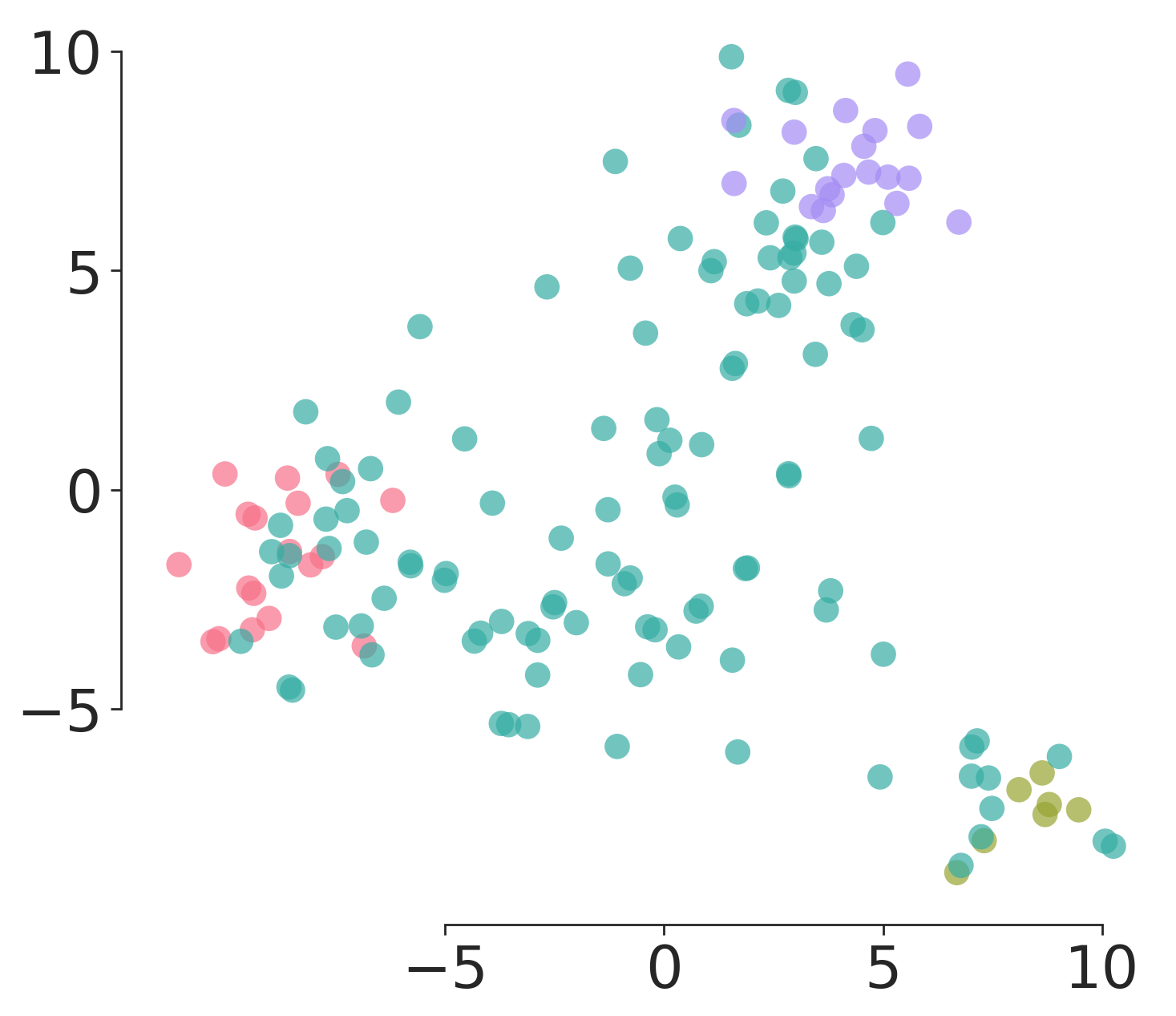}  
		\captionsetup{width=.9\linewidth}
		\caption{Cell bifurcation data with colors indicating cell group (UMAP).}
		\label{fig:cellbifurcation}
	\end{subfigure}
	\begin{subfigure}[t]{.325\textwidth}
		\centering
		\includegraphics[width=.95\linewidth]{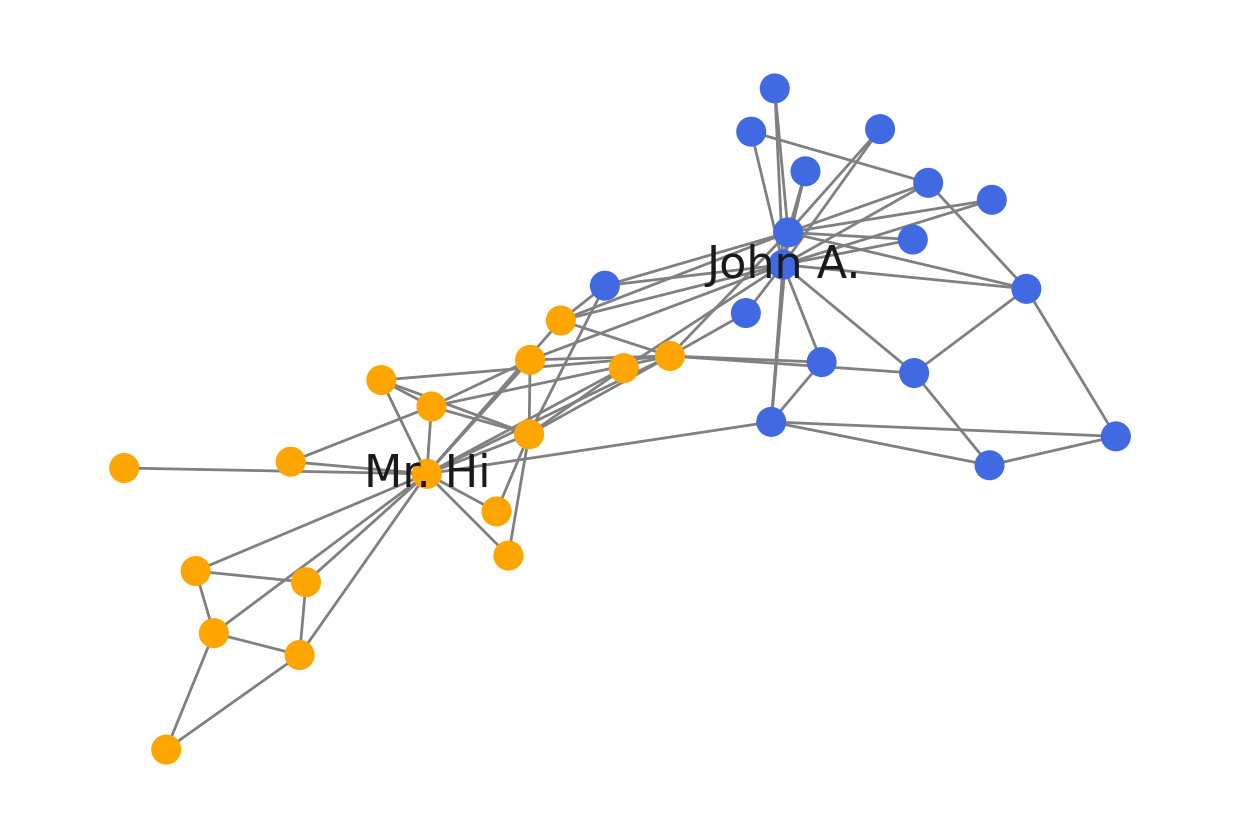}
		\caption{Karate network (NetworkX spring layout).}
		\label{fig:karate}
	\end{subfigure}
	\begin{subfigure}[t]{.65\textwidth}
		\includegraphics[width=\linewidth]{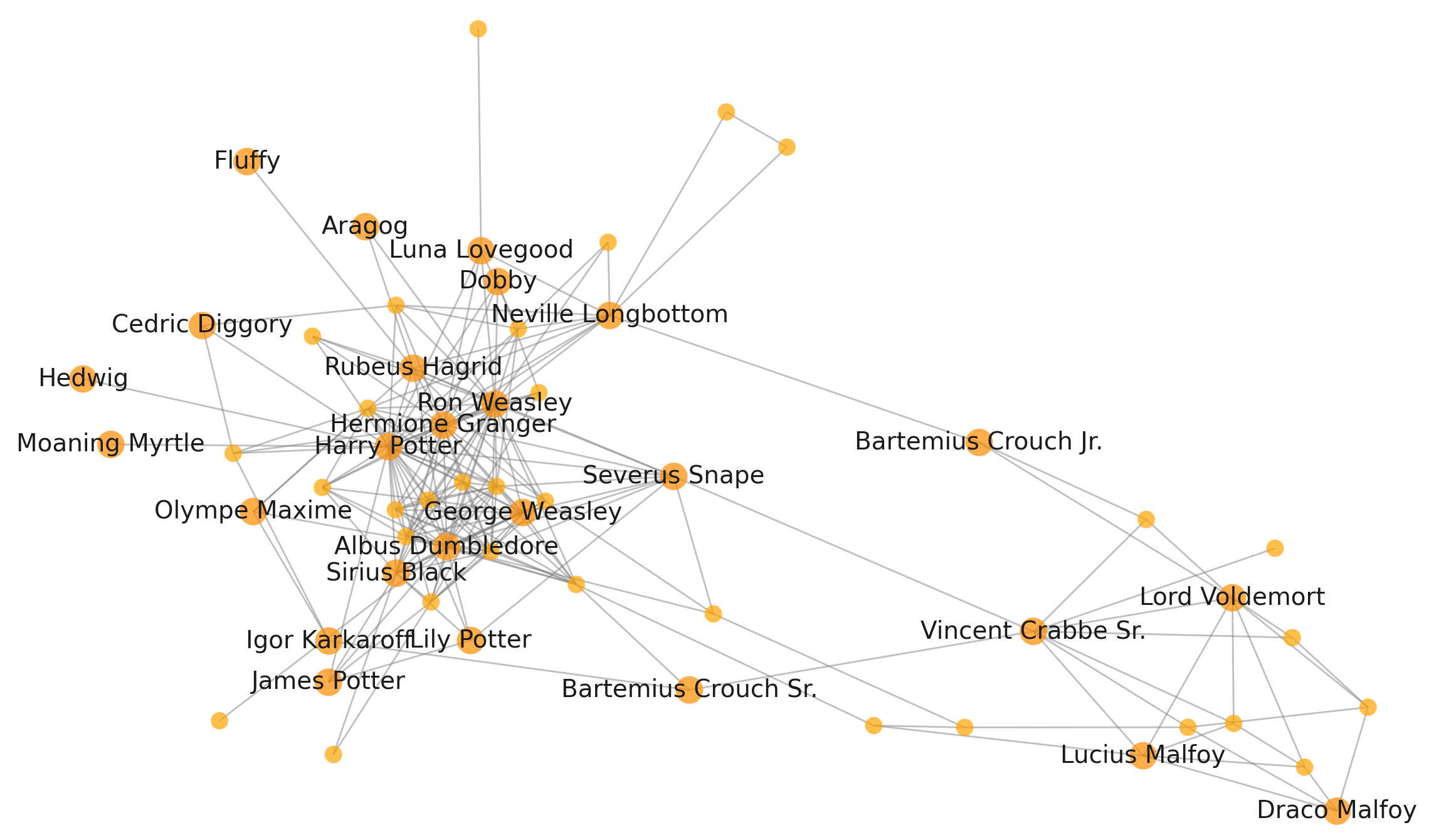}  
		\caption{The major connected component in the Harry Potter graph (NetworkX spring layout).
			Edges mark friendly relationships between characters.}
		\label{fig:harrypotter}
	\end{subfigure}
	\caption{Visualizations of the datasets used in the experiments. The method to obtain the two-dimensional embeddings is indicated in brackets.}
	\label{fig:datasets}
\end{figure*}

We evaluate the effectiveness and robustness of topological regularization on one synthetic, two biological, and two network datasets. We provide their size and ground truth model in Table \ref{tab:datasets} and visualizations in Figure \ref{fig:datasets}.

\begin{description}
	\item[Synthetic cycle] We sample $50$ points uniformly from the unit circle in $\mathbb{R}^2$. 
	We then added 500-dimensional noise, sampled uniformly from $[-0.45, 0.45]$ in each dimension.
	\item[Cell cycle] We considered a single-cell trajectory data set of 264 cells in a 6812-dimensional gene expression space \citep{robrecht_cannoodt_2018_1443566, Saelens276907}.
	This data can be considered a snapshot of the cells at a fixed time.
	The ground truth is a circular model connecting three cell groups through cell differentiation (see also Section \ref{pseudotime}). 
	\item[Cell bifurcation] We considered a second cell trajectory data set of 154 cells in a 1770-dimensional expression space \citep{robrecht_cannoodt_2018_1443566}.
	The ground truth here is a bifurcating model connecting four different cell groups through cell differentiation.
	\item[Karate] The Karate network \citep{Zac77} is a well known and studied network within graph mining that consists of two different communities. The communities are represented by two key figures (John A.\ and Mr.\ Hi), as shown in Figure \ref{fig:karate}.
	\item[Harry Potter] The nodes of this  graph (\href{https://github.com/hzjken/character-network}{https://github.com/hzjken/character-network}) are characters from the Harry Potter novel and edges mark friendly relationships between them (Figure \ref{fig:harrypotter}).
	We only use the largest connected component in our experiments. 
	The Harry Potter graph has previously been analyzed by \citet{JMLR:v21:19-1032}, who identified a circular model therein that transitions between the `good' and `evil' characters from the novel.
\end{description}

\begin{table*}
	\renewcommand*{\arraystretch}{1.3}
	\centering
	\begin{tabular}{lrrl}
		\toprule
		Name & n & d & Ground truth model \\
		\midrule
		Synthetic Cycle & 50 & 500 & circle in dims 1 and 2 \\
		Cell Cycle & 264 & 6812 & circle \\
		Cell Bifurcating & 154 & 1770 & bifurcation \\
		Karate & 34 & 78 & two clusters \\
		Harry Potter & 58 & 217 & circle \\
		\bottomrule
	\end{tabular}
	\caption{Summary of data sizes and their ground truth model.}
	\label{tab:datasets}
\end{table*}

\subsection{Effectiveness of Topological Regularization}
\label{sec:effectiveness}
In this section we design and analyze experiments with the goal of answering the following questions about the effectiveness of topological regularization:

\begin{description}
	\item[Q1.] Does topological regularization succeed in imposing the specific topological structure in the low dimensional embedding?
	\item[Q2.] How does topological regularization affect downstream prediction tasks?
	\item[Q3.] How scalable is topological regularization?
\end{description}

An overview of the dimension reduction methods and their optimization parameters is shown in Table \ref{tab:effectiveness_parameters} and the topological loss functions are provided in Table \ref{tab:effectiveness_loss}. 

\begin{table*}
	\renewcommand*{\arraystretch}{1.2}
	\centering
	\begin{tabular}{llrrrrr}
		\toprule
		Data & Method & lr & epochs & $\lambda_{\mathrm{top}}$ & $t$ w/o top & $t$ with top \\
		\midrule
		Synthetic Cycle & PCA & 0.01 & $\leq 1000$ & 0.0005 & $<$1s & 12s \\
		Cell Cycle & PCA & 0.01 & $\leq 1000$ & 0.01 & $<$1s & 3m29s \\
		Cell Bifurcating & UMAP & 0.1 & 100 & 10 & $<$1s & 4s \\
		Karate & DeepWalk & 1e-2 & 50 & 50 & 19s & 20s \\
		Harry Potter & InnerProd & 1e-1 & 100 & 1e-1 & 1m21s & 1m24s \\
		\bottomrule
	\end{tabular}
	\caption{Summarization of optimization parameters for each dataset.}
	\label{tab:effectiveness_parameters}
\end{table*}

\begin{table*}[t]
	\centering
	\renewcommand\tabcolsep{5pt}
	\renewcommand\arraystretch{1.2}
	\begin{tabular}{lllccc}
		\toprule
		Data & Figure & Topological loss function & dim & $f_{\mathcal{S}}$ & $n_{\mathcal{S}}$ \\
		\midrule
		Synthetic Cycle & \ref{fig:synthCirclePCAopt},\ref{fig:SynthCircleTop}, \ref{fig:RandTop} & $-(d_1-b_1)$ & 1 & 0.4 & 5\\
		Cell Cycle & \ref{CellCirclePCAopt}, \ref{CellCircleTop} & $-(d_1-b_1)$ & 1 & 0.25 & 10\\
		Cell Bifurcating & \ref{fig:CellBifUMAP}, \ref{fig:CellBifUMAPopt}, \ref{fig:CellBifTop} & {\small $\sum_{k=2}^\infty (d_k-b_k)-\left[d_3-b_3\right]_{\mathcal{E}_{\mE}^{-1}]-\infty,0.75]}$ } & 0 - 0 & {\color{gray} N/A} & {\color{gray} N/A} \\
		Karate & \ref{KarateDWopt}, \ref{KarateTop} & $-(d_2-b_2)$ & $0$ & $0.25$ & $10$\\
		Harry Potter & \ref{HarryIPopt}, \ref{HarryTop} & $-(d_1-b_1)$ & 1 & {\color{gray} N/A} & {\color{gray} N/A}\\
		\bottomrule
	\end{tabular}
	\caption{Summary of the topological losses computed from persistence diagrams $\mathcal{D}$ with points $(b_k, d_k)$ ordered by persistence $d_k-b_k$. 
	Note that for 0-th dimensional homology diagrams $d_1=\infty$.}
	\label{tab:effectiveness_loss}
\end{table*}

\subsubsection{Qualitative Evaluation}
\label{sec:effec_qualitative}

In this section we tackle Q1, illustrating the effect of topologically regularizing embeddings by comparing them to topologically optimized and standard embeddings for each of the five datasets. We use topological loss functions to model the ground truth structure of the data (see in Table \ref{tab:datasets}). 

We use the term \emph{optimized embeddings} when optimizing only the topological loss for a predefined number of epochs initialized with a PCA, UMAP, or DeepWalk embedding. For the optimized PCA embeddings, we do not use the reconstruction loss but still optimize over the Stiefel manifold resulting in an orthogonal projection. \emph{Regularized embeddings} are the result of minimizing a combination between the embedding and topological loss. We set the trade-off $\lambda_{\mathrm{top}}$ between these two losses as low as possible but as high as needed such that the regularized embedding is notably different from the optimized embedding.
The optimized and regularized embeddings are computed using the same number of epochs (see Table \ref{tab:effectiveness_parameters}) when using DeepWalk and UMAP.
For PCA, we use a maximum of 1000 epochs but stop the optimization when the topological loss begins to stagnate (measured by the ratio between the average topological loss of the last 100 epochs and the average of last 50 epochs).

\paragraph{Synthetic Data}
\label{SEC::synthetic}

\begin{figure*}[t]
	\centering
	\begin{subfigure}[t]{.325\textwidth}
		\centering
		\includegraphics[width=\linewidth]{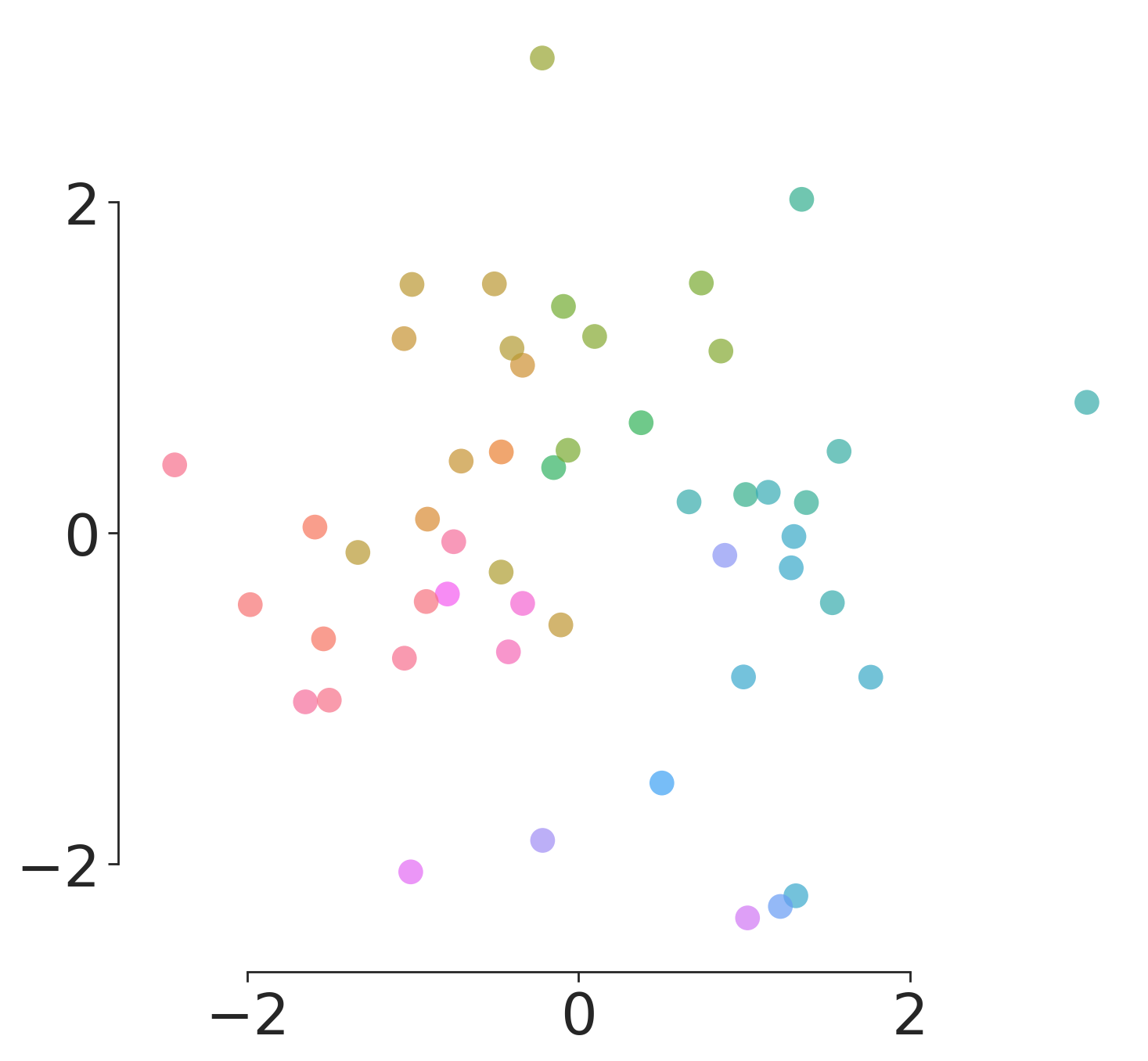}  
		\caption{Ordinary PCA embedding.}
		\label{fig:SynthCirclePCA}
	\end{subfigure}
	\hfill
	\begin{subfigure}[t]{.325\textwidth}
		\centering
		\includegraphics[width=\linewidth]{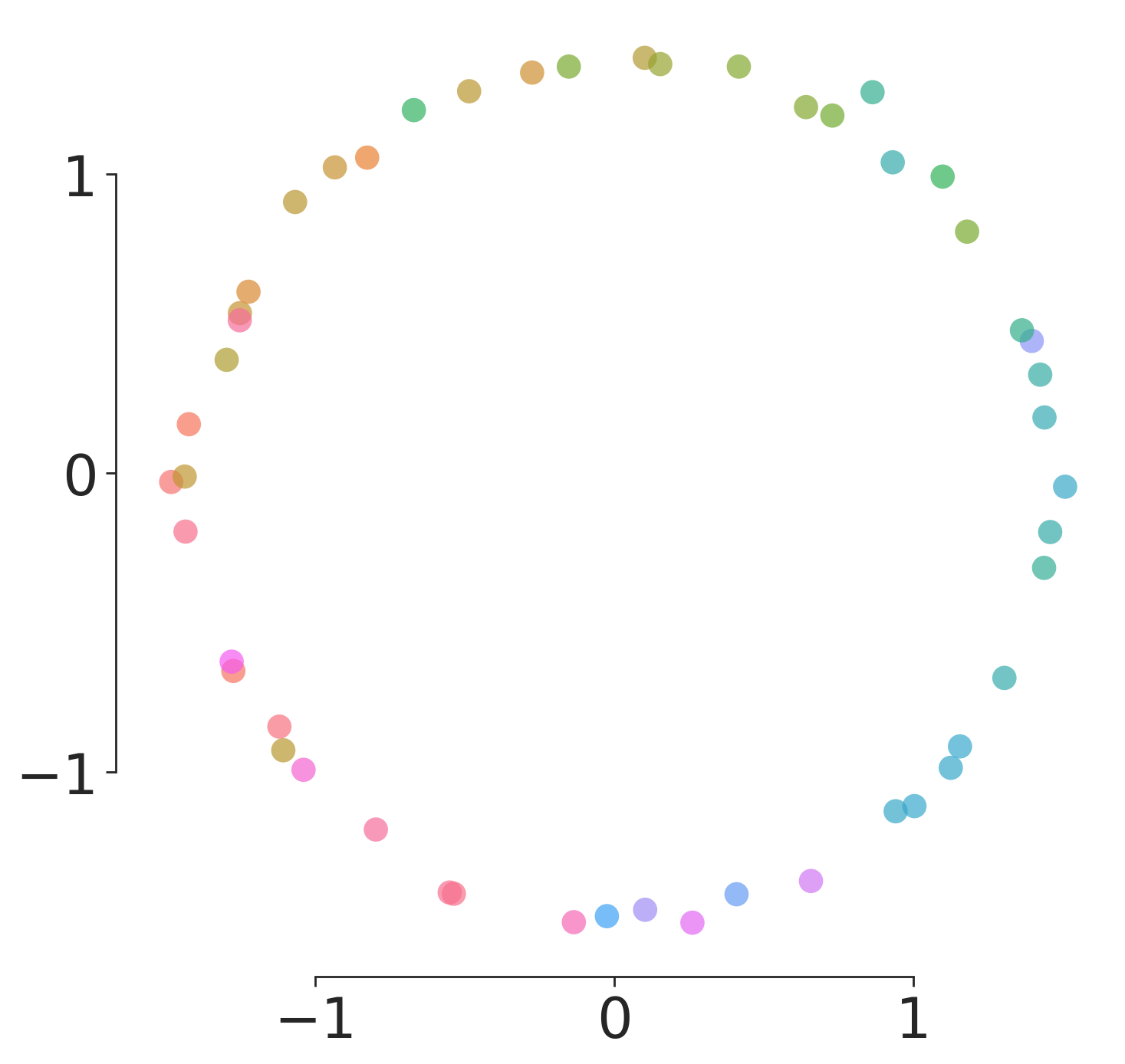}  
		\caption{Top.\ optimized embedding.}
		\label{fig:synthCirclePCAopt}
	\end{subfigure}
	\hfill
	\begin{subfigure}[t]{.325\textwidth}
		\centering
		\includegraphics[width=\linewidth]{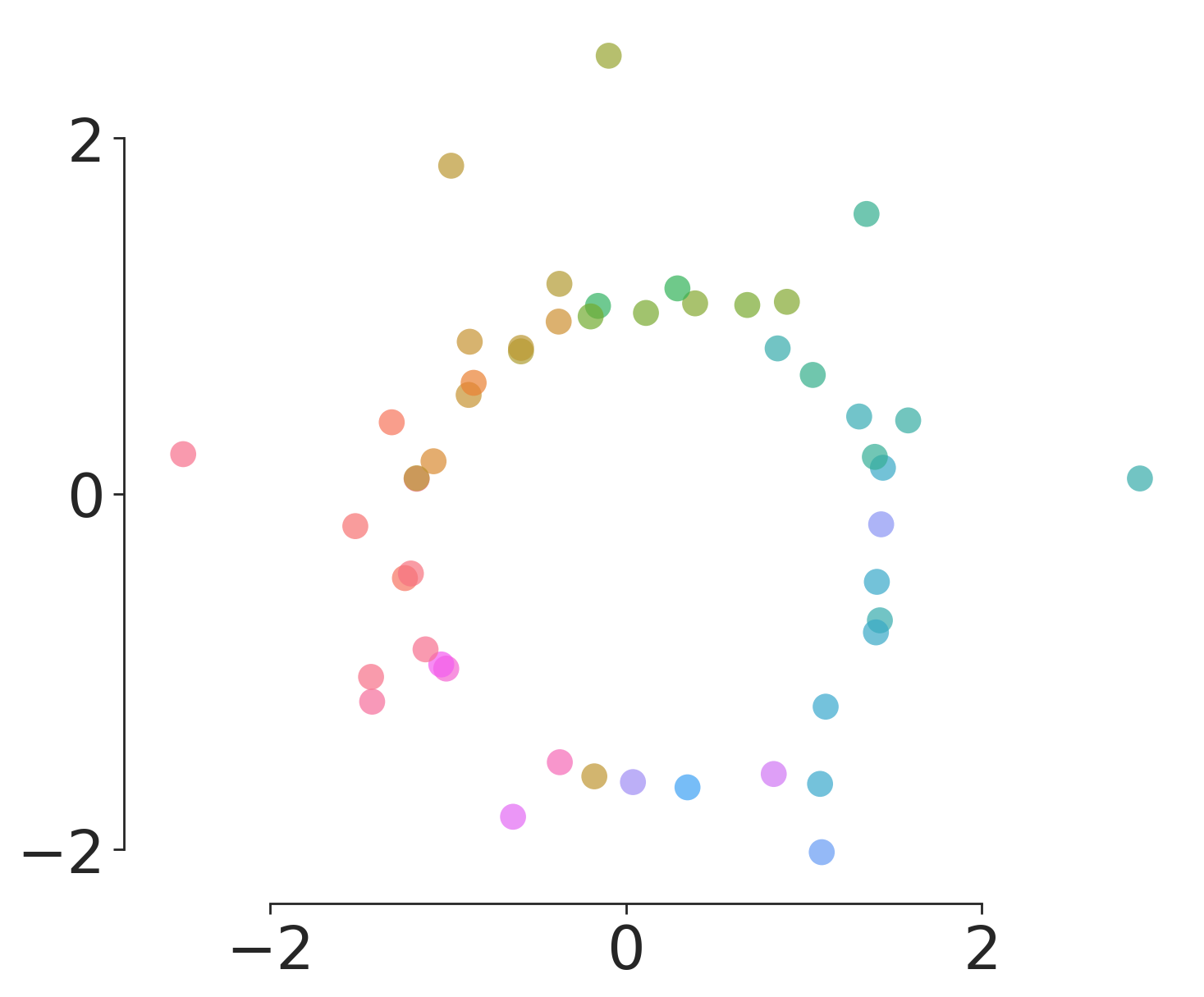}  
		\caption{Top. regularized embedding.}
		\label{fig:SynthCircleTop}
	\end{subfigure}
	\caption{Representations of the synthetic data, colored by ground truth coordinates.}
	\label{SynthCircle}
\end{figure*}

For the 500-dimensional synthetic circle, the 498 additional noisy features are irrelevant to the topological (circular) model. An ideal projection embedding would be a restriction to its first two features (see Figure \ref{fig:synthcircle12}). However, it is probabilistically unlikely that the irrelevant features will have a zero contribution to a PCA embedding of the data. Intuitively, each added feature slightly shifts the projection plane away from the plane spanned by the first two features. In Figure \ref{fig:SynthCirclePCA}, we observe that the circular hole is indeed less present in the PCA embedding of the data.
Note that \emph{we observed this to be a notable problem for `small $n$ large $p$' data sets}, as similar to other machine learning models, and as also recently studied by \citet{vandaele2022curse}, more data can significantly accommodate for the effect of noise and result in a better embedding model on its own.

To improve the visual representation of the circle, we use the topological loss function $\mathcal{L}_{\mathrm{top}}(\gD_1) = -(d_1-b_1)\,$
measuring the persistence of the most prominent 1-dimensional hole in the embedding. Since we aim for all points to lie on the boundary of the circle, we evaluate this loss on subsets of the data ($f_{\mathcal{S}}=0.4$, $n_{\mathcal{S}} = 5$). 
The resulting embedding is shown in Figure \ref{fig:SynthCircleTop}, which better captures the circular hole. For comparison, Figure \ref{fig:synthCirclePCAopt} shows the optimized embedding (initialized with PCA) without the reconstruction loss $\mathcal{L}_{\mathrm{emb}}$ with a more prominent hole.

\paragraph{Circular Cell Trajectory Data}
\label{SEC::cellcycle}

To improve the representation of the circular model in the single-cell trajectory, we use the same loss as for the synthetic data but different sampling parameters ($f_{\mathcal{S}}=0.25$, and $n_{\mathcal{S}}=10$) as the size of the single-cell dataset is larger. 

\begin{figure*}[t]
	\centering
	\begin{subfigure}[t]{.325\linewidth}
		\centering
		\includegraphics[width=\linewidth]{Images/CellCycle/CellCirclePCA}  
		\caption{Ordinary PCA embedding.}
		\label{CellCirclePCA}
	\end{subfigure}
	\begin{subfigure}[t]{.325\linewidth}
		\centering
		\includegraphics[width=\linewidth]{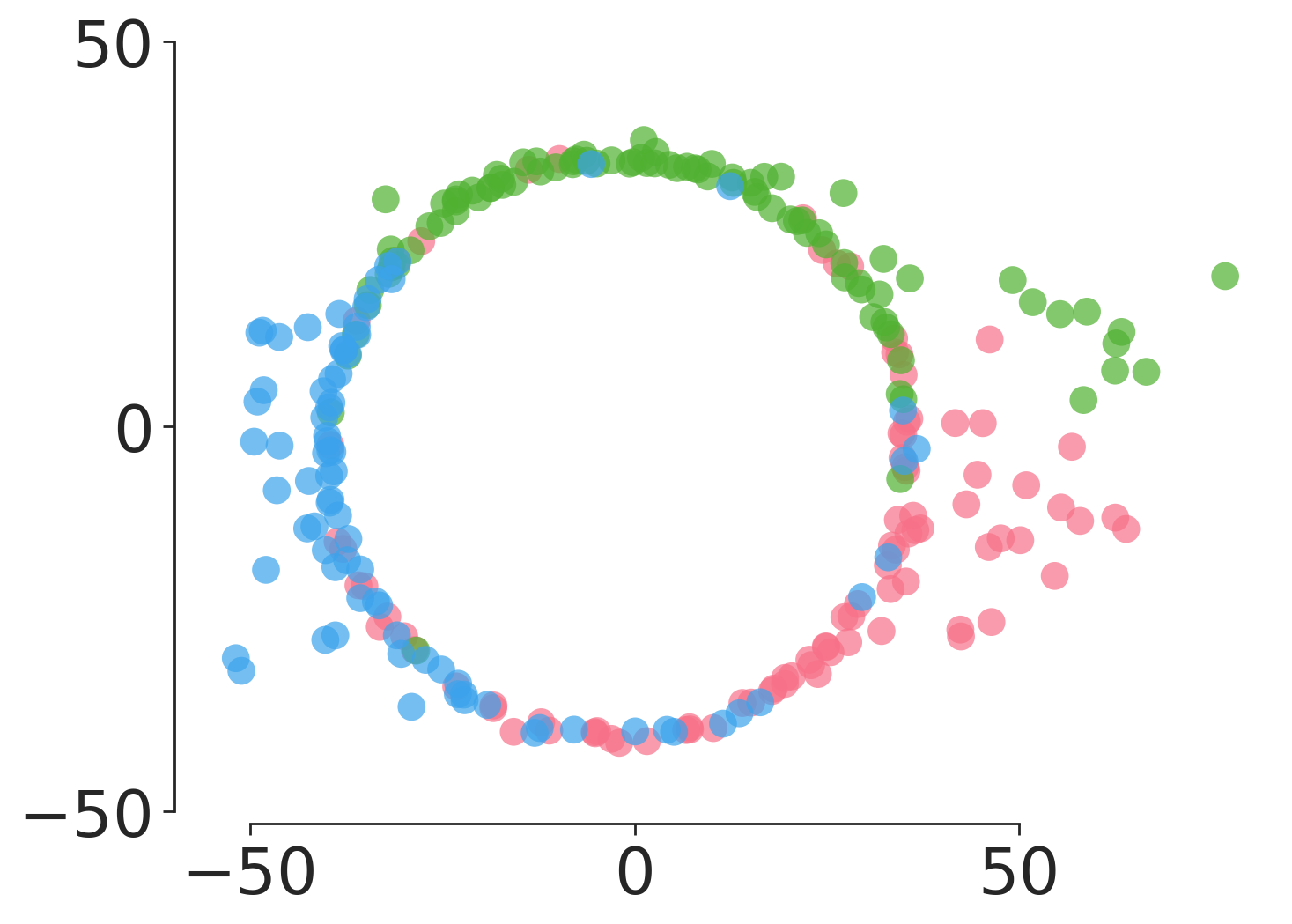}  
		\caption{Top.\ optimized embedding.}
		\label{CellCirclePCAopt}
	\end{subfigure}
	\begin{subfigure}[t]{.325\linewidth}
		\centering
		\includegraphics[width=\linewidth]{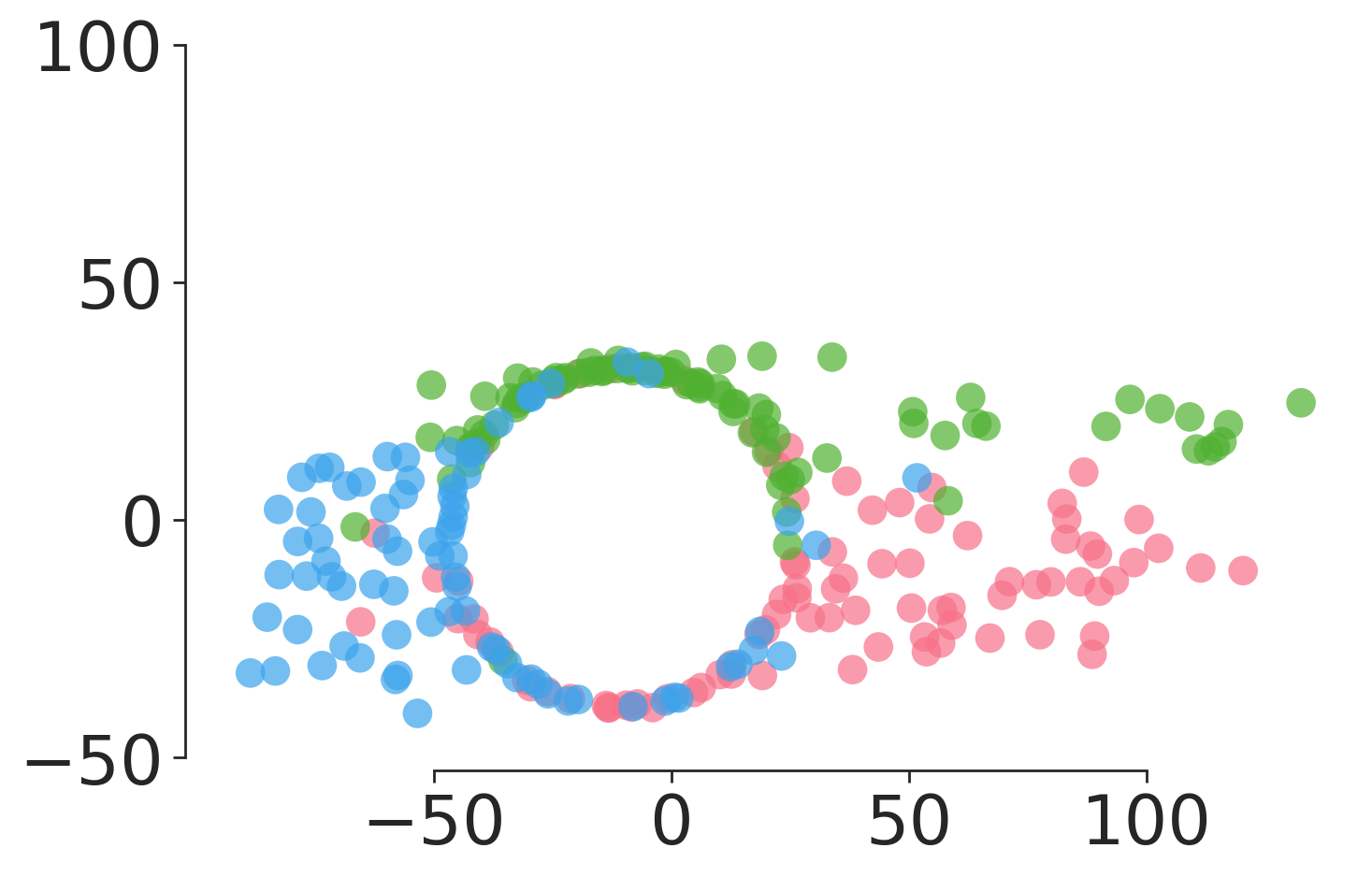}  
		\caption{Top.\ regularized embedding.}
		\label{CellCircleTop}
	\end{subfigure}
	\caption{Representations of the cyclic cell data with colors representing the cell group.}
\end{figure*}

From Figure \ref{CellCirclePCA}, we see that while the ordinary PCA embedding does somehow respect the positioning of the cell groups (marked by their color), it indeed struggles to embed the data in a manner that visualizes the cycle that we know to be present in the data. By topologically regularizing the embedding as shown in Figure \ref{CellCircleTop}, we are able to embed the data in a circular manner. Compared to the optimized embedding in Figure \ref{CellCirclePCAopt}, there are more cells that lie outside of the circle. 

\paragraph{Pseudotime inference in cell trajectory data}
\label{pseudotime}

Single-cell omics include various types of data collected on a cellular level, such as transcriptomics, proteomics and epigenomics.
Studying the topological model underlying the data may lead to better understanding of the dynamic processes of biological cells and the regulatory interactions involved therein.
Such dynamic processes can be modeled through trajectory inference (TI) methods, also called \emph{pseudotime analysis}, which order cells along a trajectory based on the similarities in their expression patterns \citep{Saelens276907}.

For example, the \emph{cell cycle} is a well-known biological differentiation model that describes a cell as it grows and divides. The cell cycle consists of different stages, namely growth (G1), DNA synthesis (S), growth and preparation for mitosis (G2), and mitosis (M). The latter two stages are often grouped in a G2M stage. Hence, by studying expression data of cells that participate in the cell cycle differentiation model, one may identify the genes involved in and between particular stages of the cell cycle \citep{liu2017reconstructing}. Pseudotime analysis allows such study by assigning to each cell a time during the differentiation process in which it occurs, and thus, the relative positioning of all cells within the cell cycle model.

\begin{figure}[t]
	\centering
	\begin{subfigure}[t]{.9\linewidth}
		\includegraphics[height=.4cm]{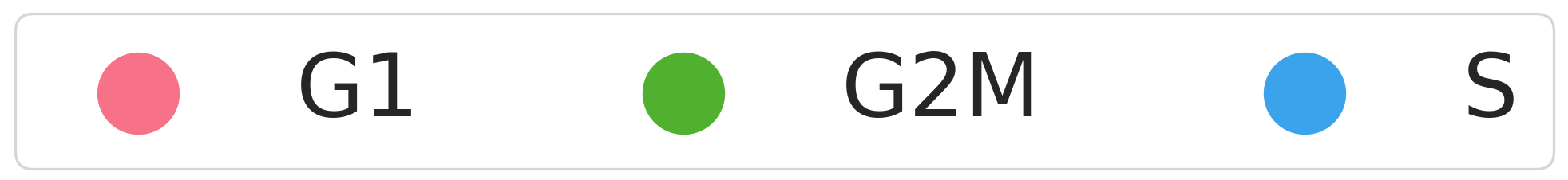}
	\end{subfigure}\\
	\begin{subfigure}[t]{.32\linewidth}
		\centering
		\includegraphics[height=3.7cm]{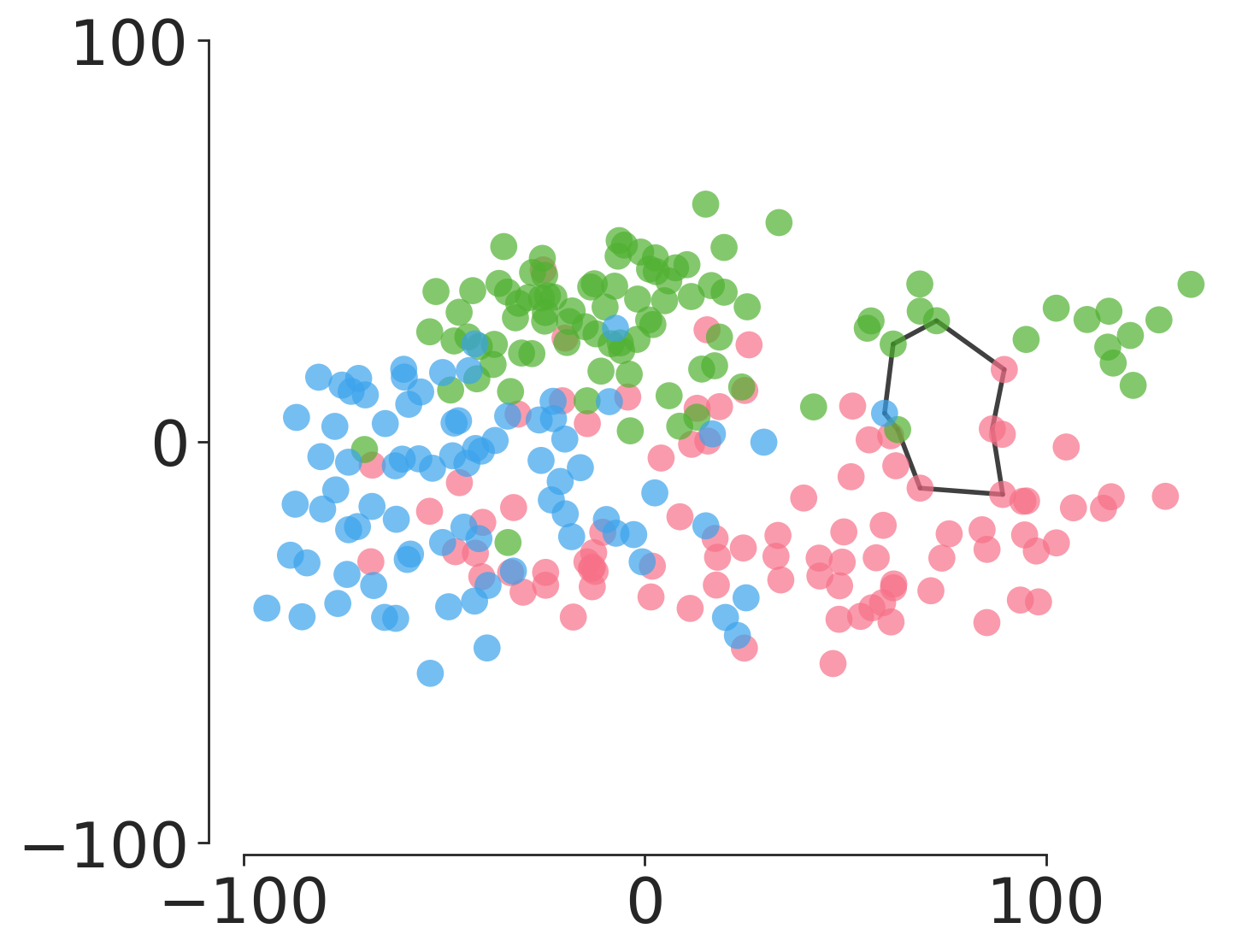}  
		\caption{The representation of the most prominent cycle obtained through persistent homology.}
		\label{fig:diode_cycle_pca}
	\end{subfigure}\hfill%
	\begin{subfigure}[t]{.3\linewidth}
		\centering
		\includegraphics[height=3.8cm]{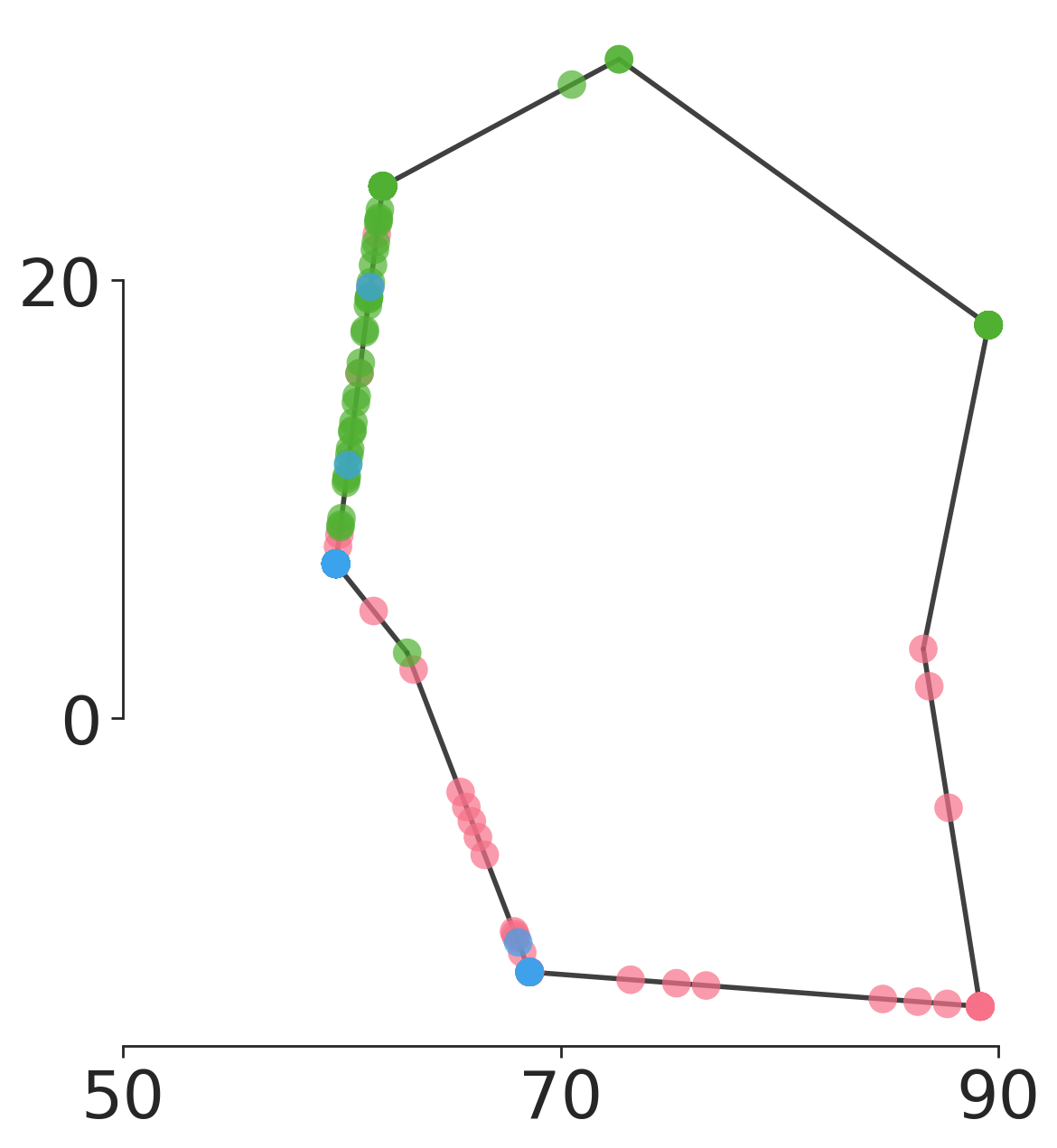}  
		\caption{Orthogonal projection of the embedded data onto the cycle representation.}
		\label{fig:diode_proj_pca}
	\end{subfigure}\hfill%
	\begin{subfigure}[t]{.34\linewidth}
		\centering
		\includegraphics[height=3.8cm]{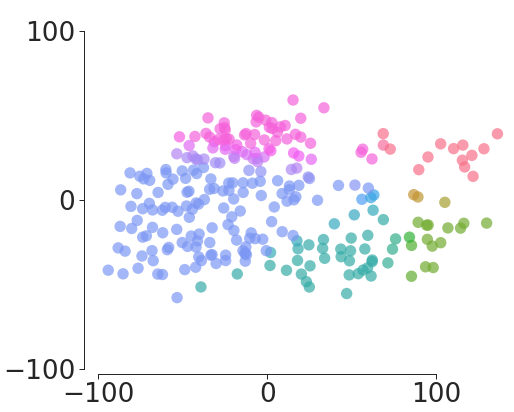}  
		\caption{The pseudotimes inferred from the projection in (b), quantified on a continuous color scale.}
		\label{fig:diode_pseudo_pca}
	\end{subfigure}
	\caption{Automated pseudotime inference of real cell cycle data through persistent homology, from the PCA embedding of the data.}
\end{figure}

\begin{figure}[t]
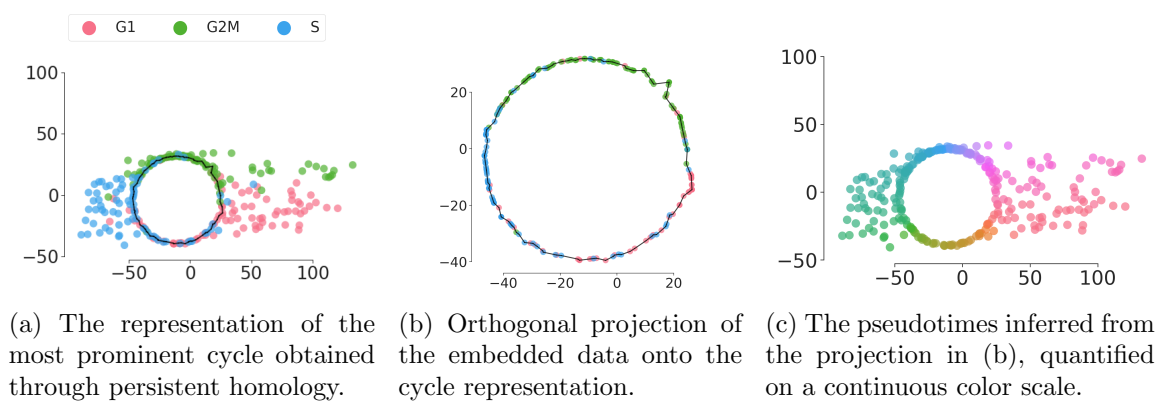

	\centering
	\begin{subfigure}[t]{.9\linewidth}
		\includegraphics[height=.4cm]{Images/CellCycle/CellCircleLegend}
	\end{subfigure}\\
	\begin{subfigure}[t]{.32\linewidth}
		\centering
		\includegraphics[height=3.3cm]{Images/CellCycle/CellCircle_diodeTimeCircle}  
		\caption{The representation of the most prominent cycle obtained through persistent homology.}
		\label{fig:diode_cycle_reg}
	\end{subfigure}
	\hfill
	\begin{subfigure}[t]{.3\linewidth}
		\centering
		\includegraphics[height=3.3cm]{Images/CellCycle/CellCircle_diodeTimeProjection}  
		\caption{Orthogonal projection of the embedded data onto the cycle representation.}
		\label{fig:diode_proj_reg}
	\end{subfigure}
	\hfill
	\begin{subfigure}[t]{.34\linewidth}
		\centering
		\includegraphics[height=3.3cm]{Images/CellCycle/CellCircle_diodeTimeCoordinates}  
		\caption{The pseudotimes inferred from the projection in (b), quantified on a continuous color scale.}
		\label{fig:diode_pseudo_reg}
	\end{subfigure}
	\caption{Automated pseudotime inference of real cell cycle data through persistent homology, from the topologically regularized PCA embedding of the data.}
	\label{fig:diode_reg}
\end{figure}

Analyzing single cell cycle data is a use case where prior topological information is available. As the signal-to-noise ratio is commonly low in high-dimensional expression data \citep{libralon2009pre, zhang2021noise}, this data is usually preprocessed through a dimensionality reduction method before automated pseudotime inference \citep{Cannoodt2016, Saelens276907}. Topological regularization provides a tool to enhance the desired topological signal during the embedding procedure, and as such, facilitates automated inference that depends on this signal-to-noise ratio.

To illustrate this, we used persistent homology for an automated (cell) cycle and pseudotime inference method, with and without topological regularization during the PCA embedding of the data.
For this experiment, we use the PCA and topologically regularized embeddings of the cell cycle data which has also been analyzed by \citet{buettner2015computational}.
Our automated pseudotime inference method consists of the following steps.
\begin{enumerate}
	\item First, a representation of the most prominent cycle in the embedding is obtained through persistent homology from the $\alpha$-filtration, using the Python software library Dionysus (\url{https://pypi.org/project/dionysus/}).
	It can be seen as a circular representation---discretized in edges between data points---of the point in the 1st-dimensional persistence diagram that corresponds to the most persisting cycle (Figures \ref{fig:diode_cycle_pca} \& \ref{fig:diode_cycle_reg}).
	\item An orthogonal projection of the embedded data onto the representative cycle is obtained.
	This is an intermediate step to derive continuous pseudotimes from a discretized topological representation, as earlier described by \citet{Saelens276907} (Figures \ref{fig:diode_proj_pca} \& \ref{fig:diode_proj_reg}).
	\item The lengths between consecutive points on the orthogonal projection are used to obtain circular pseudotimes between $0$  and $2\pi$ (Figures \ref{fig:diode_pseudo_pca} \& \ref{fig:diode_pseudo_reg}).
\end{enumerate}
Note that within the scope of the current paper, we do not advocate this to be a new and generally applicable automated pseudotime inference method for single-cell data following the cell cycle model. 
However, we chose this particular method because it illustrates well what persistent homology identifies in the data and what we aim to optimize through topological regularization.

For example, we observe that (the representation of) the most prominent cycle in the ordinary PCA embedding is rather spurious and mostly linearly separates G1 and G2M cells.
Projecting cells onto the edges of this cycle places most of the cells onto a single edge (Figure \ref{fig:diode_proj_pca}).
The resulting pseudotimes are continuous for cells projected onto this edge, whereas they are mainly discrete for all other cells (Figure \ref{fig:diode_pseudo_pca}).
However, by incorporating prior topological knowledge into the PCA embedding, the (representation of) the most prominent cycle in the embedding now better characterizes the transition between the G1, G2M, and S stages in the cell cycle model (Figure \ref{fig:diode_cycle_reg}).
The automated procedure for pseudotime inference also reflects a more continuous transition between the cell stages (Figures \ref{fig:diode_proj_reg} and \ref{fig:diode_pseudo_reg}).

\paragraph{Bifurcating Cell Trajectory Data}
We use the UMAP loss to illustrate the effect of topological regularization on the bifurcating model of this single-cell dataset. The topological loss is $\mathcal{L}_{\mathrm{top}}\equiv\mathcal{L}_{\mathrm{conn}}+\mathcal{L}_{\mathrm{flare}}$, where $\mathcal{L}_{\mathrm{conn}}(\gD_0) = \sum_{k=2}^\infty (d_k-b_k)$ measures the total (sum of) finite 0-dimensional persistence in the embedding to encourage connectedness of the representation and $\mathcal{L}_{\mathrm{flare}} = -\left[d_3-b_3\right]_{\mathcal{E}_{\mE}^{-1}]-\infty,0.75}$ optimizes for a `flare' with (at least) three clusters away from the embedding mean.

\begin{figure*}[t]
	\centering
	\begin{subfigure}[t]{.325\linewidth}
		\centering
		\includegraphics[width=\linewidth]{Images/CellBifurcating/CellBifUMAP}  
		\caption{Ordinary UMAP embedding.}
		\label{fig:CellBifUMAP}
	\end{subfigure}
	\begin{subfigure}[t]{.325\linewidth}
		\centering
		\includegraphics[width=\linewidth]{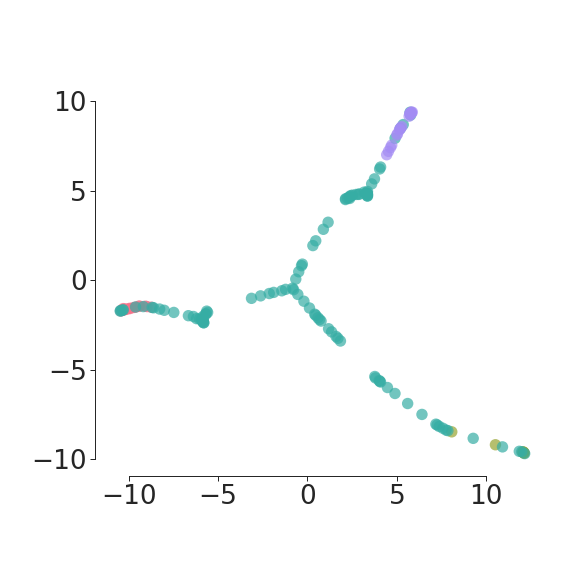}  
		\caption{Top.\ optimized embedding.}
		\label{fig:CellBifUMAPopt}
	\end{subfigure}
	\begin{subfigure}[t]{.325\linewidth}
		\centering
		\includegraphics[width=\linewidth]{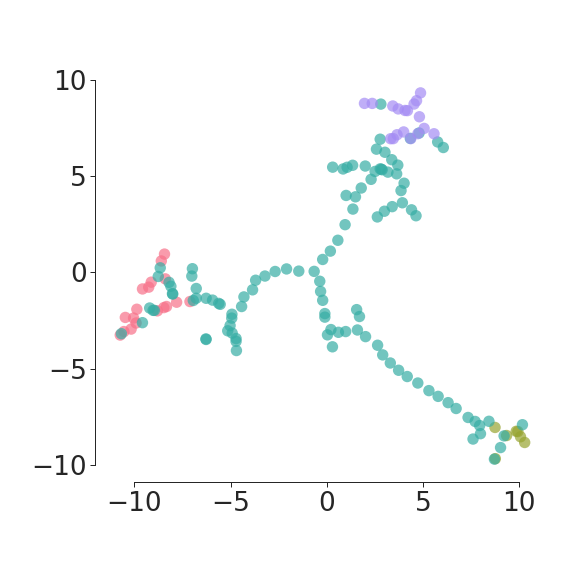}  
		\caption{Top.\ regularized embedding.}
		\label{fig:CellBifTop}
	\end{subfigure}
	\caption{Embeddings of the bifurcating cell data with colors representing the cell group.}
	\label{fig:CellBif}
\end{figure*}

We observe that while the ordinary UMAP embedding is more dispersed (Figure \ref{fig:CellBifUMAP}), the topologically regularized embedding is more constrained towards a connected bifurcating shape (Figure \ref{fig:CellBifTop}).
For comparison, we conducted topological optimization for the loss $\mathcal{L}_{\mathrm{top}}$ of the initialized UMAP embedding without the UMAP embedding loss.
The resulting embedding is now more fragmented (Figure \ref{fig:CellBifUMAPopt}). 
We thus see that topological optimization may also benefit from the embedding loss.
Both losses are needed: the topological loss to impose the topological prior knowledge, and the embedding loss to ensure the learned representation is faithful to the learned representation.

\begin{figure}[t]
	\begin{subfigure}[t]{.325\textwidth}
		\centering
		\includegraphics[width=\linewidth]{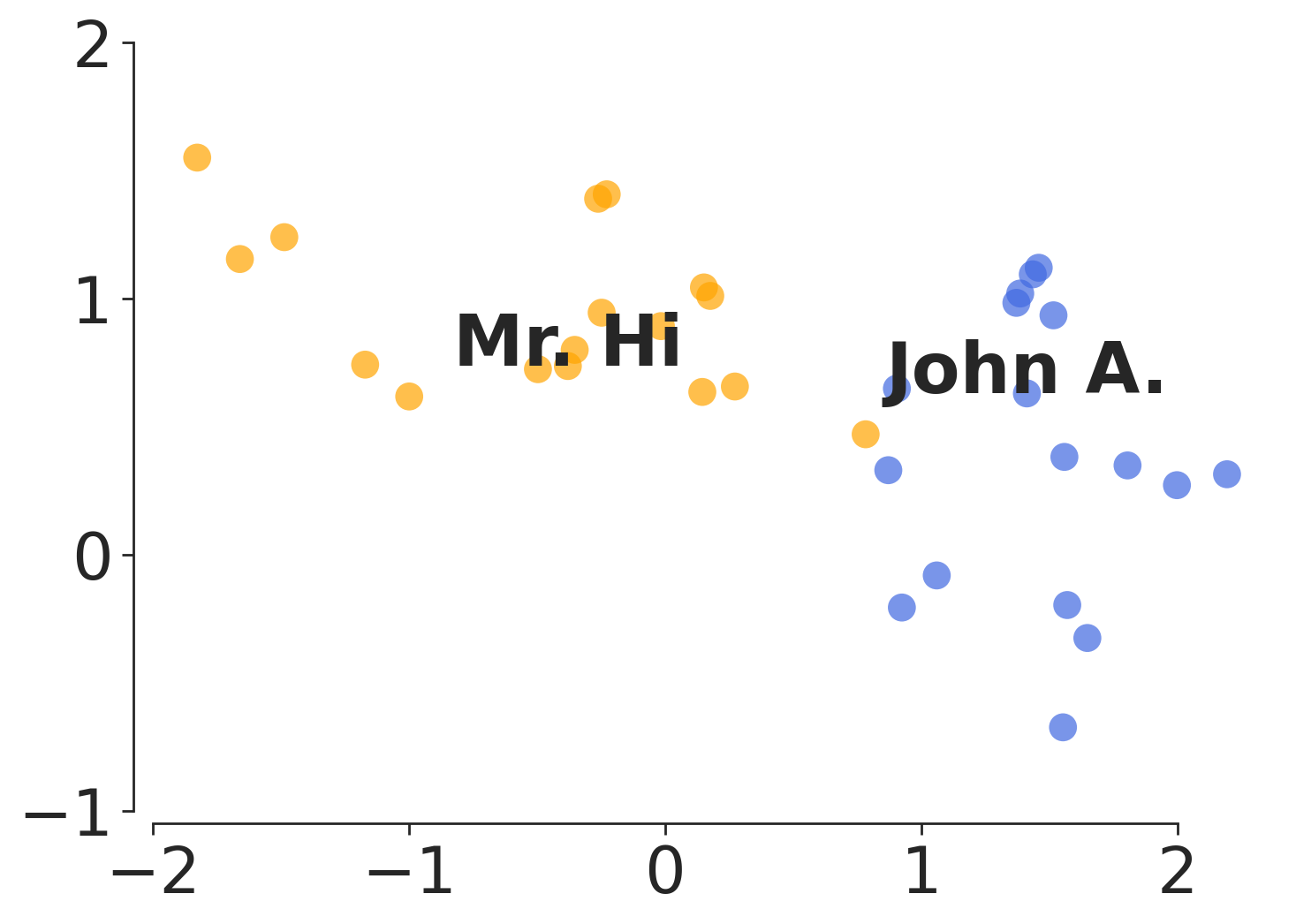}  
		\captionsetup{width=.8\linewidth}
		\caption{Ordinary DeepWalk embedding.}
		\label{KarateDW}
	\end{subfigure}
	\begin{subfigure}[t]{.325\textwidth}
		\centering
		\includegraphics[width=\linewidth]{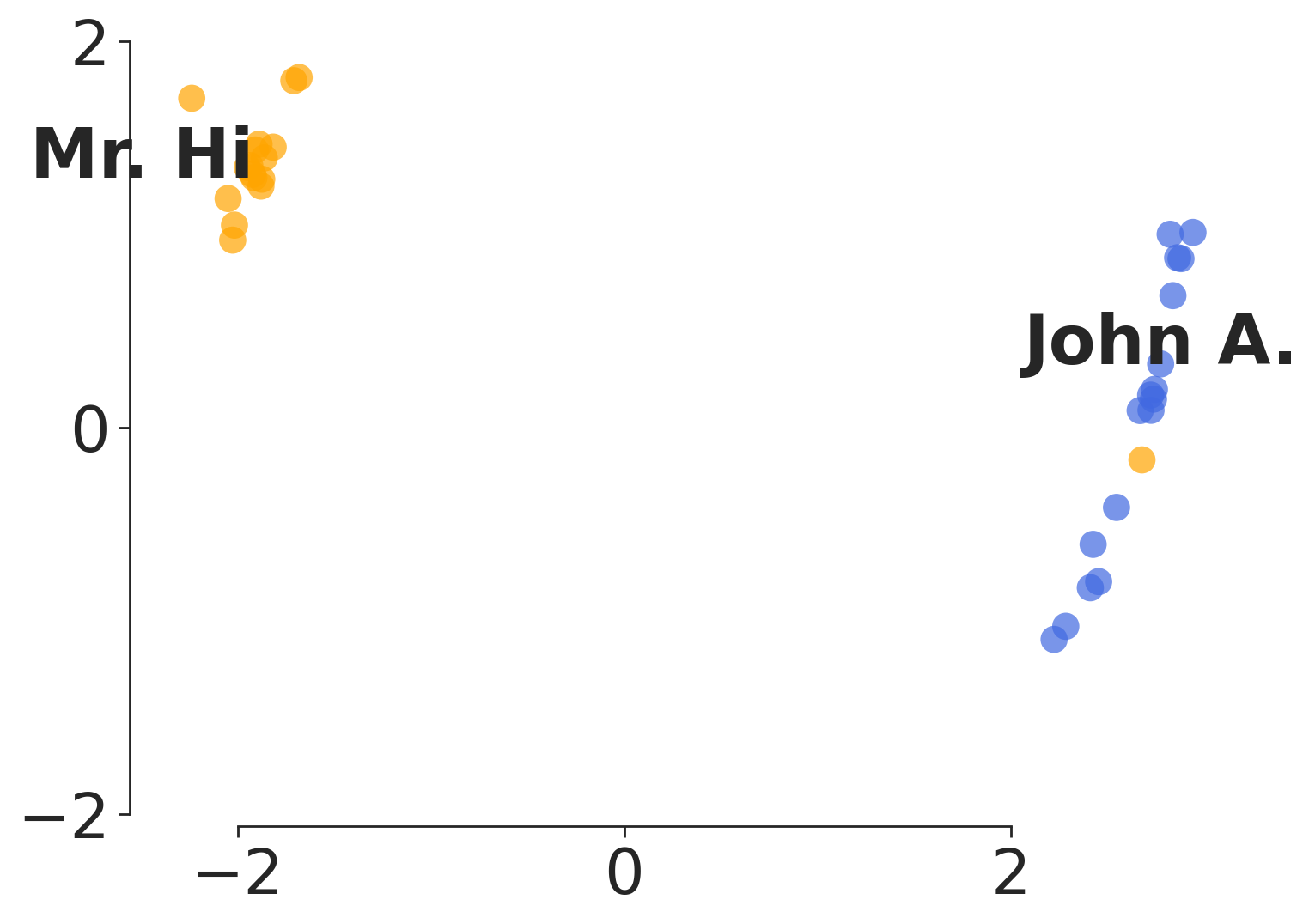}
		\caption{Top.\ optimized embedding.}
		\label{KarateDWopt}
	\end{subfigure}
	\begin{subfigure}[t]{.325\textwidth}
		\centering
		\includegraphics[width=\linewidth]{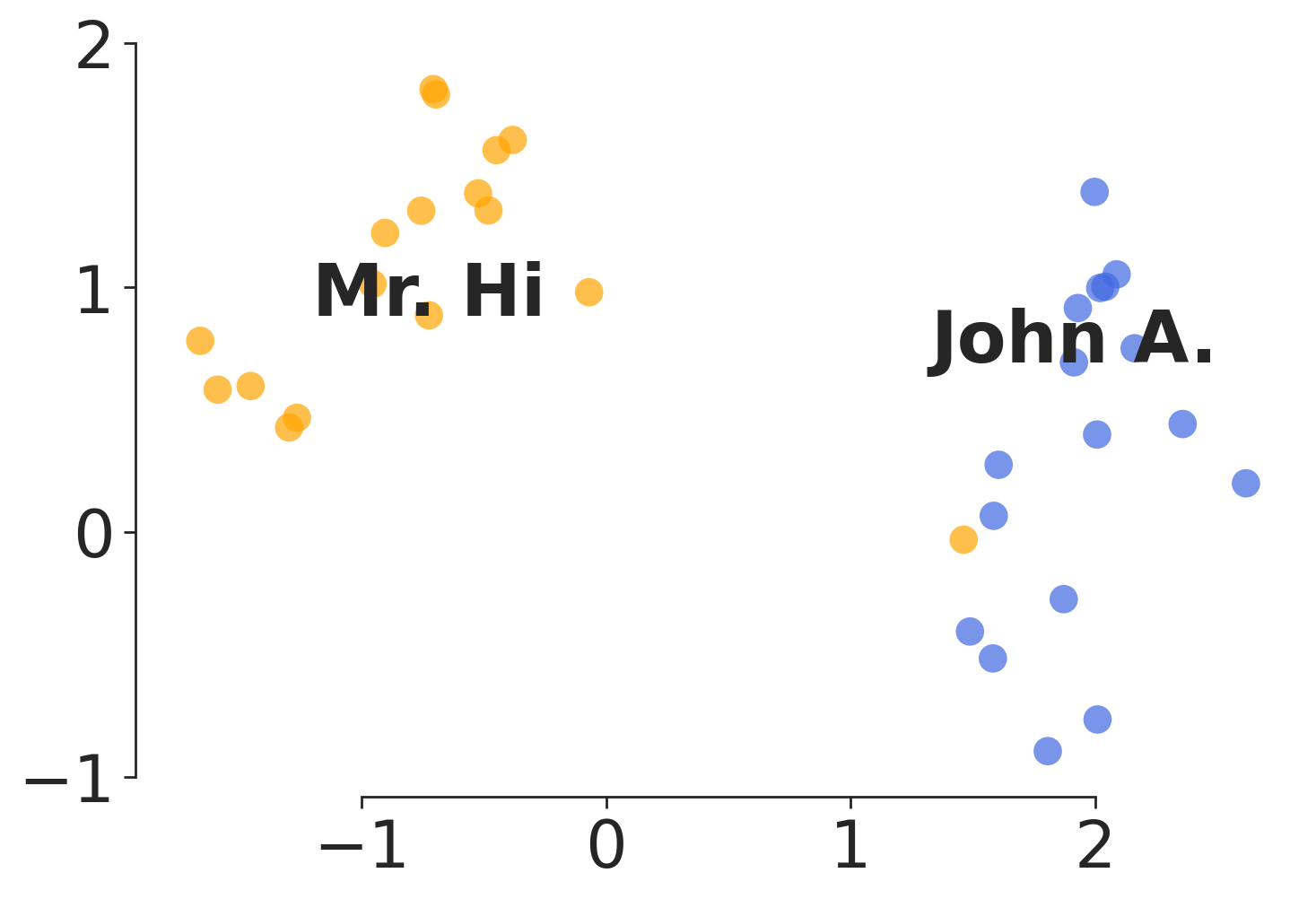}
		\caption{Top.\ regularized embedding.}
		\label{KarateTop}
	\end{subfigure}
	\caption{The Karate network and various of its embeddings.}
\end{figure}

\paragraph{Karate}
The Karate network consists of two different communities represented by their key figures (John A.\ and Mr.\ Hi), as shown in Figure \ref{fig:karate}. 
To embed the graph, we used a DeepWalk variant adapted from \citet{graph_nets}.
While the ordinary DeepWalk embedding (Figure \ref{KarateDW}) well respects the ordering of points according to their communities, the communities remain close to each other.
We thus regularized this embedding for (at least) two clusters using the topological loss with sampling as defined by (\ref{sampleloss}), where $\mathcal{L}_{\mathrm{top}}(\gD_0) = -(d_2 - b_2)$ measures the persistence of the second most prominent 0-dimensional hole, and $f_{\mathcal{S}}=0.25$, $n_{\mathcal{S}}=10$.

The resulting embedding (Figure \ref{KarateTop}) now nearly perfectly separates the two ground truth communities present in the graph.
Further optimizing the ordinary DeepWalk embedding with the same topological loss but without the DeepWalk loss, results in a larger separation of the two communities (Figure \ref{KarateDWopt}). Compared to the regularized embedding, the visible structure within the two communities is reduced, again showing the importance of including both embedding loss and topological loss.

\paragraph{Harry Potter}
\label{SEC::HarryPotter}

\begin{figure*}[t]
	\centering
	\begin{subfigure}[t]{.30\textwidth}
		\centering
		\includegraphics[width=\linewidth]{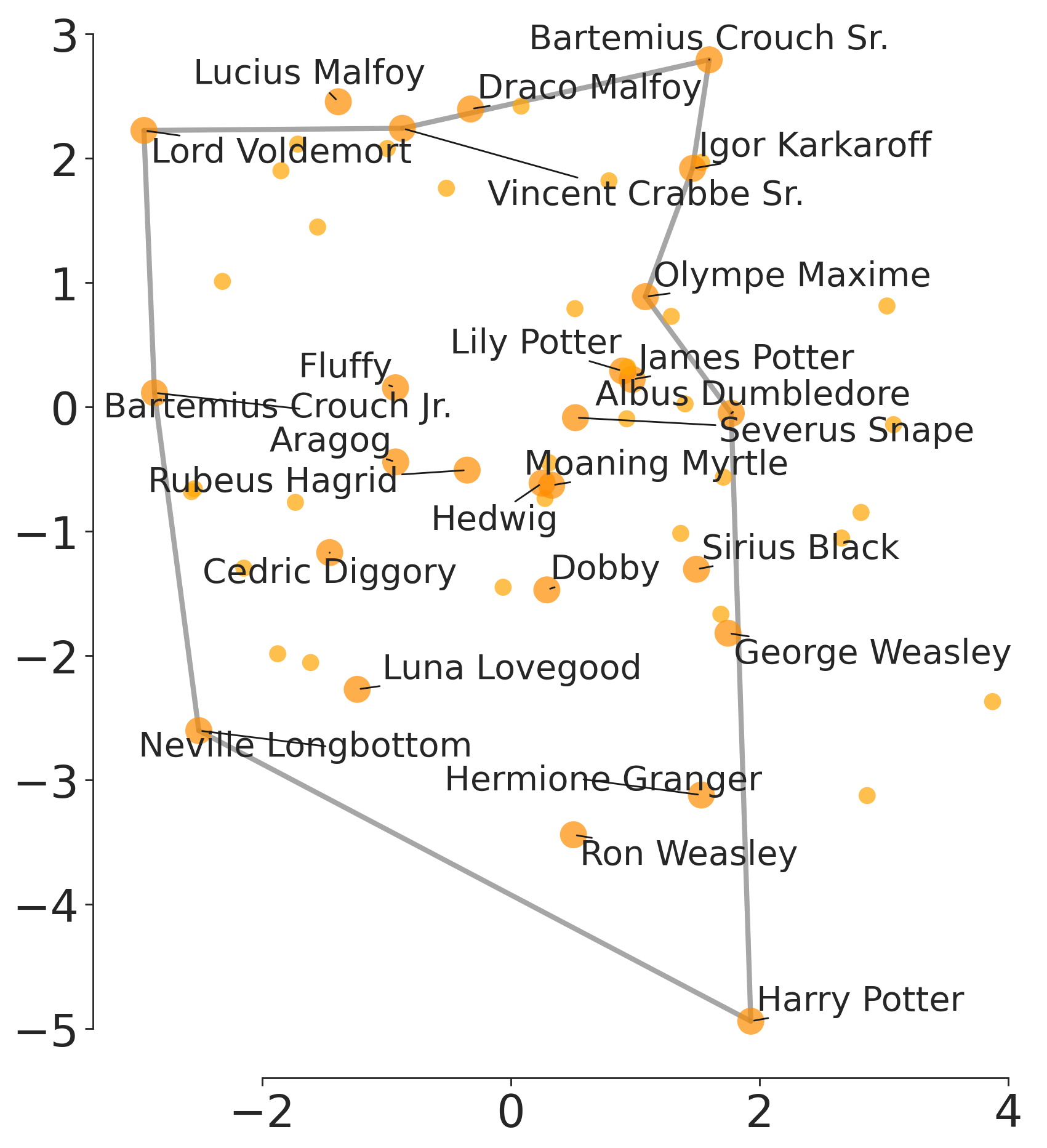}  
		\caption{Ordinary graph embedding.}
		\label{HarryIP}
	\end{subfigure}\hfill
	\begin{subfigure}[t]{.325\textwidth}
		\centering
		\includegraphics[width=\linewidth]{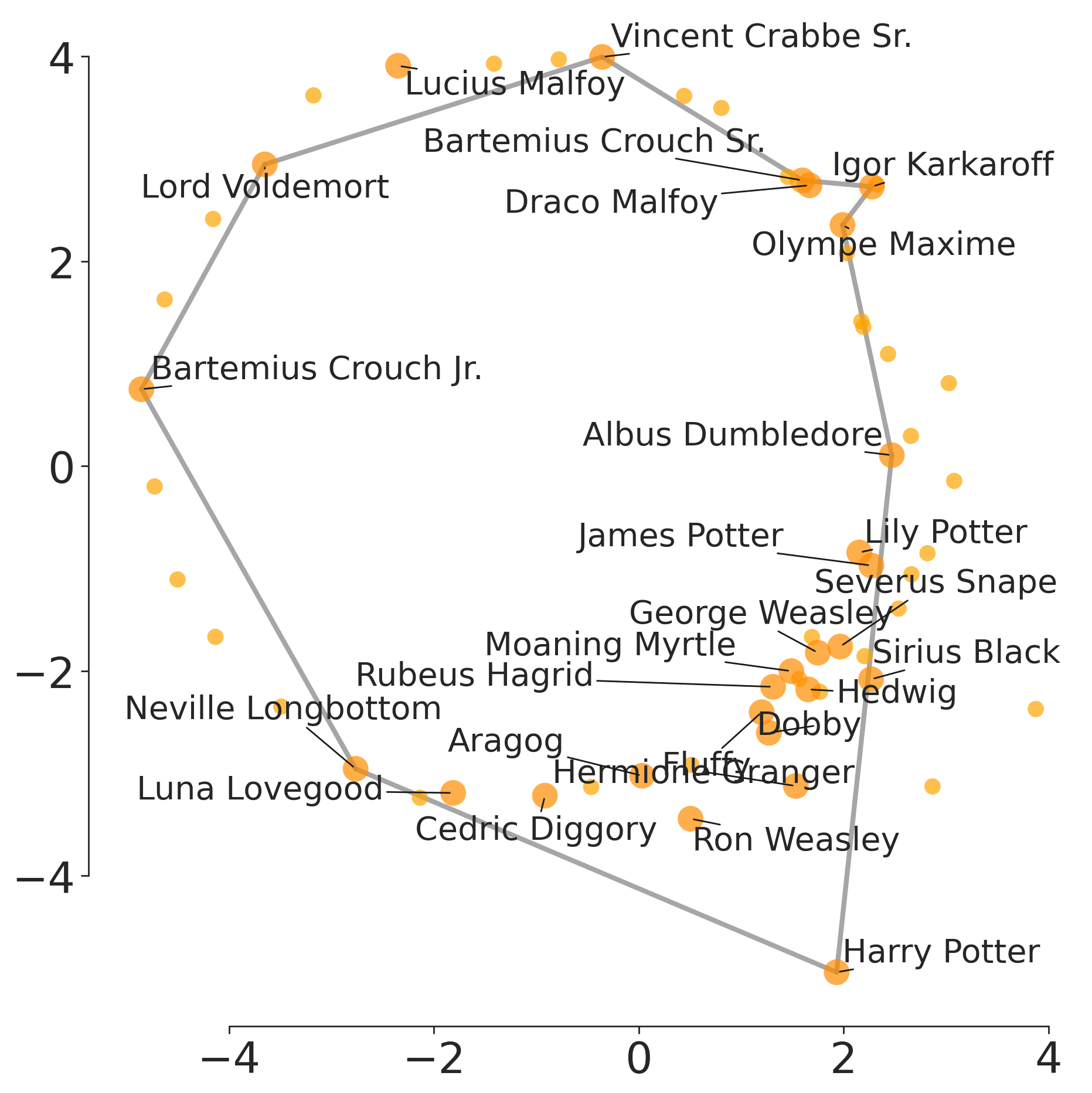}  
		\caption{Topologically optimized embedding (initialized with the ordinary graph embedding).}
		\label{HarryIPopt}
	\end{subfigure}\hfill
	\begin{subfigure}[t]{0.35\textwidth}
		\centering
		\includegraphics[width=\linewidth]{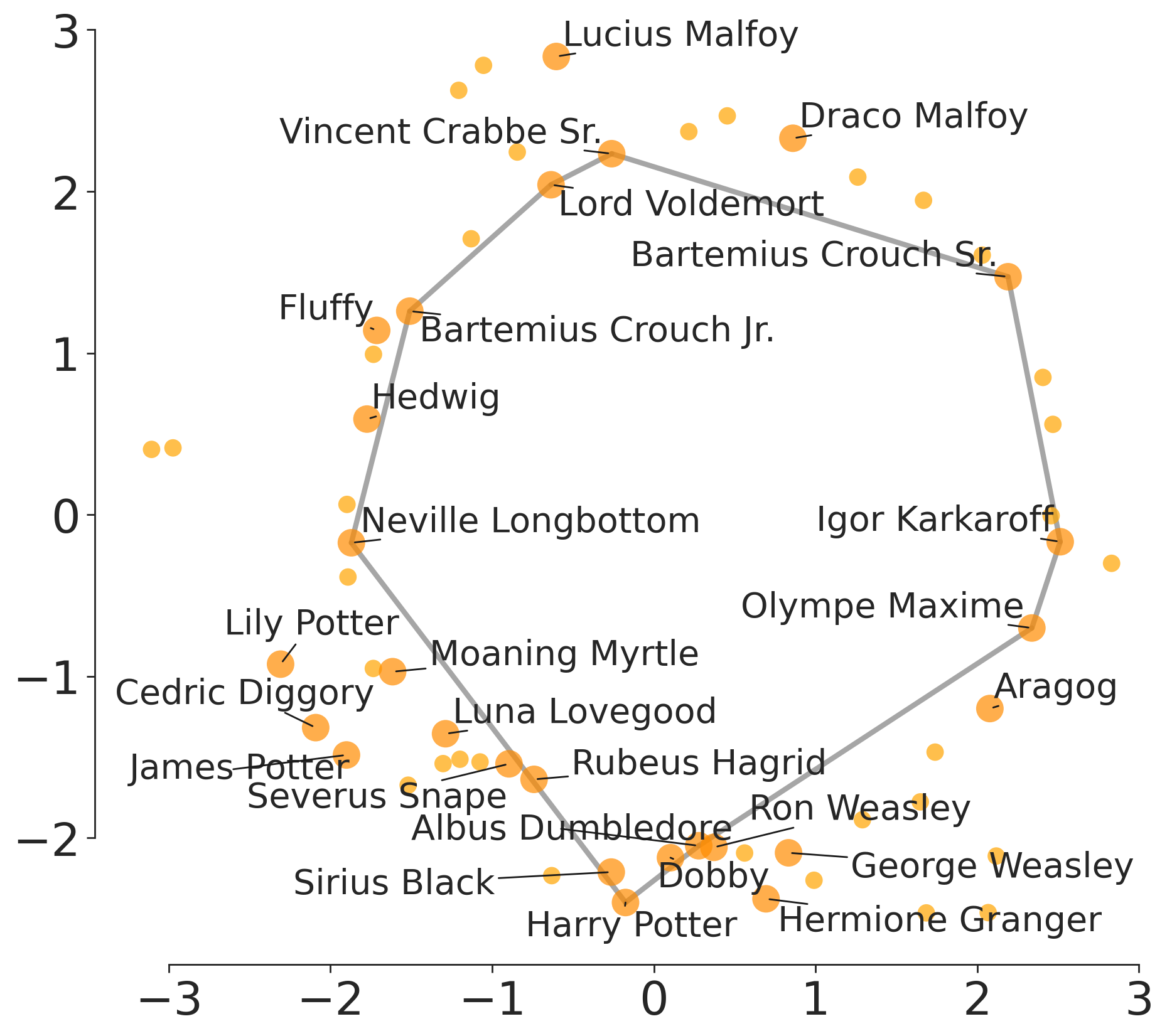}  
		\caption{Topologically regularized embedding.}
		\label{HarryTop}
	\end{subfigure}
	\caption{Various embeddings of the Harry Potter graph and the circular model therein.}
	\label{HPembed}
\end{figure*}

To embed the Harry Potter graph, we used a simple graph embedding model where the sigmoid of the inner product between embedded nodes quantifies the (Bernoulli) probability of an edge occurrence \citep{rendle2020neural}.
Thus, this probability will be high for nodes close to each other in the embedding, and low for more distant nodes. These probabilities are then optimized to match the binary edge indicator vector.
Figure \ref{HarryIP} shows the result of this embedding, along with the circular model presented by \citet{JMLR:v21:19-1032}.
For clarity, character labels are only annotated for a subset of the nodes (the same as by \citet{JMLR:v21:19-1032}). 

To better represent the circular model that transitions between the `good` and `evil` characters, we regularized the embedding using a topological loss function $\mathcal{L}_{\mathrm{top}}(\gD_1) = -(d_1-b_1)$ that measures the persistence of the most prominent 1-dimensional hole.
The resulting embedding is shown in Figure \ref{HarryTop}.
Interestingly, the topologically regularized embedding now better captures the circularity of the model identified by \citet{JMLR:v21:19-1032}, and focuses more on distributing the characters along it.
Note that although this previously identified model is included in the visualizations, it is not used to derive the embeddings, nor is it derived from them.
For comparison, Figure \ref{HarryIPopt} shows the result of optimizing the ordinary graph embedding (used as initialization) for the same topological loss, but without the graph embedding loss.

We conclude that topological regularization does successfully impose the specified structure in the embeddings.

\subsubsection{Quantitative Evaluation}
\label{quantitative}

In this section we analyze the embedding and topological losses from the previous experiments and investigate Q2 by quantifying changes in prediction performance on the different embeddings.  

Table \ref{losstable} summarizes the losses we obtained for the ordinary embeddings, the topologically optimized embeddings (initialized with the ordinary embeddings, but not using the embedding loss), as well as for the topologically regularized embeddings.
As one would expect, topological regularization balances the embedding losses between the embedding losses of the ordinary and topologically optimized embeddings.
We also observe that there are more significant differences in the obtained topological losses than in the embedding losses with and without regularization.
This suggests that the optimum region for the embedding loss may be somewhat flat with respect to the corresponding region for the topological loss. 
Thus, slight shifts in the local embedding optimum by topological regularization may result in much better topological embedding models.

We evaluated the quality of the embedding visualizations presented in Section \ref{sec:effec_qualitative}, by assessing how informative they are for predicting ground truth data labels.
For the Synthetic Cycle data, these labels are the 2D coordinates of the noise-free data on the unit circle in $\mathbb{R}^2$, and we used a multi-output support vector regressor model.
For the cell trajectory data and Karate network, we used the ground truth cell groupings and community assignments, respectively, and a support vector machine model.  
The points in the 2D embeddings were then split into 90\% points for training and 10\% for testing with exception of the synthetic cycle dataset, where we used an 80/20 split.
Consecutively, we used 5-fold CV on the training data to tune the regularization hyperparameter $C\in\{1\mathrm{e}-2, 1\mathrm{e}-1, 1, 1\mathrm{e}1, 1\mathrm{e}2\}$.
Other settings were the default from \textsc{scikit-learn}.
The performance of the final tuned and trained model was then evaluated on the test data, through the $r^2$ coefficient of determination for the regression problem, and the accuracy for all classification problems. 

The averaged test performance metrics and their standard deviations, obtained over 100 random train-test splits, are summarized in Table \ref{quanttable}.
We observe that quantitative differences between the ordinary, optimized, and regularized embeddings are small. However, aligning the embedding with the expected topology does not seem to decrease the prediction performance.

\begin{table}[tp]
	\renewcommand\arraystretch{1.2}
	\centering
	\begin{tabular}{lccc|rrr}
		\toprule
		 & \multicolumn{3}{c}{Embedding loss} & \multicolumn{3}{c}{Topological loss}\\
		Data & ord. & top.\ opt. & top.\ reg. & ord. & top.\ opt. & top.\ reg. \\
		\midrule
		Synthetic Cycle & $\bf{0.0632}$ & $0.0639$ & $0.0634$ & $-0.42$ & $\bf{-1.63}$ & $-1.38$ \\
		Cell Cycle & $\bf{6.70}$ & $7.00$ & $6.78$ & $-10.18$ & $\bf{-68.34}$ & $-63.52$ \\
		Cell Bifurcating & $\bf{8496}$ & $9865$ & $8896$ & $116.03$ & $\bf{24.03}$ & $63.90$ \\
		Karate & $\bf{2006}$ & $3239$ & $2112$ & $-0.59$ & $\bf{-4.58}$ & $-1.82$ \\
		Harry Potter & $\bf{0.20}$ & $4.25$ & $0.23$ & $-0.82$ & $\bf{-5.84}$ & $-3.05$ \\
		\bottomrule
	\end{tabular}
	\caption{Embedding/reconstruction and topological losses of the final embeddings.
	Lowest in bold. If the topological loss function was computed on random subsets of the data, we report the loss on the full dataset.}
	\label{losstable}
\end{table}

\begin{table}[tp]
	\renewcommand\arraystretch{1.2}
	\centering
	\begin{tabular}{llccc}
		\toprule
		Data & Metric & Ord.\ emb. & Top.\ opt. & Top.\ reg. \\
		\midrule
		Synthetic Cycle & $r^2$ & $0.62\pm0.15$ & $0.60\pm0.14$ & $\bf{0.67\pm0.20}$\\ %
		Cell Cycle & accuracy & $\bf{0.79\pm0.07}$ & $\bf{0.79\pm0.07}$ & $0.78\pm0.07$  \\
		Cell Bifurcating & accuracy & $0.78\pm0.08$ & $\bf{0.85\pm0.07}$ & $0.82\pm0.08$ \\
		Karate & accuracy & $\bf{0.97\pm0.08}$ & $\bf{0.97\pm0.08}$ & $\bf{0.97\pm0.08}$\\
		\bottomrule
	\end{tabular}
	\caption{Embedding performance evaluations for label prediction.
	Highest in bold.}
	\label{quanttable}
\end{table}

\subsubsection{Scalability}
\begin{figure}[tp]
	\centering
	\includegraphics[width=.4\textwidth]{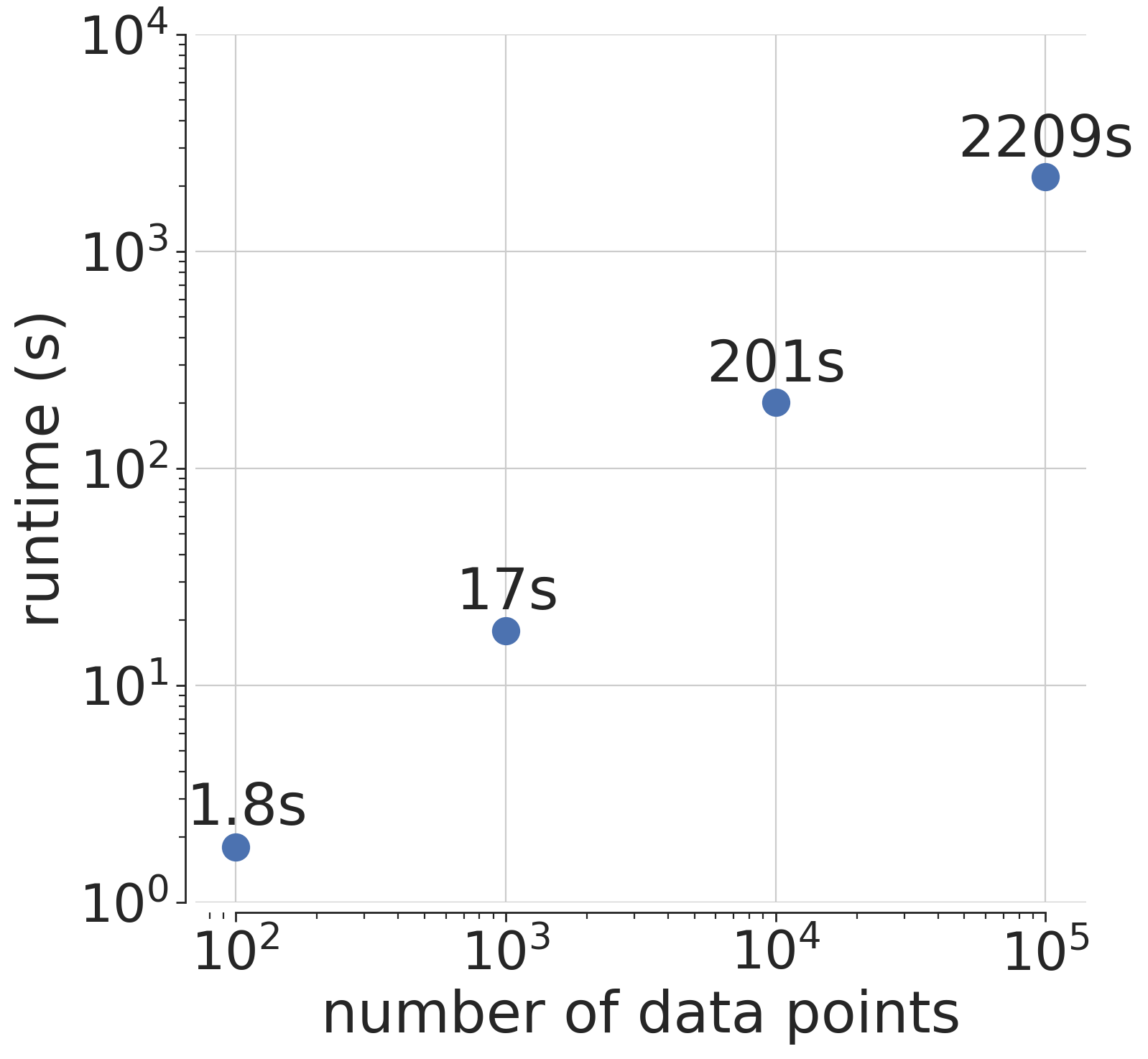}
	\caption{Log-log plot showing dataset size vs. runtime to optimize for a one-dimensional hole for 100 iterations without sampling.}
	\label{fig:runtime}
\end{figure}

To illustrate the empirical runtime of topological regularization, we optimized a random 2-dimensional dataset of size $n$ using a circle loss but no embedding loss for 100 iterations (learning rate $0.01$). A circle is computationally the most challenging topological hole to optimize in a 2D embedding.
The experiments have been carried out on a Dell Latitude 5511 laptop equipped with an Intel(R) Core(TM) i7-10850H 2.70GHz processor, with 16.0 GB of RAM. 
We show the runtime without sampling in a log-log plot in Figure \ref{fig:runtime}. 
Computing the topological loss using sampling with $f_{\mathcal{S}} = 0.1$ and $n_{\mathcal{S}} = 10$, takes $1.9, 17, 191, \text{and } 2284$ seconds. This is comparable to the runtime for 100 iterations without sampling, but we might need fewer iterations to reach a similar result. This is because the topological loss for a one-dimensional hole has nonzero gradients only for $4$ points (see Section \ref{SUBSUBSEC::pointcloudopt}). By aggregating the loss for $n_{\mathcal{S}}$ subsets, the gradient of the topological loss can affect up to $n_\mathcal{S}\cdot 4$ points. 

\subsection{Robustness of Topological Regularization}
\label{sec:robustness}

In this section we provide more intuition about the limits of topological regularization and seek to answer the following questions:

\begin{description}
	\item[Q4.] How does the precise formulation of the topological loss affect the final embedding? 
	\item[Q5.] How does a weak signal to noise ratio affect the optimization?
	\item[Q6.] What is the effect of regularizing with wrong prior information?
\end{description}

An overview of the topological loss functions that regularize the embeddings presented in this section is given in Table \ref{tab:robustness_loss}. 

\begin{table*}[t]
	\centering
	\renewcommand\tabcolsep{5pt}
	\renewcommand\arraystretch{1.5}
	\begin{tabular}{p{2.8cm}lllp{1.3cm}}
		\toprule
		Data & Fig. & Topological loss function & Dim & $f_{\mathcal{S}}, n_{\mathcal{S}}$ \\
		\midrule
		Synthetic Cycle & \ref{fig:robustness_sampling} & $-(d_1-b_1)$ & 1 & $0.25, 4$\newline $0.5, 2$\newline $1, 1$\\
		Random \newline Synthetic Cycle & \ref{fig:RandTop}, \ref{fig:robustness_zscore} & $-(d_1-b_1)$ & 1 & $0.25, 4$ \\
		Synthetic Cycle & \ref{fig:SynthCircle_secondcircle} & $-(d_2-b_2)$ & 1 & \textcolor{gray}{NA} \\
		Synthetic Cycle & \ref{fig:SynthCircle_connected} & {\small $\sum_{k=2}^\infty d_k$ } & 0 & \textcolor{gray}{NA} \\
		Cell Bifurcating & \ref{fig:bifotherp} & {\small $\sum_{k=2}^\infty (d_k-b_k)^{p_{\text{CC}}}-\left[(d_3-b_3)^{p_{\text{Flare}}}\right]_{\mathcal{E}_{\mE}^{-1}]-\infty,0.75]}$} & 0 - 0 & \textcolor{gray}{NA} \\
		Cell Bifurcating & \ref{CellBifTau} & {\small $\sum_{k=2}^\infty (d_k-b_k)-\left[d_3-b_3\right]_{\mathcal{E}_{\mE}^{-1}]-\infty,\tau]}$ } & 0 - 0 & \textcolor{gray}{NA} \\
		Cell Bifurcating & \ref{fig:CellBifCircleStrong}  & $-(d_1-b_1)$  & 1 & $0.25, 10$  \\
		\bottomrule
	\end{tabular}
	\caption{Summary of the topological losses used to regularize embeddings in Section \ref{sec:robustness}. Note that for 0-th dimensional homology diagrams $d_1=\infty$.}
	\label{tab:robustness_loss}
\end{table*}

\subsubsection{Varying the Parameters of Topological Priors\label{SEC:robust_params}}
In Section \ref{SEC::topreg} we introduced topological loss functions with sampling (see Equation~(\ref{sampleloss})) and for the flare shape (see Equation~(\ref{functionalloss})). Here we investigate how topological regularization reacts to different choices of the hyperparameters that are used in these loss functions.
We vary the sampling fraction $f_{\mathcal{S}}$ and number of repeats $n_{\mathcal{S}}$ in Equation~(\ref{sampleloss}), the functional threshold $\tau$ in Equation~(\ref{functionalloss}), and the function to evaluate the points in the persistence diagram.
Note that although changing these parameters may change the \emph{geometrical} characteristics that are specified, the \emph{topological} characteristics remain roughly the same with respect to the chosen topological loss function.

\begin{figure*}[t]
	\centering
	\begin{subfigure}[t]{.325\textwidth}
		\centering
		\includegraphics[width=\linewidth]{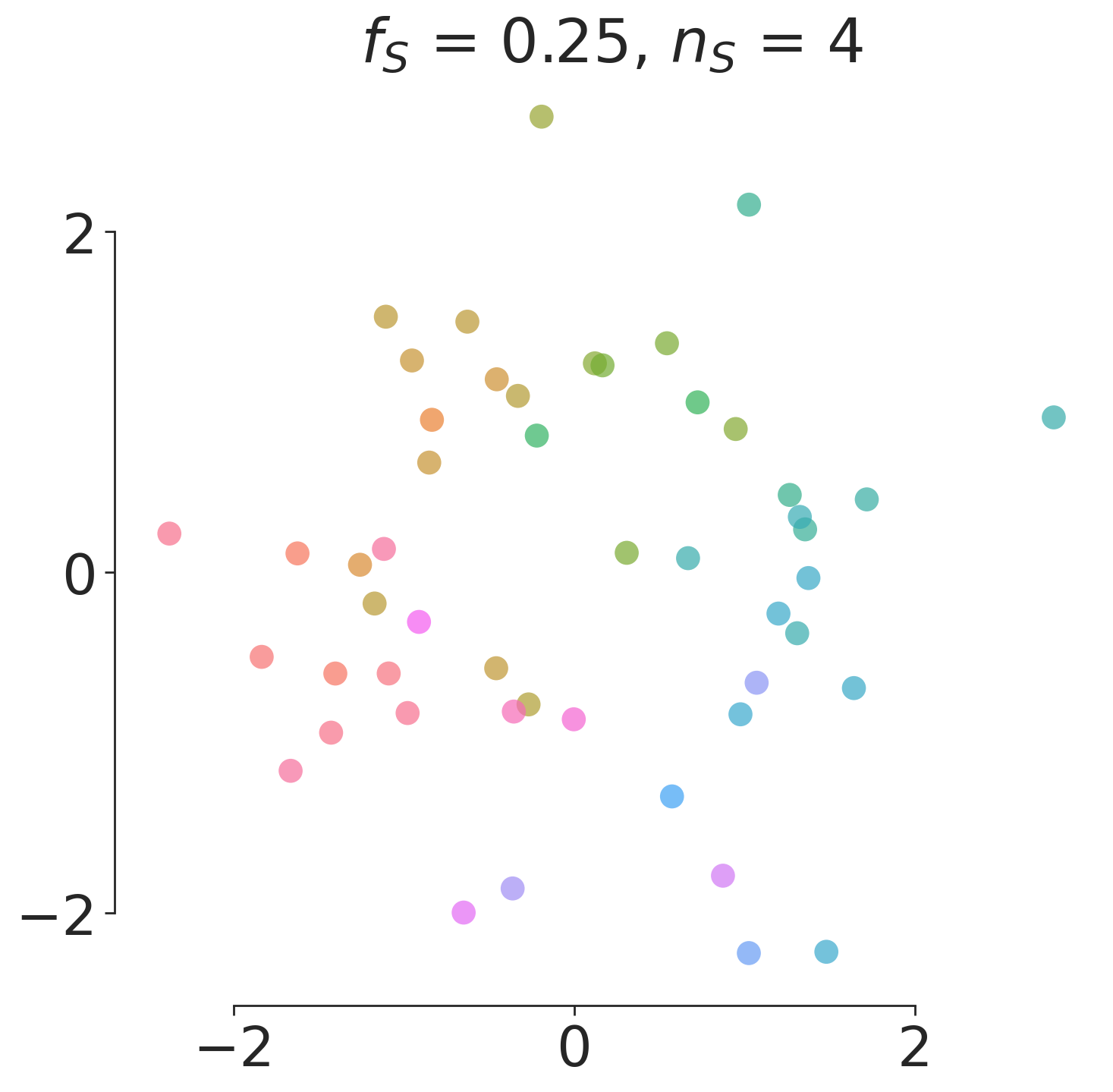}  
	\end{subfigure}\hfill
	\begin{subfigure}[t]{.325\textwidth}
		\centering
		\includegraphics[width=\linewidth]{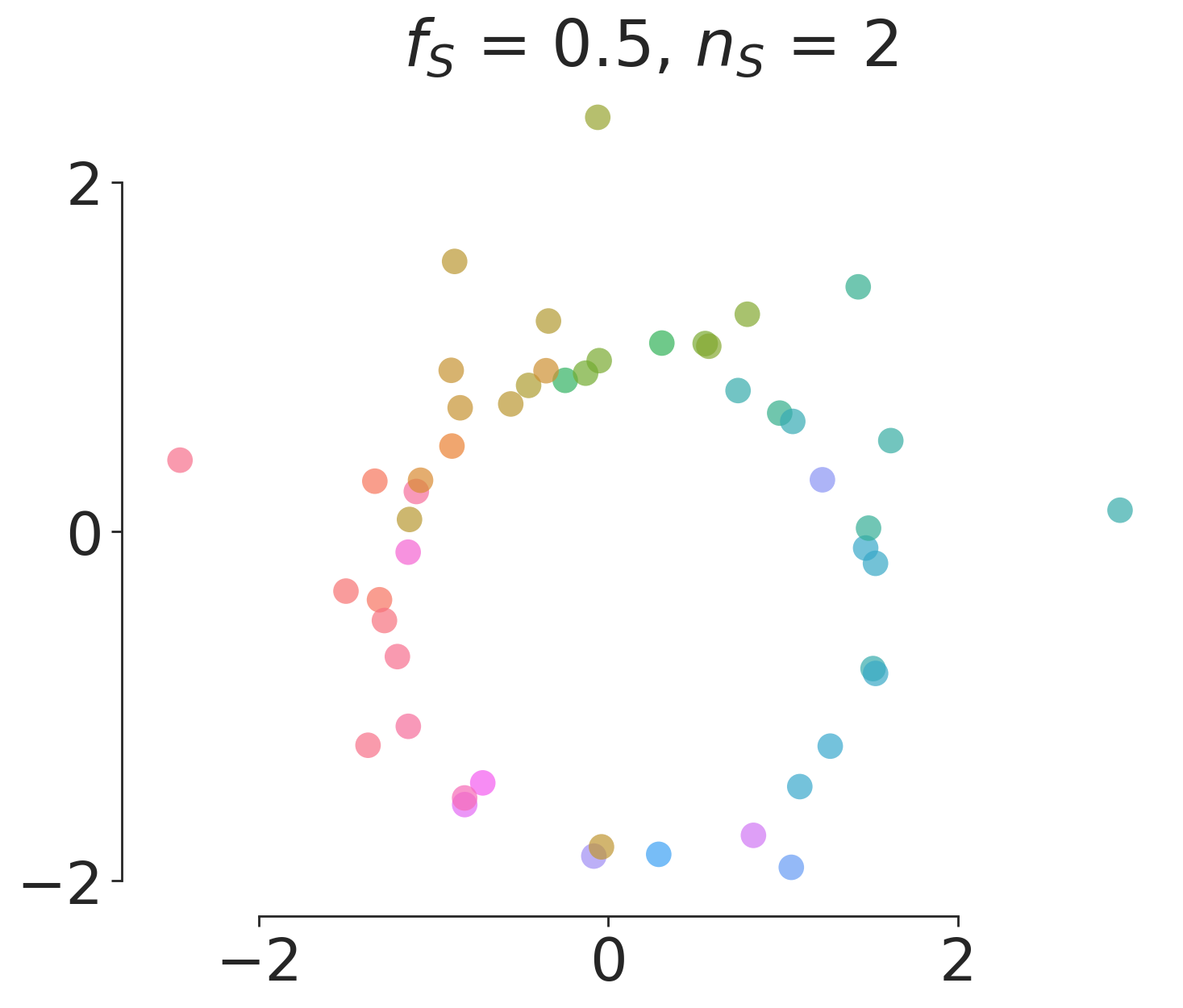}  
	\end{subfigure}\hfill
	\begin{subfigure}[t]{.325\textwidth}
		\centering
		\includegraphics[width=\linewidth]{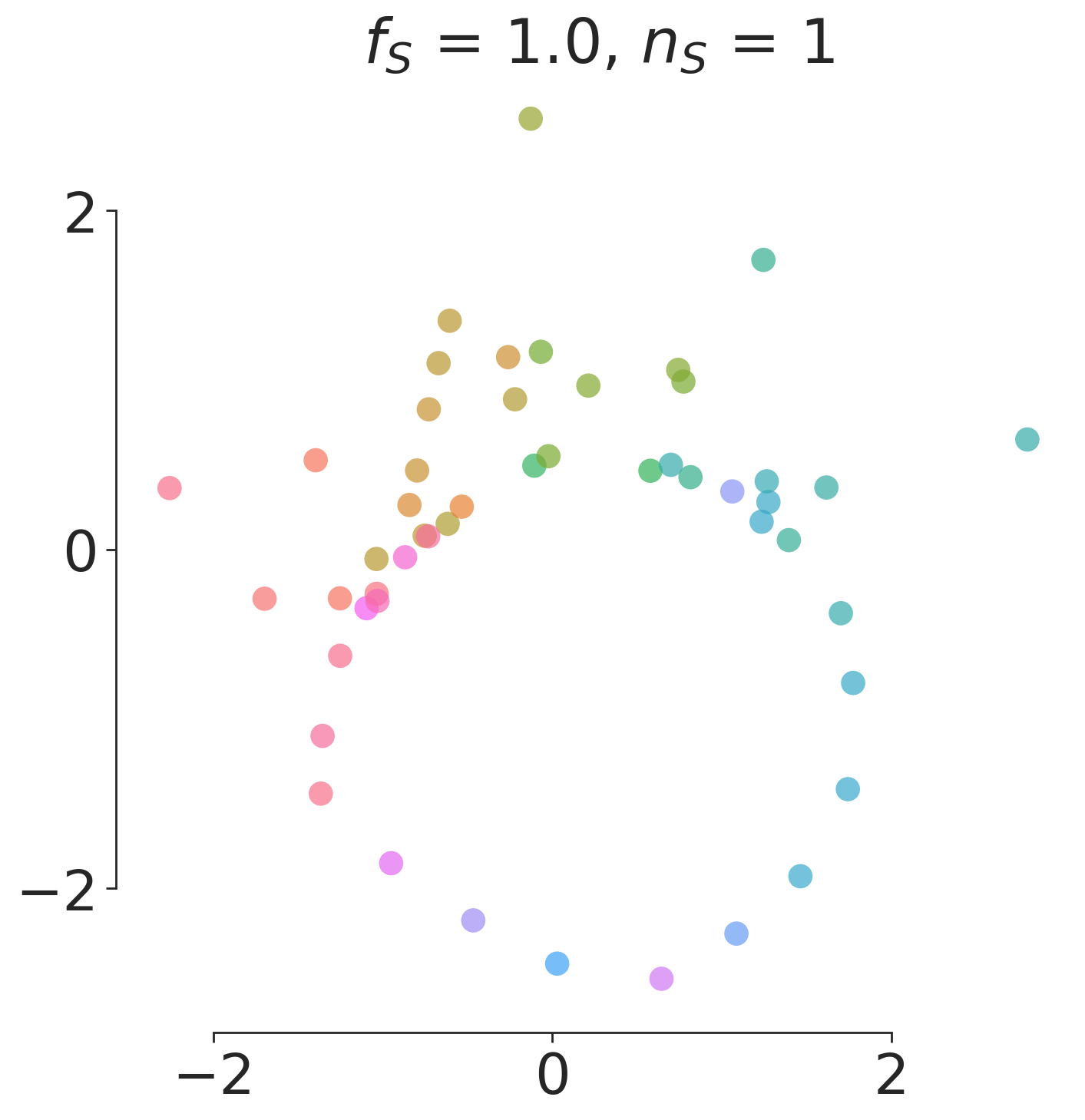}  
	\end{subfigure}
	\caption{Topologically regularized PCA embeddings of the synthetic cycle dataset for varying sampling fraction $f_{\mathcal{S}}$ and number of repeats $n_{\mathcal{S}}$.}
	\label{fig:robustness_sampling}
\end{figure*}

\paragraph{Varying $f_{\mathcal{S}}$ and $n_{\mathcal{S}}$}
Figure \ref{fig:robustness_sampling} shows the topologically regularized PCA embedding of the synthetic cycle for varying choices of the values $f_{\mathcal{S}}$ and $n_{\mathcal{S}}$.
We again optimize for a one-dimensional hole (see Table \ref{tab:robustness_loss}) and use the same parameter setting as before, stated in Table \ref{tab:effectiveness_parameters}. We chose the number of repeats $n_{\mathcal{S}}$ such that computing the embeddings requires approximately the same runtime.
We observe that by increasing $f_{\mathcal{S}}$, the hole tends to become more defined and no points are inside the boundary.
More specifically, one can better visually deduce a subset of the data that represents and lies on a nearly perfect circle in the Euclidean plane.
Nevertheless, we also notice that without sampling ($f_{\mathcal{S}} = 1$), the embedded circle does not align perfectly with the ground truth and only part of the data lies on it.
We conclude that the sampling fraction should be small enough to avoid optimizing spurious holes but large enough to ensure the $d$-dimensional features of interest are present and stable. In this example, regularizing with $f_{\mathcal{S}} = 0.25$ is not stable, as the largest hole in the embedding will change depending on whether the two points in the center are in the sample or not.

\paragraph{Different Loss Functions for the same Topological Prior (varying $\bm{g}$)}
\label{PARAGRAPH::vary_g}
For one particular type of prior topological information, one can design different topological loss functions by varying the real-valued function $g$ evaluated on the persistence diagram points that is used in the loss function (\ref{eq:topo_loss}).
In particular, our topological loss function (\ref{eq:kdim-hole-loss}), where we let $g\colonequiv b_k-d_k$, is inspired by the topological loss function
\begin{equation}
	\label{fromloss}
	\mathcal{L}_{\mathrm{top}}(\mE)\coloneqq\mathcal{L}_{\mathrm{top}}(\mathcal{D})=\mu\sum_{k=i, d_k<\infty}^{|\mathcal{D}^{\mathrm{reg}}|}\left(d_k-b_k\right)^p\left(\frac{d_k+b_k}{2}\right)^q,\hspace{2em}\mbox{ where } d_1-b_1\geq d_2-b_2\geq\ldots,
\end{equation}
introduced by \citet{gabrielsson2020topology}, and more formally investigated within an algebraic setting by \citet{adcock2013ring}.
Here, $p$ and $q$ are hyperparameters that control the strength of penalization, whereas $i$ and the choice of persistence diagram (or homology dimension) is used to postulate the topological information of interest.
The choice of $\mu\in\{1,-1\}$ determines whether one wants to increase ($\mu=-1$) or decrease ($\mu=1$) the topological information for which the topological loss function in the data is designed.
The term $(d_k-b_k)^p$ can be used to control the prominence, i.e., the persistence, of the most prominent topological holes, whereas the term $\left(\frac{d_k+b_k}{2}\right)^q$ can be used to control whether prominent topological features persist either early or later in the filtration.

Our topological loss function (\ref{eq:kdim-hole-loss}) is identical to (\ref{fromloss}) with $p=1$ and $q=0$, and the additional change that we sum to the hyperparameter $j$ instead of $|\mathcal{D}^{\mathrm{reg}}|$, which allows for even more flexible prior topological information. 
Note that this is one of the most simple and intuitive, yet flexible, manners to postulate topological loss functions.
Indeed, $(d_k-b_k)^{p=1}$ directly measures the persistence of the $k$-th most prominent topological hole for a chosen dimension of interest.
The role of $q\neq 0$ is to be further investigated within the context of topological regularization and formulating prior topological information.

\begin{figure*}[tp]
	\centering
	\begin{subfigure}[t]{.325\textwidth}
		\centering
		\includegraphics[width=\linewidth]{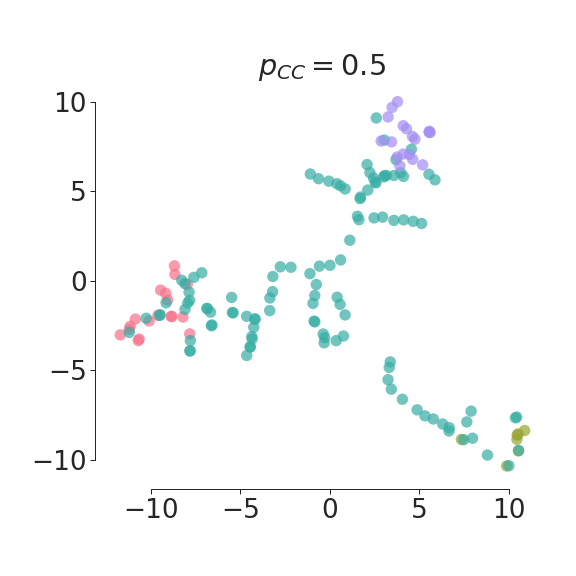}  
	\end{subfigure}\hfill
	\begin{subfigure}[t]{.325\textwidth}
		\centering
		\includegraphics[width=\linewidth]{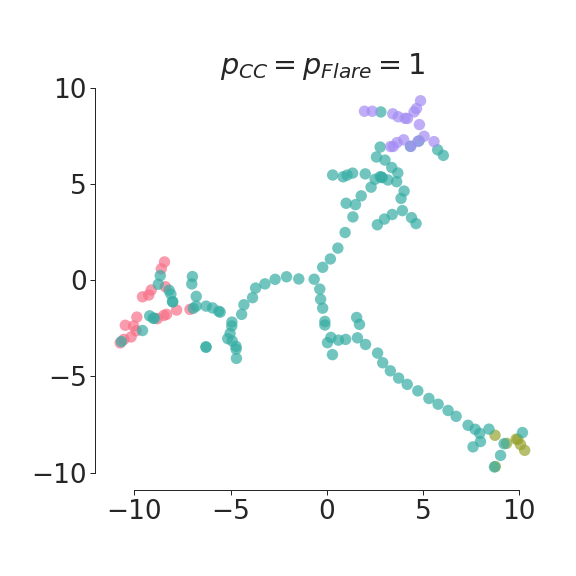}  
	\end{subfigure}\hfill
	\begin{subfigure}[t]{.325\textwidth}
		\centering
		\includegraphics[width=\linewidth]{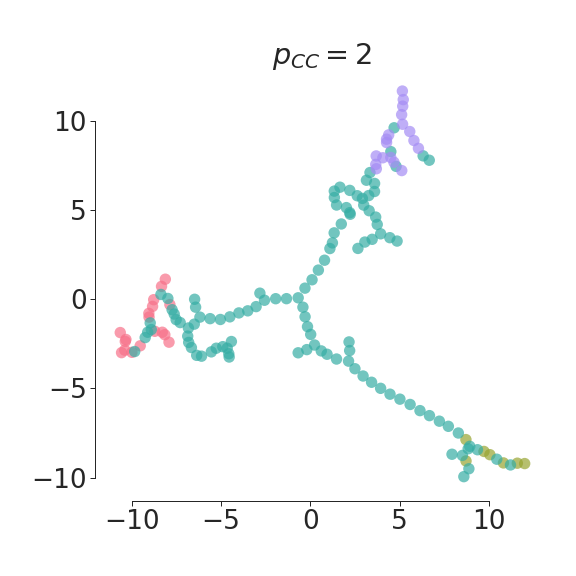}  
	\end{subfigure}
	\begin{subfigure}[t]{.325\textwidth}
		\centering
		\includegraphics[width=\linewidth]{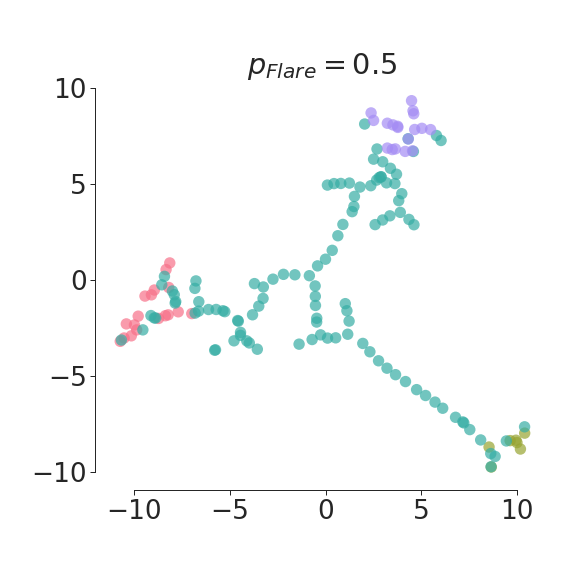}  
	\end{subfigure}\hspace{2mm}
	\begin{subfigure}[t]{.325\textwidth}
		\centering
		\includegraphics[width=\linewidth]{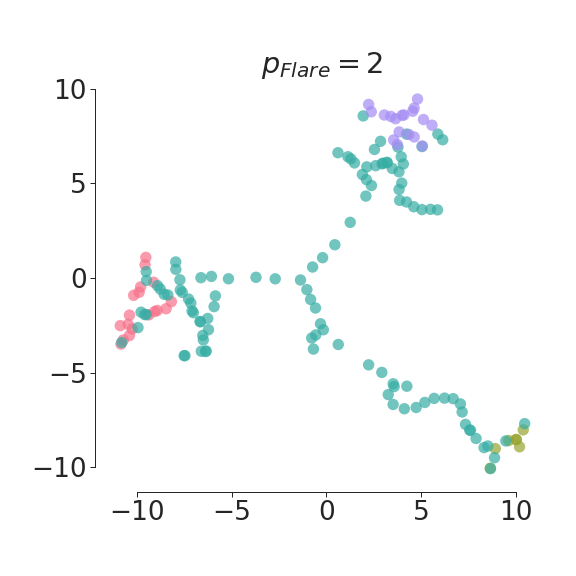}  
	\end{subfigure}
	\caption{The topologically regularized UMAP embeddings of the bifurcating real single cell data  according to (\ref{bifloss}), for various values of $p$.}
	\label{fig:bifotherp}
\end{figure*}

It may already be clear that the same topological information can be postulated through different choices for $p>0$.
To explore this, we considered topologically regularized UMAP embeddings of the real bifurcating single cell data, for the topological loss function as stated in Table \ref{tab:robustness_loss} where the parameter $p$ is varied, i.e.,
\begin{equation}
	\label{bifloss}
	\mathcal{L}_{\mathrm{top}}(\mE)=\sum_{k=2}^\infty (d_k-b_k)^{p_{\text{CC}}}-\left[(d_3-b_3)^{p_{\text{Flare}}}\right]_{\mathcal{E}_{\mE}^{-1}]-\infty,0.75]}.
\end{equation}
We only change one of the exponents, $p_{\text{CC}}$ or $p_{\text{Flare}}$, to investigate their effects independently and show the embeddings in Figure \ref{fig:bifotherp}.

We observe that a small value for $p_{CC}$ leads to many overlapping points as even the smallest gaps contribute significantly to the loss. For $p_{CC} = 2$ there are no large gaps anymore and all points have approximately the same distance to their nearest neighbor. 
With increasing values of $p_{Flare}$, points that are sufficiently far away from the center are pulled towards the leaves of the represented topological model. This increases the persistence of the three clusters. However, there are a few points at the center of the bifurcation that will keep everything connected as regularized through the first part of the topological loss. 
Nevertheless, although different values of $p$ appear to affect the overall spacing between points, we see that the embedded topological shape remains overall recognizable and consistent.

\begin{figure*}[tp]
	\centering
	\begin{subfigure}[t]{.325\textwidth}
		\centering
		\includegraphics[width=\linewidth]{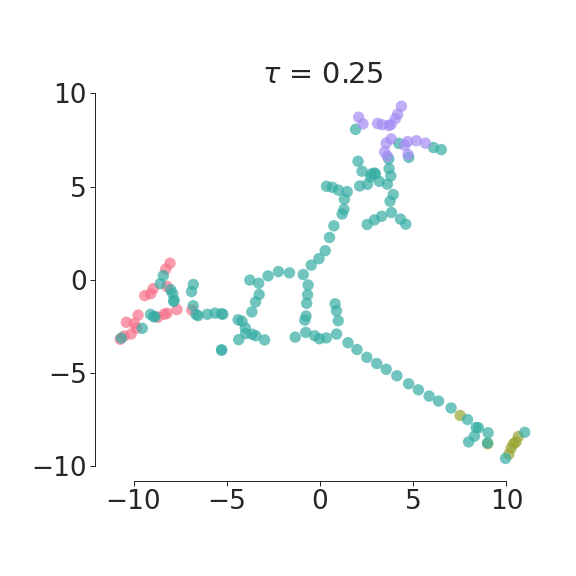}  
	\end{subfigure}\hfill
	\begin{subfigure}[t]{.325\textwidth}
		\centering
		\includegraphics[width=\linewidth]{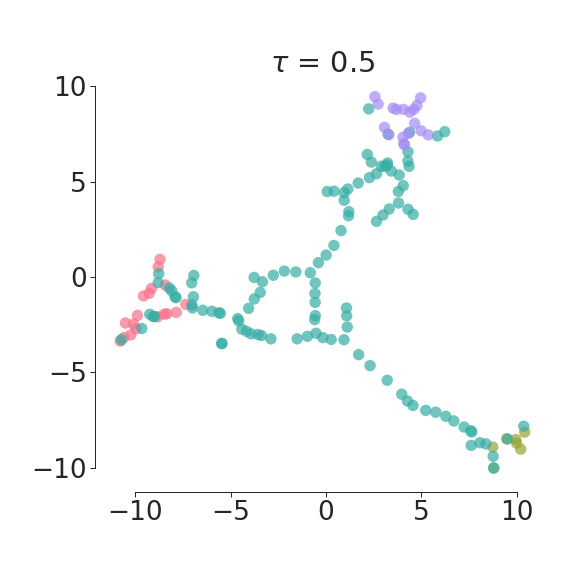}  
	\end{subfigure}\hfill
	\begin{subfigure}[t]{.325\textwidth}
		\centering
		\includegraphics[width=\linewidth]{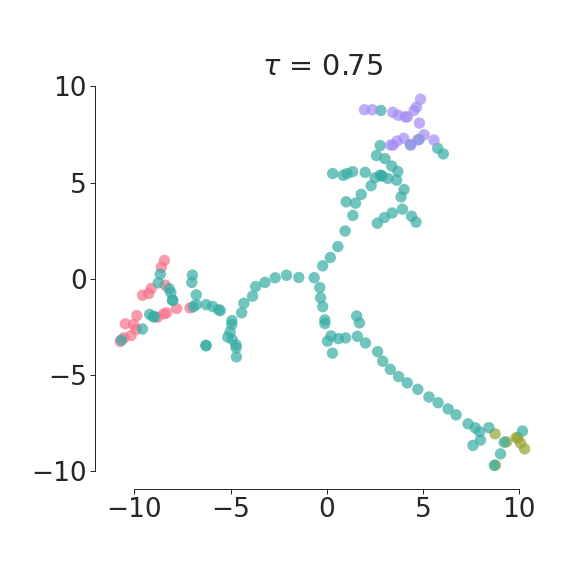}  
	\end{subfigure}
	\caption{The topologically regularized UMAP embeddings of the bifurcating real single cell data, for various functional thresholds $\tau$.}
	\label{CellBifTau}
\end{figure*}

\paragraph{Varying $\tau$}
Figure \ref{CellBifTau} shows the topologically regularized UMAP embedding of the real single-cell data following a bifurcating cell differentiation model, for varying choices of the functional threshold $\tau$.
Recall that $0\leq\tau\leq1$ is a threshold on the normalized distance of the embedded points to their mean as defined in (\ref{centrality}).
Intuitively, $\tau$ specifies how close one looks to the embedding mean, with higher values of $\tau$ allowing more points (thus closer to the embedding mean) to be included when optimizing for three clusters away from the center.
While little differences can be observed between $\tau=0.25$ and $\tau=0.5$, for $\tau=0.75$, points near the center of bifurcation are more stretched towards the endpoints in the embedded model.
This is related to the fact that for higher values of $\tau$, more points close to the center are included when optimizing for three separated clusters.

\subsubsection{Regularization with Weak Signal to Noise Ratio}
\label{sec:weaksignal}
In the synthetic data we used throughout the experiments, the cycle was by construction in the dimensions with highest variance. Thus, the embedding initialization with PCA is already somewhat aligned with the underlying circular structure. Even when initialized differently, the PCA (MSE) loss would contribute to reach this configuration. 
We investigate a setup with a 10-dimensional synthetic dataset (n=100) that again contains a cycle in the first two dimensions, but standardize all dimensions to have equal variance. 

\begin{figure*}[t]
	\centering
	\begin{subfigure}[t]{.35\textwidth}
		\centering
		\includegraphics[width=\linewidth]{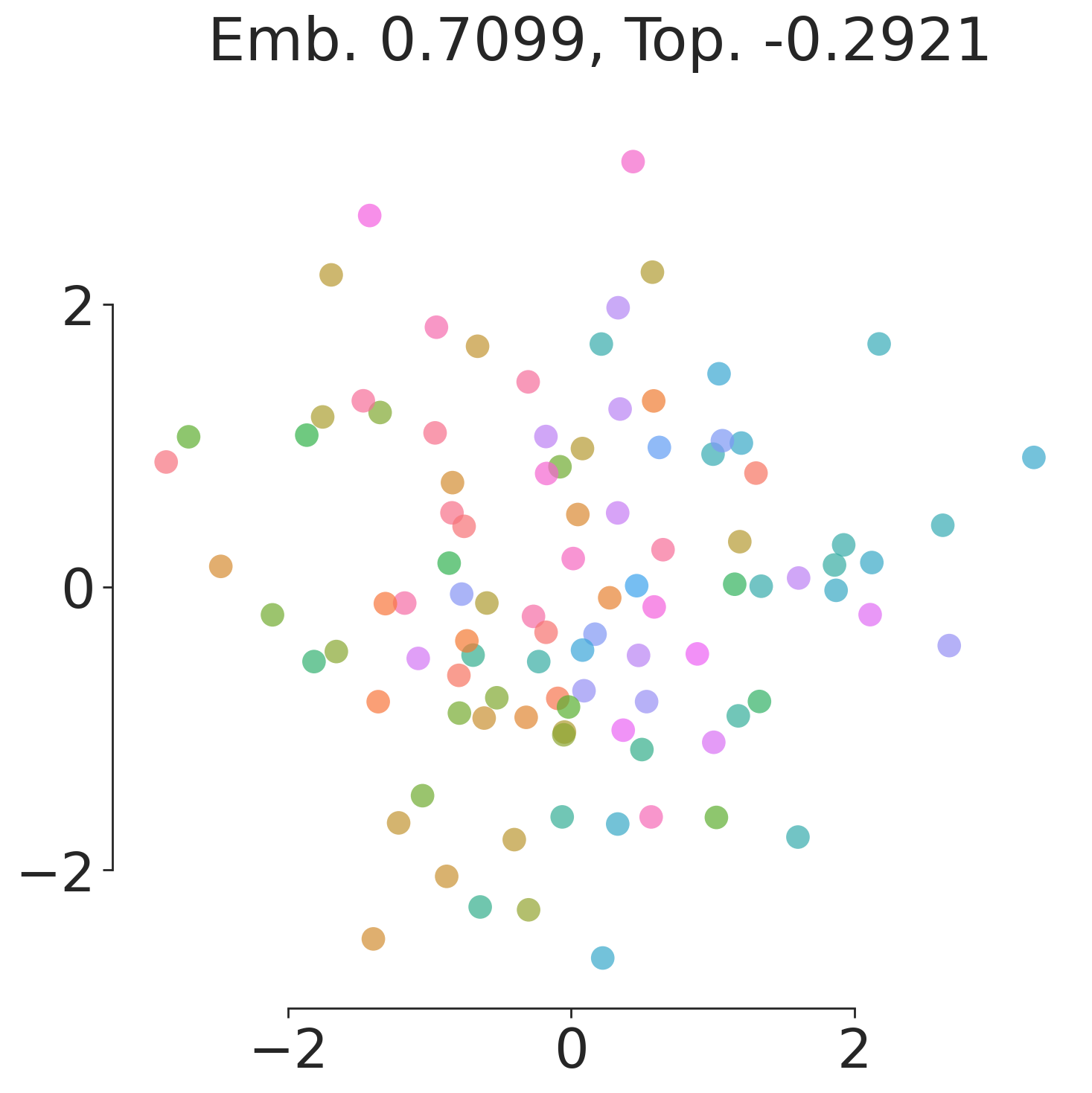}
		\caption{PCA projection.} 
		\label{fig:robustness_zscore_pca}
	\end{subfigure}\hspace{.1\textwidth}
	\begin{subfigure}[t]{.35\textwidth}
		\centering
		\includegraphics[width=\linewidth]{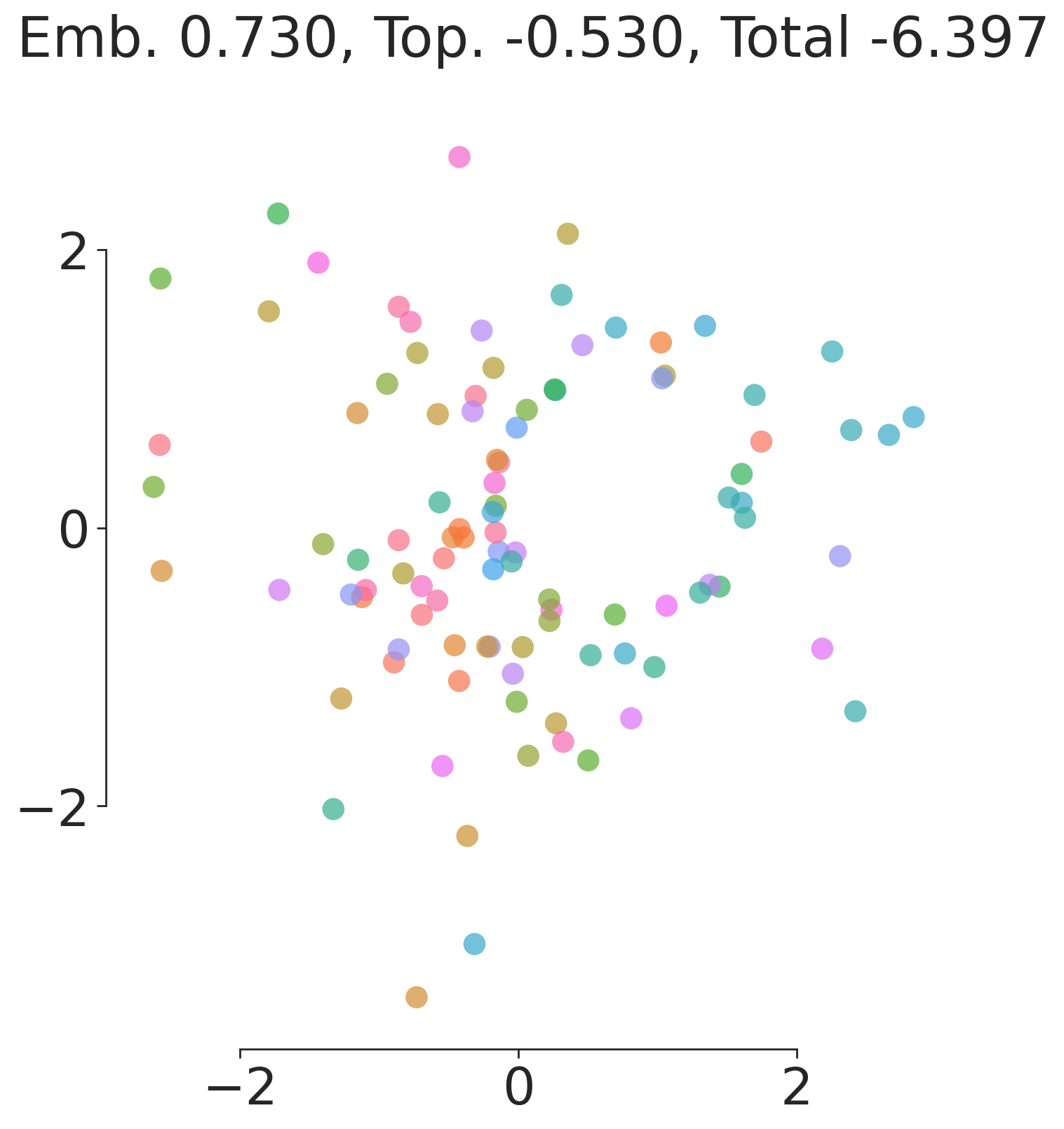} 
		\subcaption{Topologically regularized embedding initialized with \ref{fig:robustness_zscore_pca}.}
		\label{fig:robustness_zscore_pca_regularized}
	\end{subfigure}
	\begin{subfigure}[t]{.35\textwidth}
		\centering
		\includegraphics[width=\linewidth]{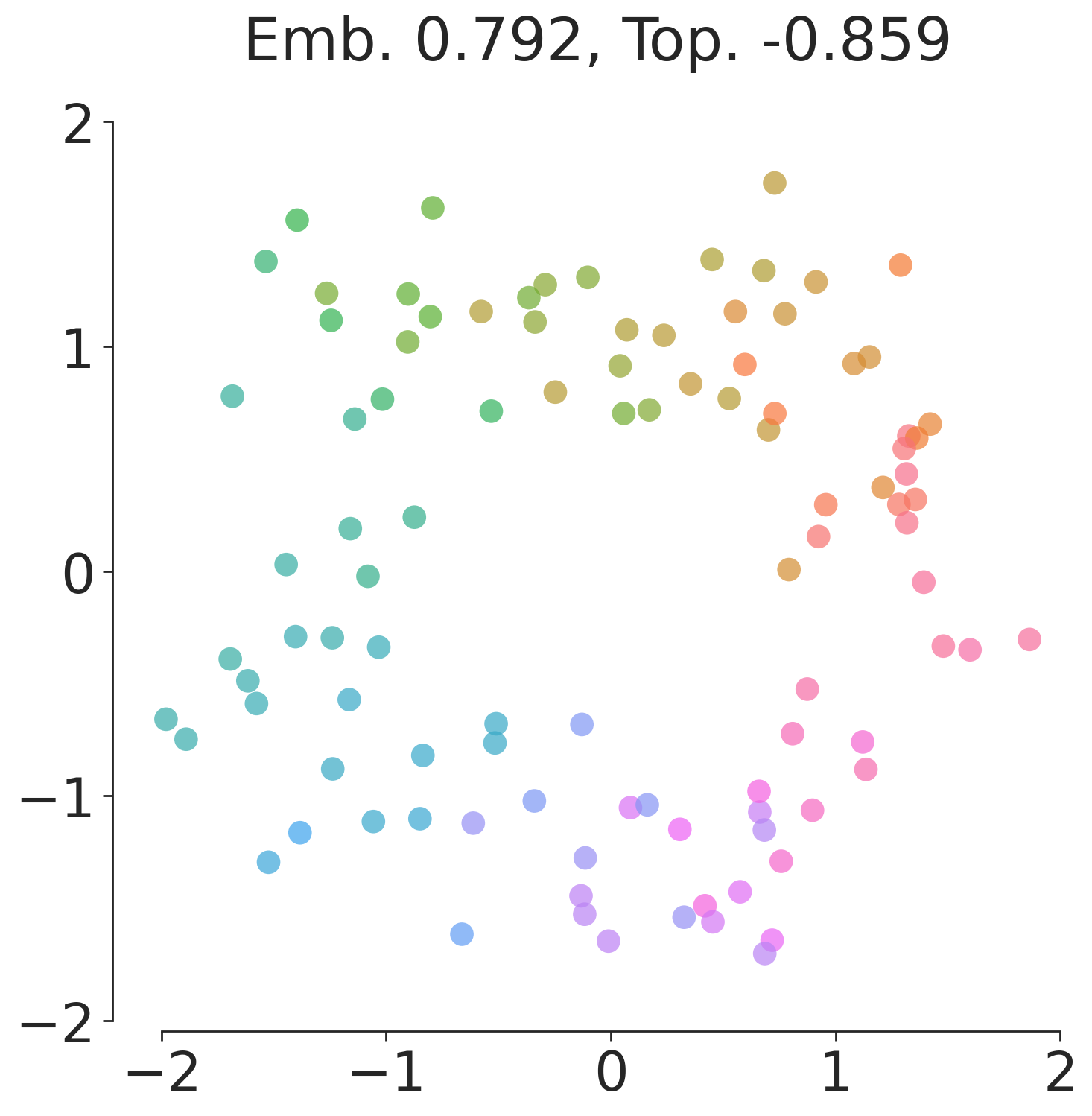} 
		\subcaption{Visualization of the first two data dimensions.}
		\label{fig:robustness_zscore_d12}
	\end{subfigure}\hspace{.1\textwidth}
	\begin{subfigure}[t]{.35\textwidth}
		\centering
		\includegraphics[width=\linewidth]{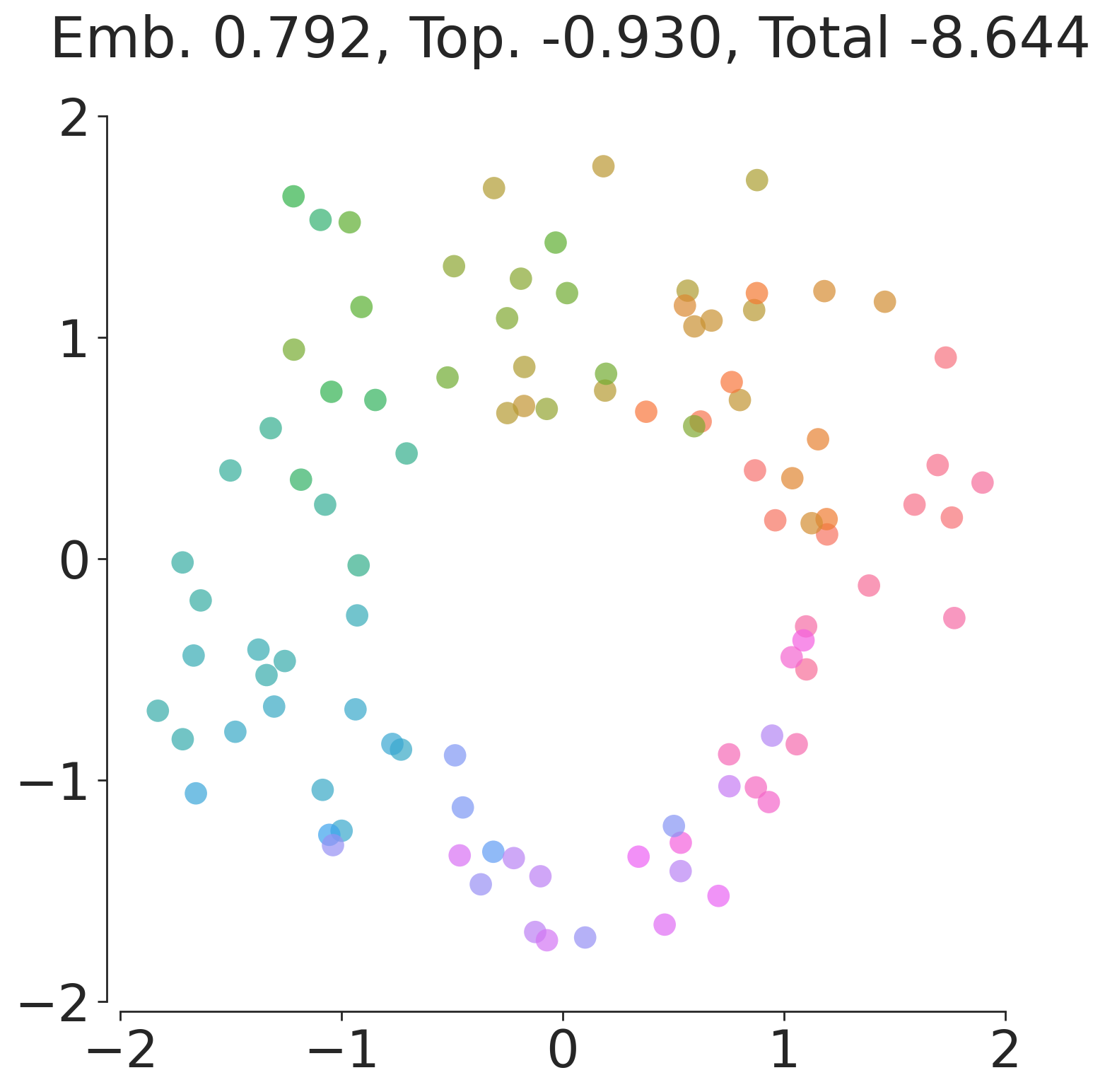}  
		\subcaption{Topologically regularized embedding initialized with \ref{fig:robustness_zscore_d12}.}
		\label{fig:robustness_zscore_d12_regularized}
	\end{subfigure}
	\caption{Regularized embeddings of (n=100, d=10) z-score standardized data in (b) and (d), initialized with the PCA projection (a) and the first two dimensions corresponding to the ground truth circle (c) respectively. The total objective is computed as $\mathcal{L}_{\mathrm{emb}} + 10 \cdot \mathcal{L}_{\mathrm{top}}$.}
	\label{fig:robustness_zscore}
\end{figure*}

The PCA projection (Figure \ref{fig:robustness_zscore_pca}) of this data does not necessarily align with the first two dimensions and does not show the circle. 
Topological regularization initialized with this PCA projection results in an embedding with a spurious hole (Figure \ref{fig:robustness_zscore_pca_regularized}). 
A projection on the first two dimensions and its topologically regularized optimization (Figures \ref{fig:robustness_zscore_d12} and \ref{fig:robustness_zscore_d12_regularized}) both have better objective values due to the lower topological loss. 
It appears that the topologically regularized embedding using PCA initialization in \ref{fig:robustness_zscore_pca_regularized} is a local optimum of the loss that the current local optimization technique is not able to overcome. 
As the total loss of the projection using only the first two dimensions is substantially lower, and the regularized embedding initialized with the ground truth even more so, we presume that using different optimization techniques, or a more intelligent initialization, it could be possible to still identify the correct cycle in this challenging setting. However, this also highlights the loss is not easy to optimize well in more difficult settings.

\subsubsection{Optimizing for Different Topological Priors}
\label{sec:diffloss}
As a user may not always have strong prior expert topological knowledge available, we investigate Q6 by studying how topological regularization reacts to different topological loss functions that do not model the ground truth topological structure of the data.

\paragraph{Wrong Prior on Synthetic Cycle}
We explore the effect of two different topological loss functions regularizing the embedding of the synthetic cycle. For this dataset we know that there is exactly one topological model, i.e., a circle, that generates the data. We compute the regularized PCA embedding with the same parameters as before  (see Table \ref{tab:effectiveness_parameters}) and summarize the loss functions in Table \ref{tab:robustness_loss}.

\begin{figure*}[t]
	\centering
	\begin{subfigure}[t]{.325\textwidth}
		\centering
		\includegraphics[width=\linewidth]{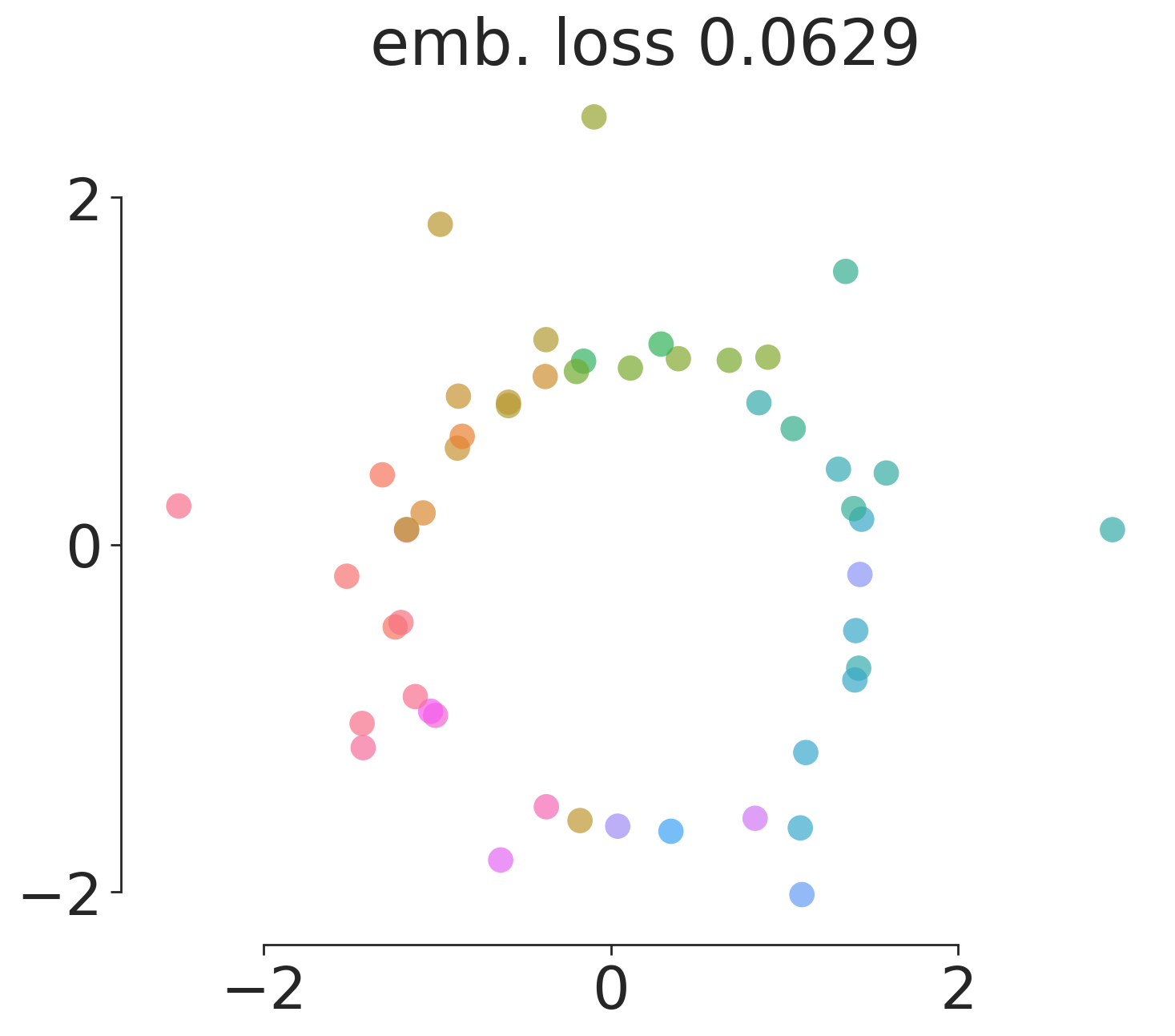} 
		\caption{One circle}
		\label{fig:SynthCircle_onecircle}
	\end{subfigure}\hfill
	\begin{subfigure}[t]{.325\textwidth}
		\centering
		\includegraphics[width=\linewidth]{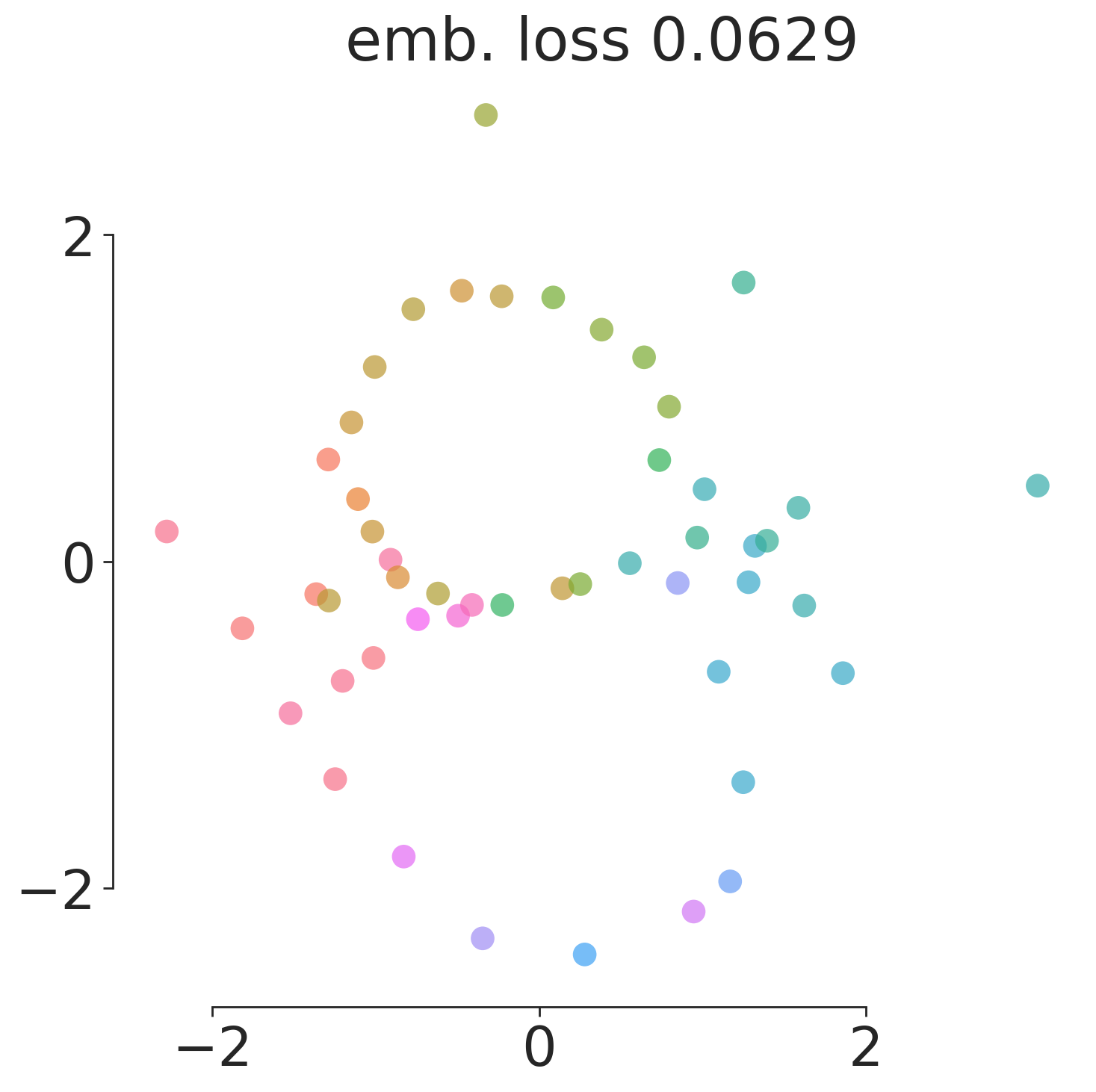}
		\caption{Two equally sized circles}
		\label{fig:SynthCircle_secondcircle}
	\end{subfigure}\hfill
	\begin{subfigure}[t]{.325\textwidth}
		\centering
		\includegraphics[width=\linewidth]{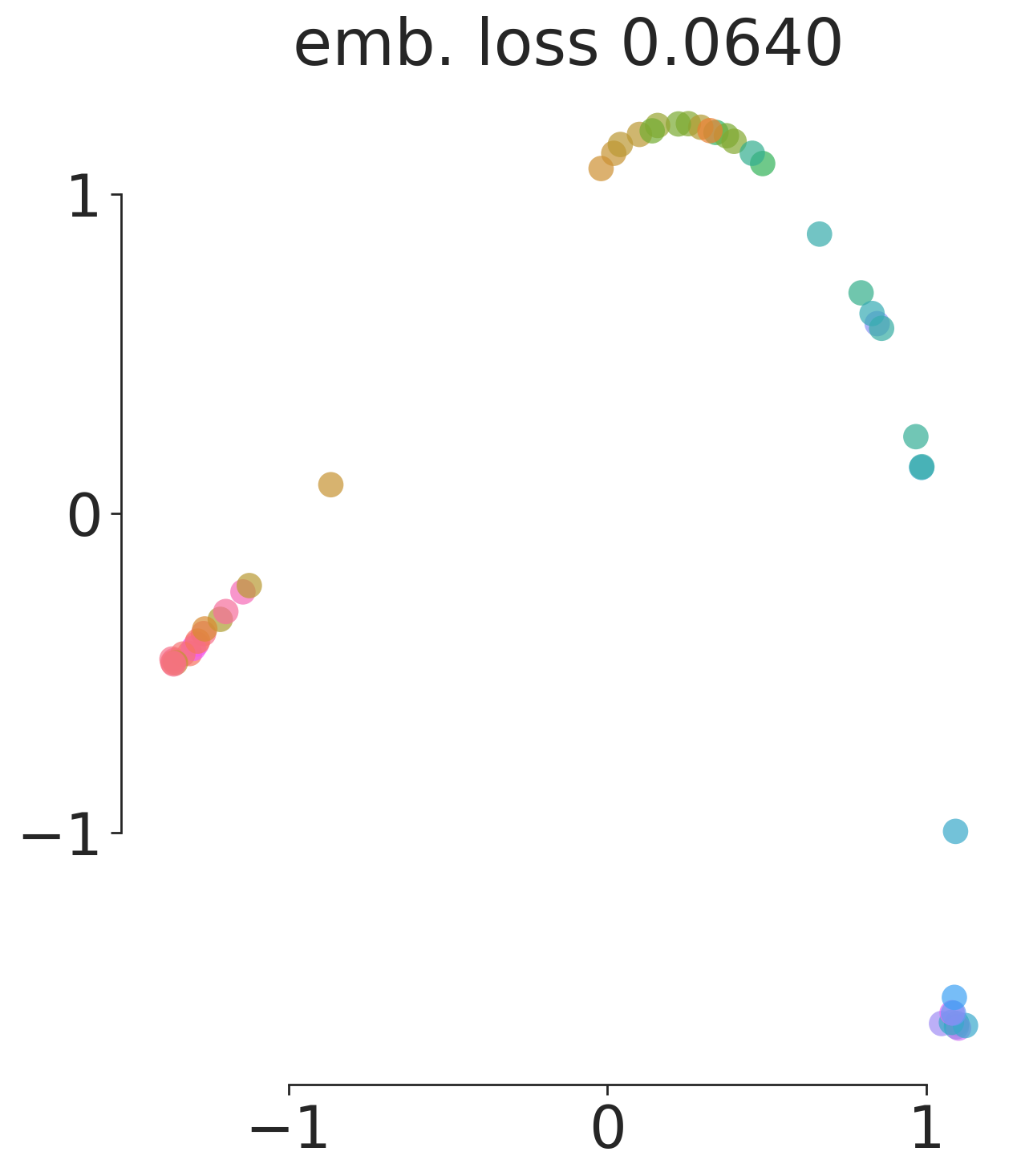} 
		\caption{Connected component}
		\label{fig:SynthCircle_connected}
	\end{subfigure}
	\caption{Topologically regularized embeddings of the 500-dimensional synthetic data for which the ground truth model is a circle. For easier visual comparison, we include in (a) the embedding already shown in \ref{fig:SynthCircleTop}.}
	\label{fig:SynthCircleOther}
\end{figure*}

In Figure \ref{fig:SynthCircle_secondcircle} we show the result of regularizing the embedding with a topological loss maximizing the persistence of the second most prominent cycle. We consider this loss to specify a partially wrong form of prior information, as it imposes that the persistence of the most prominent cycle in the embedding must be high as well. The effect is that the two cycles have similar persistence.
The embedding shown in Figure \ref{fig:SynthCircle_connected} was regularized with a topological loss to minimize all finite (0-dimensional) death times, leading to smaller distances between neighboring points. 
We also show the embedding regularized with the correct circular prior (\ref{fig:SynthCircle_onecircle}) for comparison. 

These experiments show that it is possible to optimize the embedding using a wrong topological prior. In this setting, the reconstruction error for the embedding with one circle and two circles is very similar and do not allow a conclusion which one is correct. Thus, the results must be interpreted with caution: the visual presence of the imposed topological information can in general not be taken as a confirmation that this topological structure is indeed present in the data.

\paragraph{Topological Models in Random High-Dimensional Data}
In the previous experiment, the synthetic data set is very small compared to its dimensionality (n=50, d=500), which gives the projection lots of flexibility: it allows the projection to change the low-dimensional embedding of one point without affecting the others. Thus, the result of that experiment is hardly surprising, and a more interesting question is \emph{in which situations} topological regularization can yield low-dimensional embeddings which show spurious topologies.

\begin{figure*}[t]
	\centering
	\begin{subfigure}[t]{.32\textwidth}
		\centering
		\includegraphics[width=\linewidth]{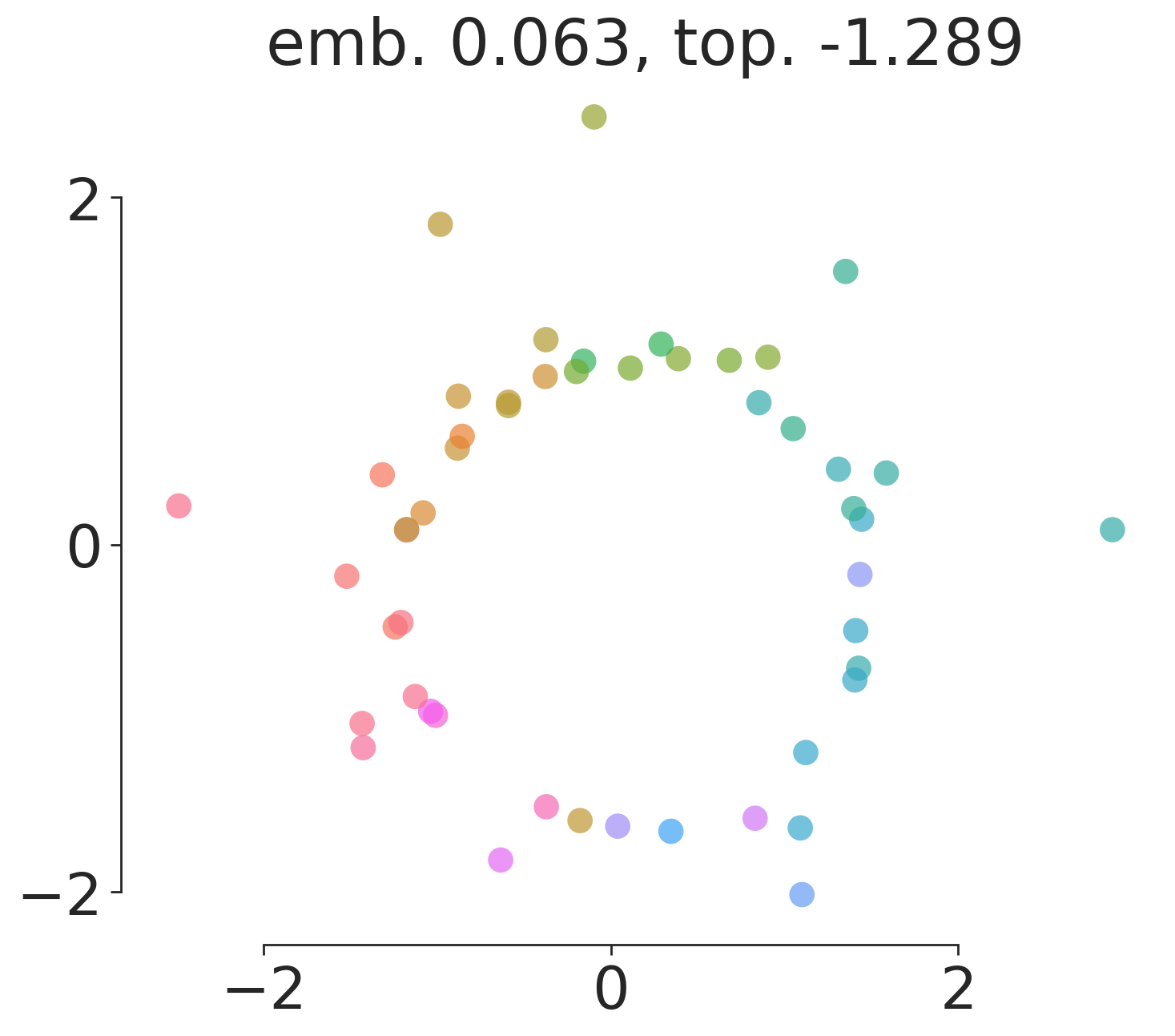} 
		\caption{500-dimensional synthetic data containing a cycle.}
	\end{subfigure}\hfill
	\begin{subfigure}[t]{.3\textwidth}
		\centering
		\includegraphics[width=\linewidth]{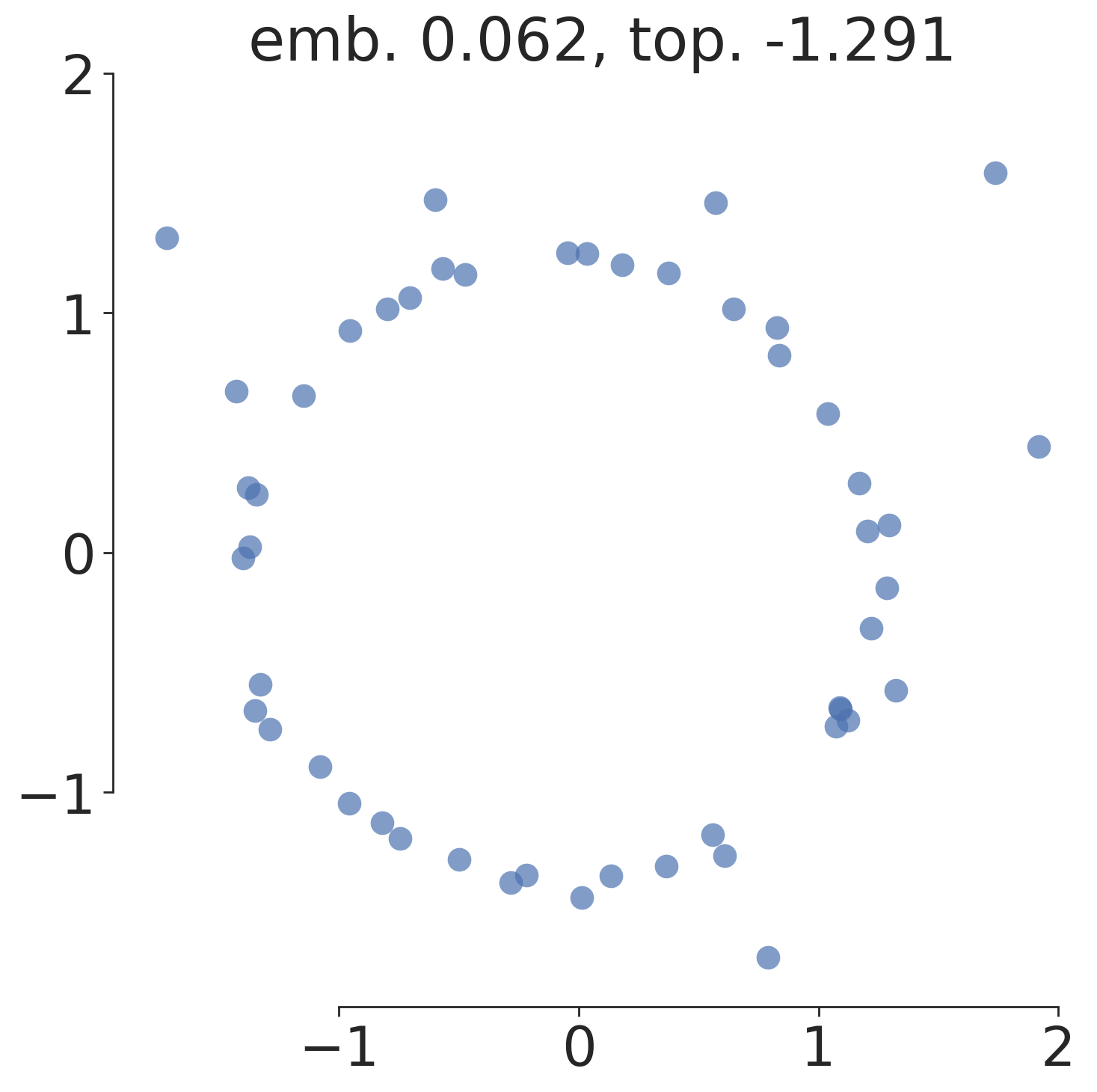}
		\caption{500-dimensional point cloud without circular ground truth.}
		\label{fig:robustness_random_500}
	\end{subfigure}\hfill
	\begin{subfigure}[t]{.33\textwidth}
		\centering
		\includegraphics[width=\linewidth]{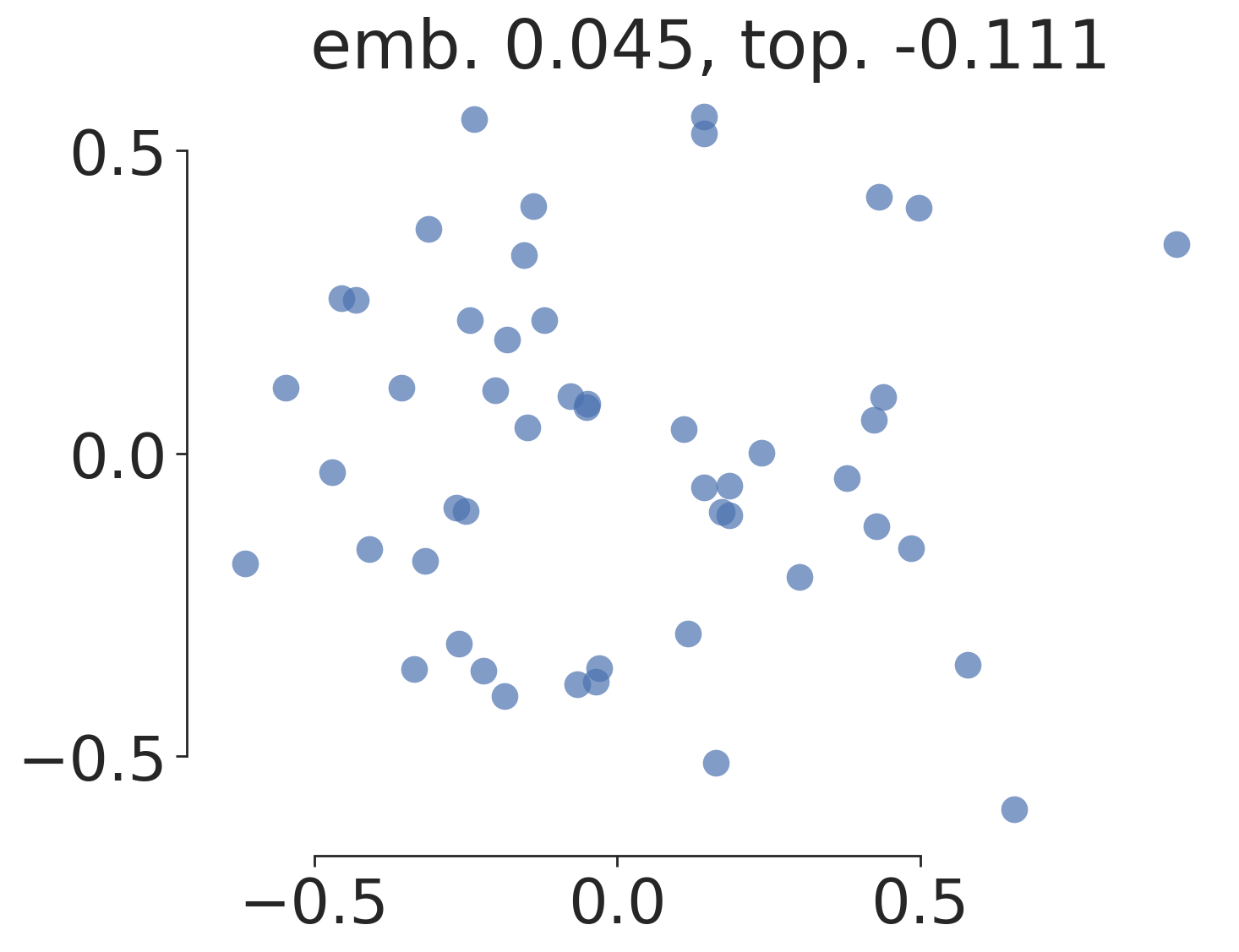}
		\captionsetup{width=.8\linewidth}
		\caption{10-dimensional point cloud without circular ground truth.}
		\label{fig:robustness_random_10}
	\end{subfigure}
	\caption{Topologically regularized embeddings of the 500-dimensional dataset containing the cycle compared to 500 and 10-dimensional point clouds regularized with the same topological loss. To embed a circular structure in the projection of the 10-dimensional data, we used a higher weight on the topological loss ($\lambda_{\mathrm{top}} = 10$).}
	\label{fig:RandTop}
\end{figure*}

Without aiming to be exhaustive, we investigate this question by topologically regularizing a random high-dimensional dataset of the same size and dimensionality (n=50, d=500), and with the same amount of noise as the synthetic cycle, but without the circular signal. The topological loss function for regularization also remains the same as stated in Table \ref{tab:robustness_loss}. We observe that, arguably unsurprisingly, regularizing the embedding of the random data (Figure \ref{fig:robustness_random_500}) results in a visually similar circle than the embedding of the synthetic cycle. However, when reducing the dimensionality of the data to d=10, the cycle is much less pronounced (Figure \ref{fig:robustness_random_10}), even when using a higher weight for the topological loss ($\lambda_{\mathrm{top}}$ = 10).

These experiments show that in data where the number of dimensions is high as compared to the number of data points, the result might show the specified topological structure regardless of whether it is meaningfully present in the data. This may happen even with a small weight for the topological loss. When the number of data points is sufficiently large as compared to the number of dimensions, however, topological regularization is less vulnderable to this. While these results may not be surprising to most readers, we believe it is important to stress that topologically regularized embeddings cannot and should not be used as a test to check whether a certain prior about the data is correct. 

\paragraph{Circular Prior on Bifurcating Single Cell Data}
In the previous experiments, we showed the effects of regularizing an orthogonal projection with a wrong topological prior in different settings. Here, we consider the effect on the non-linear embeddings by UMAP. We assume they may be more prone to model wrong prior topological information as the point coordinates are directly optimized in the embedding space. This offers substantial flexibility even regardless of the dimensionality of the input data.
To explore this, we embed the real bifurcating single cell data for a circular prior through the topological loss function $-(d_1-b_1)$ with $f_{\mathcal{S}} = 0.25$ and $n_{\mathcal{S}} = 10$, evaluated on the 1st-dimensional persistence diagram.

\begin{figure*}[t]
	\centering
	\begin{subfigure}[t]{.32\textwidth}
		\centering
		\includegraphics[width=\linewidth]{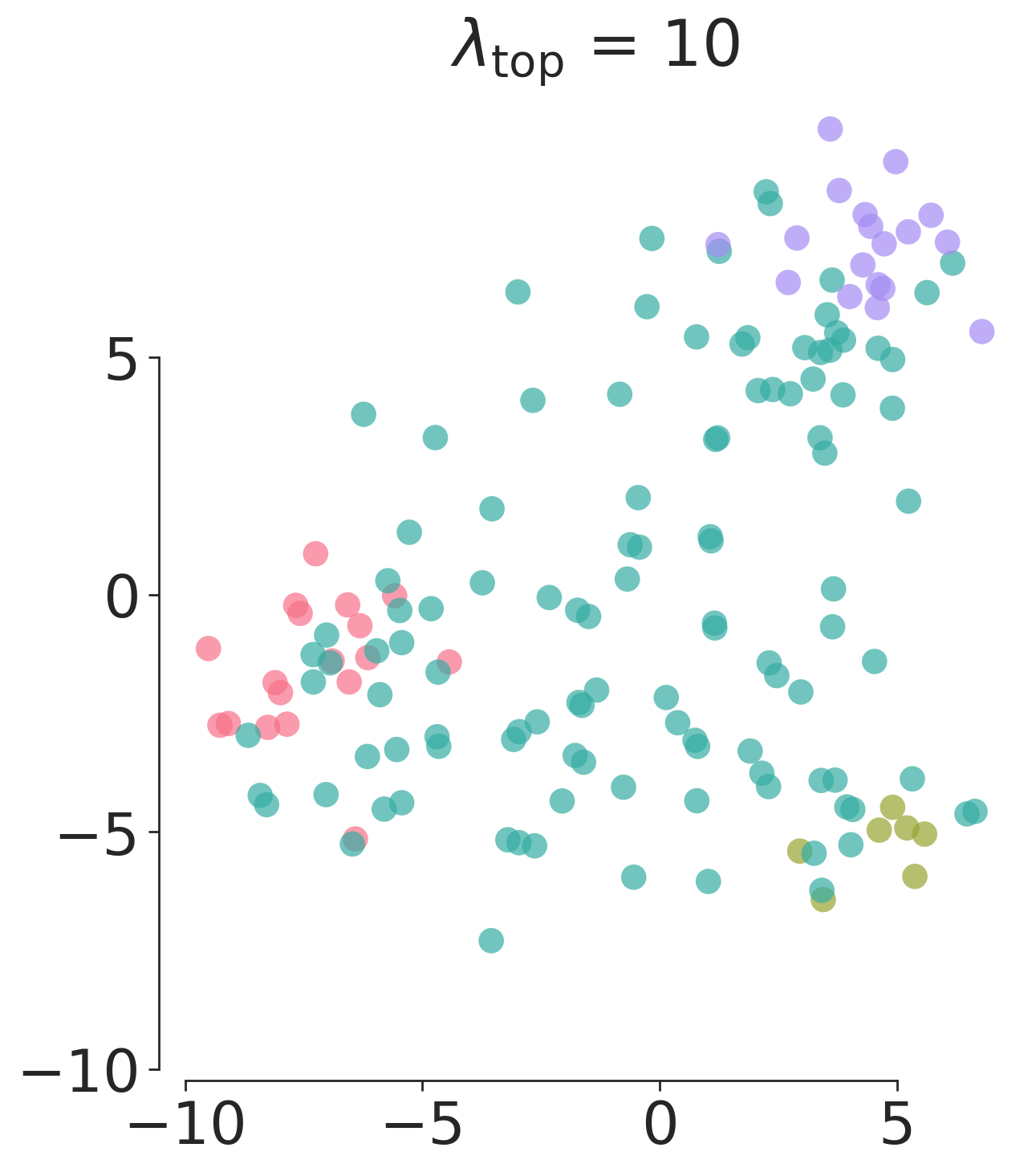}  
	\end{subfigure}\hfill
	\begin{subfigure}[t]{.32\textwidth}
		\centering
		\includegraphics[width=\linewidth]{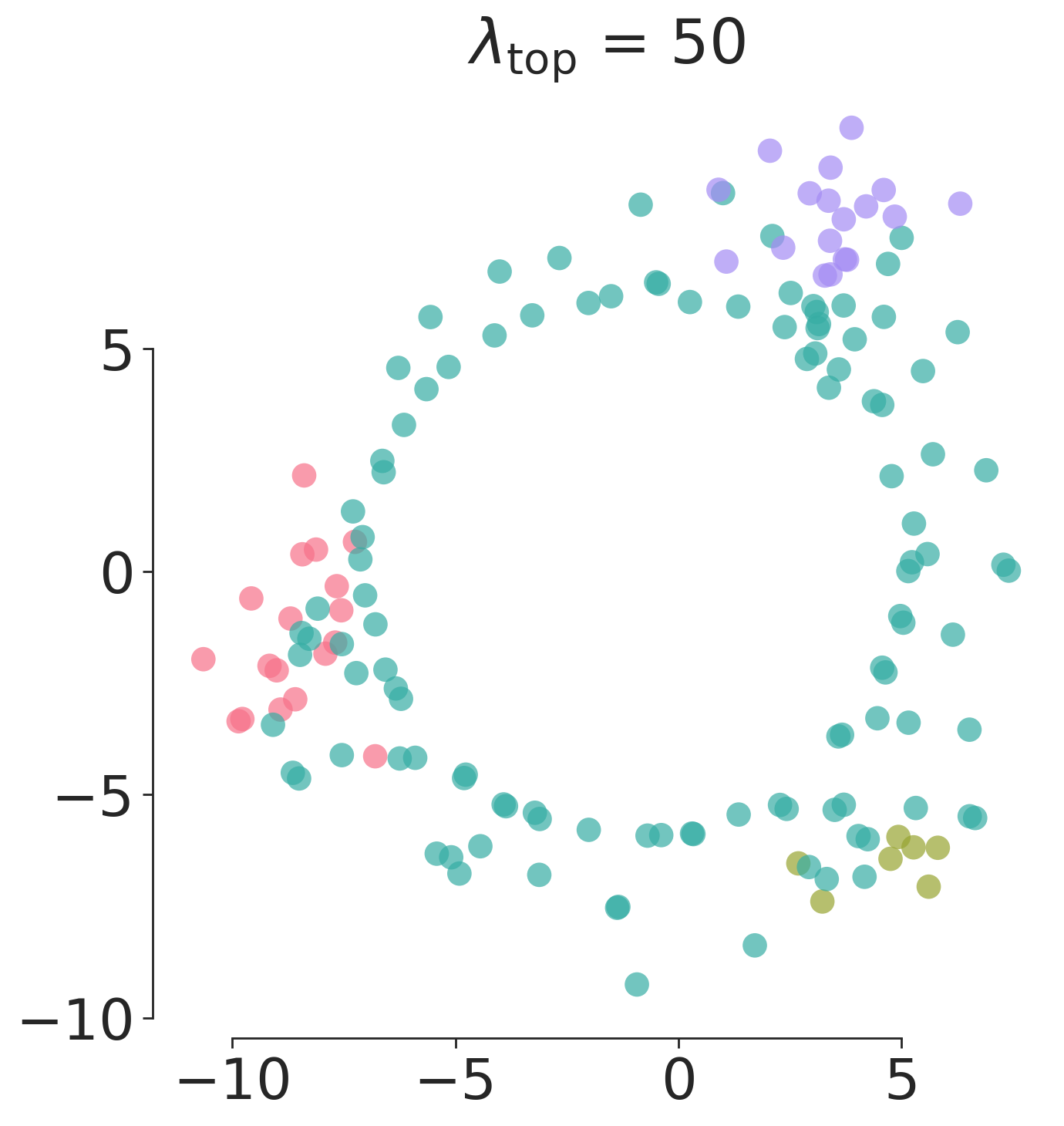}  
	\end{subfigure}\hfill
	\begin{subfigure}[t]{.32\textwidth}
		\centering
		\includegraphics[width=\linewidth]{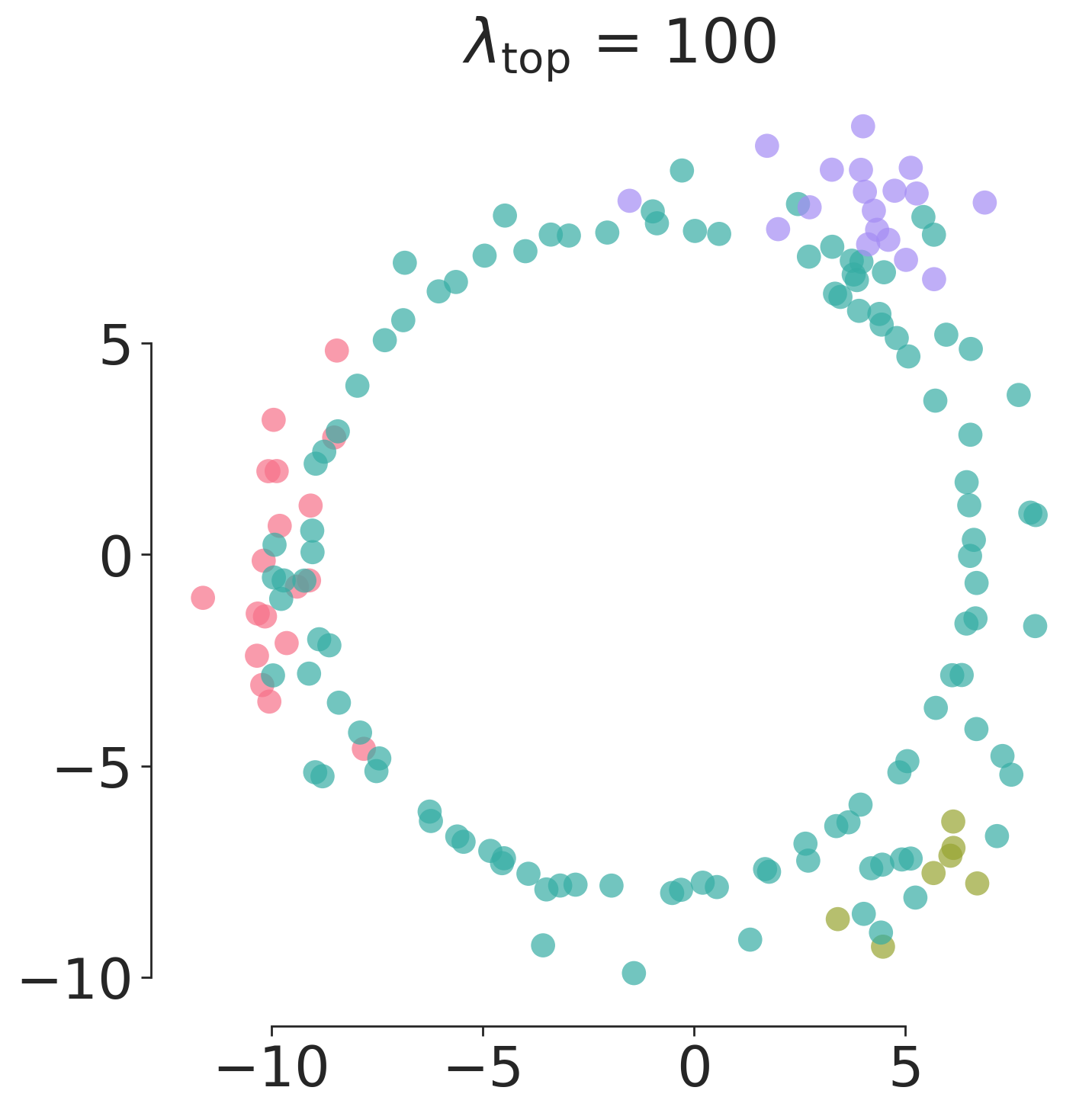}  
	\end{subfigure}
	\caption{Topologically regularized UMAP embeddings of the bifurcating cell data, using potentially wrong prior topological (circular) information. We show embeddings optimized for 250 epochs with various topological regularization strengths $\lambda_{\mathrm{top}}$.}
	\label{fig:CellBifCircleStrong}
\end{figure*}

Figure \ref{fig:CellBifCircleStrong} shows the topologically regularized UMAP embeddings for the circular prior for different topological regularization strengths $\lambda_{\mathrm{top}}$, optimized for 250 epochs. 
Naturally, higher topological regularization strengths $\lambda_{\mathrm{top}}$ will more significantly impact the topological representation in the data embedding. 
We observe that with $\lambda_{\mathrm{top}}=10$ the UMAP loss is preventing the embedding to show a hole. 
With a regularization strength of $\lambda_{\mathrm{top}}=50$ we observe the circle in the regularized embedding. With $\lambda_{\mathrm{top}}=100$ the radius enlarges, but points from the endpoints of the bifurcation lie mostly outside of the circle. 

Returning to research question Q6, we conclude that topological regularization cannot and should not be used to test whether a data set contains a certain topological structure. We showed that especially for high-dimensional data with few samples, and for flexible embedding methods like UMAP, arbitrary structures can be optimized. In addition, the suitability of a topological loss function should not be assessed by the required value of $\lambda_{\mathrm{top}}$ to have a visible effect on the embedding. This value mainly depends on the relative difference between the embedding and topological loss. 
Thus, caution is warranted when no reliable topological prior information is available. 

\section{Conclusion}
\label{SEC::discconc}

We proposed a new approach for representation learning under the name of \emph{topological regularization}, which builds on the recently developed differentiation frameworks for topological optimization. To make topological regularization as accessible as possible, we included an extensive introduction to the minimal theoretical required background on persistent homology.
Our approach led to a versatile and effective way for embedding data according to prior expert topological knowledge, directly postulated through our newly introduced topological loss functions. The proposed topological loss functions are able to model a broad range of global and local topological structures. Moreover, by introducing a sampling approach in combination with these loss functions, we were able to overcome an important robustness challenge. The experiments show that, except in the hardest (low signal to noise ratio) circumstances, topological regularization is highly effective in increasing the saliency of an imposed topological structure. We also showed that including prior topological knowledge provides a promising way to improve subsequent---even non-topological---learning tasks (see also Section \ref{pseudotime}).

\section{Future Work}
\label{sec:future_work}

Our experiments showed that there are various directions for future work. First, a clear limitation of topological regularization is that prior topological knowledge is not always available. How to select the best from a list of priors is thus open to further research (see also Section \ref{sec:diffloss}). An interesting question is whether this limitation can be overcome by combining the prior knowledge with, or partially deriving it from, inferred (topological) features of the high-dimensional data. 

Second, a sensible range for the trade-off parameter between the embedding and the topological loss could alleviate the problem that almost any topological prior can be optimized in embeddings of datasets with few data points as compared to the number of dimensions, or when using flexible (non-linear) embedding methods such as UMAP. Research into how to best set this parameter may open the possibility of using topological regularization also to test whether a data set contains a specified topological structure.

Third, a more robust (non-local) nonlinear optimization method or informed initialization scheme might be necessary for problems with a low signal to noise ratio. As we have seen in Section \ref{sec:weaksignal}, our proof of concept implementation of topologically regularized PCA did not reach the intended configuration, despite the lower objective value. 

Fourth, the design of topological loss functions is arguably still rather complex, and it would be useful to study how to facilitate that design process for lay users. From a foundational perspective, our work provides new research opportunities into extending the developed theory for topological optimization \citep{carriere2021optimizing} to our newly introduced losses, and their integration into data embeddings.

Finally, topological optimization based on combinatorial structures other than the $\alpha$-complex may be of theoretical and practical interest.
For example, point cloud optimization based on graph-approximations such as the minimum spanning tree \citep{vandaele2021stable}, or varying the functional threshold $\tau$ in the loss (\ref{functionalloss}) alongside the filtration time \citep{chazal2009gromov}, may lead to natural topological loss functions with fewer hyperparameters.

\acks{The research leading to these results has received funding from the European Research Council under the European Union's Seventh Framework Programme (FP7/2007-2013) (ERC Grant Agreement no. 615517), and under the European Union’s Horizon 2020 research and innovation programme (ERC Grant Agreement no. 963924), from the Special Research Fund (BOF) of Ghent University (BOF20/IBF/117), from the Flemish Government under the ``Onderzoeksprogramma Artificiële Intelligentie (AI) Vlaanderen'' programme, and from the FWO (project no. G0F9816N, 3G042220, 11J2322N).}

\newpage

\vskip 0.2in
\bibliography{references}

\end{document}